%% file: main.tex
\documentclass[10pt, letter, onecolumn]{arxiv}

\usepackage{kantlipsum, lipsum}
\usepackage{dm-colors}
\usepackage{amsmath}
\usepackage{pstricks, pst-node}
\usepackage{verbatim}
\usepackage{multirow}
\usepackage{array}
\usepackage{longtable}
\usepackage{pdflscape}
\usepackage{scalerel}
\usepackage{booktabs}
\usepackage{enumitem}
\usepackage{xspace}
\usepackage{bm}
\usepackage{bbm}
\usepackage{mathtools}
\usepackage{soul}
\usepackage{epsfig}
\usepackage{graphicx}
\usepackage{amssymb}
\usepackage{colortbl}
\usepackage{csquotes}
\usepackage{setspace}
\usepackage{colortbl}
\usepackage{bbding}
\usepackage{threeparttable}
\usepackage{tabularx,ragged2e}
\usepackage{placeins}
\usepackage{bbding}
\definecolor{darkpastelgreen}{rgb}{0.13, 0.55, 0.13}
\definecolor{darkpastelred}{rgb}{0.55, 0.13, 0.13}
\definecolor{mygray}{rgb}{1, 1, 1}
\usepackage{lmodern}
\usepackage[bottom]{footmisc}

\usepackage{nameref}
\usepackage{varioref}
\usepackage{amssymb}
\usepackage{pifont}
\usepackage{rotating}
\usepackage{graphicx}

\usepackage[pagebackref=false,breaklinks=false,%
            colorlinks=true,bookmarks=true,citecolor=ourdarkblue,%
            urlcolor=ourdarkblue,linkcolor=ourdarkblue]{hyperref}
\usepackage[noabbrev,capitalize]{cleveref}
\usepackage{etoc}
\usepackage{lineno}
\usepackage{algorithm}
\usepackage{makecell} 
\usepackage{algpseudocode}

\usepackage{amsmath}
\usepackage{thmtools}
\usepackage[most]{tcolorbox}
\usepackage{lmodern}

\declaretheoremstyle[
    spaceabove=6pt, spacebelow=6pt,
    headfont=\bfseries, headpunct={.}, headformat={\NAME\ \NUMBER},
    bodyfont=\normalfont,
    postheadspace=0.5em
]{promptstyle}

\declaretheorem[name=Prompt, style=promptstyle]{prompt}

\tcolorboxenvironment{prompt}{
    colback=gray!10!,
    colframe=gray!75!,
    fonttitle=\bfseries,
    title=Prompt,
    boxrule=0.5pt,
    sharp corners,
    breakable
}

\graphicspath{{figures/}}
\definecolor{mygray}{rgb}{0.85, 0.85, 0.85}
\definecolor{midgreen}{rgb}{0.0, 0.75, 0.0}

\newcommand{\change}[1]{{\textcolor{black}{#1}}}

\title{\Large{Evolving Interactive Diagnostic Agents \\in a Virtual Clinical Environment}}

\author[1,2,$\ast$]{Pengcheng Qiu} 
\author[1,2,$\ast$]{Chaoyi Wu} 
\author[3,4,$\ast$]{Junwei Liu} 
\author[1,2]{Qiaoyu Zheng}
\author[1,2]{Yusheng Liao}
\author[3]{\\Haowen Wang}
\author[3]{Yun Yue}
\author[3]{Qianrui Fan}
\author[3]{Shuai Zhen}
\author[3]{Jian Wang}
\author[3]{Jinjie Gu}
\author[1,2]{\\Yanfeng Wang} 
\author[1,2,$\dag$]{Ya Zhang}
\author[1,2,$\dag$]{Weidi Xie}

\affil[1]{\normalsize Shanghai Jiao Tong University, Shanghai, China \authorcr \vspace{0.1cm}}

\affil[2]{\normalsize Shanghai Artificial Intelligence Laboratory, Shanghai, China  \authorcr \vspace{0.1cm}}

\affil[3]{\normalsize Intelligence Healthcare Department, AntGroup, Hangzhou, China  \authorcr \vspace{0.1cm}}

\affil[4]{\normalsize Intelligence Computing and Sensing Laboratory, Peking University, Beijing, China  \authorcr \vspace{0.1cm}}

\affil[$\ast$]{\normalsize Equal contributions\hspace{1cm}}
\affil[$\dag$]{\normalsize Corresponding author\authorcr Ya Zhang: ya\_zhang@sjtu.edu.cn; Weidi Xie: weidi@sjtu.edu.cn}

\begin{document}
\begin{abstract}
\input{content_npj/00_Abstract}

\end{abstract}

\maketitle


\input{content_npj/01_Introduction}

\input{content_npj/02_Results}

\input{content_npj/03_Discussion}

\input{content_npj/04_Methodology}
\input{content_npj/05_Conclusion}

\clearpage
\bibliographystyle{unsrt}
\bibliography{references} 

\section{Acknowledgments}
This work is supported by the National Key R\&D Program of China (No. 2022ZD0160702), and the Scientific Research Innovation Capability Support Project for Young Faculty~(ZY-GXQNJSKYCXNLZCXM-I22).  

\section{Author Contributions}
All listed authors clearly meet the ICMJE 4 criteria. P.Q., C.W., and J.L. contribute equally to this work. Y.Z. and W.X. are the corresponding authors. Specifically, P.Q., C.W., J.L., Q.Z., Y.L., H.W., Y.Y., Q.F., S.Z., J.W., J.G., Y.W., Y.Z., and W.X. all make contributions to the conception or design of the work, and P.Q., C.W., and J.L. further perform acquisition, analysis, or interpretation of data for the work. In writing, P.Q., C.W., and J.L. draft the work. Q.Z., Y.L., H.W., Y.Y., Q.F., S.Z., J.W., J.G., Y.W., Y.Z., and W.X. review it critically for important intellectual content. All authors approve of the version to be published and agree to be accountable for all aspects of the work to ensure that questions related to the accuracy or integrity of any part of the work are appropriately investigated and resolved.

\clearpage


\appendix

\input{content_npj/06_Appendix}

\end{document}

%% file: content_npj/00_Abstract.tex

In this paper, we present a framework for training large language models (LLMs) as diagnostic agents with reinforcement learning, enabling them to manage multi-turn interactive diagnostic processes, adaptively select examinations, and commit to final diagnoses. Unlike instruction-tuned models trained on static case summaries, our method acquires diagnostic strategies through dynamic exploration and outcome-based feedback, mapping evolving patient states to the next optimal examination and subsequent diagnosis.

Our contributions include:
(i) we present a diagnostics world model trained with electronic health records~(EHRs), termed as \textbf{DiagGym}, which enables to emit examination outcomes conditioned on patient history and recommended examination, serving as a virtual clinical environment to support closed-loop in-silico training and evaluation for interactive diagnosis;
(ii) we train an interactive diagnostic agent, \textbf{DiagAgent}, via end-to-end, multi-turn reinforcement learning within the environment, 
to learn dynamic diagnostic policies that optimizes both interactive effectiveness and final accuracy;
\change{(iii) We introduce a comprehensive multi-center diagnostic benchmark, \textbf{DiagBench}, designed to evaluate multi-turn diagnostic interaction trajectories. The benchmark comprises 2.2K physician-validated cases sourced from four distinct distributions, alongside 3.3K physician-written rubrics for granular process-oriented evaluation.}
\change{(iv) Extensive evaluations demonstrate DiagAgent's superior performance \textbf{across both in-domain and out-of-domain (OOD) settings.} DiagAgent significantly outperforms 11 state-of-the-art LLMs (including DeepSeek-v3 and Claude-4-Sonnet) and 2 prompt-engineered agents. In the end-to-end setting, it delivers a {11.20\%} increase in diagnostic accuracy and a 17.58\% boost in examination recommendation F1 score, while consistently maintaining state-of-the-art performance \textbf{across all three external OOD centers}. Furthermore, in rubric-based evaluations, it surpasses the next-best model by 7.1\% in weighted rubric score.}
These findings indicate that learning policies in interactive clinical environments confers dynamic and clinically meaningful long-term diagnostic management abilities that are unattainable through passive training alone.

%% file: content_npj/01_Introduction.tex
\section{Introduction}

Large language models~(LLMs) have made tremendous progress in advancing medical AI, achieving strong performance on rigorous benchmarks, including USMLE-style examinations, and across diverse clinical tasks~\cite{singhal2025toward, singhal2023large, sandmann2025benchmark, mcduff2025towards, gaber2025evaluating}. 
Recent advances in reasoning~\cite{jaech2024openai, guo2025deepseek} and post-training methods such as supervised fine-tuning (SFT)~\cite{wu2024pmc, qiu2024towards, qiu2025quantifying} have spurred interest in diagnostic applications, where the challenge is not simply answering a question but managing a complex, evolving patient case.

Clinical diagnosis, however, is not a static prediction problem. 
It is inherently an \change{interactive} long-term decision-making process under uncertainty: 
clinicians must synthesize partial information, decide which examination to recommend, or whether to commit to a diagnosis, that balance informativeness, timeliness, cost, and safety. Yet, existing LLMs are predominantly trained on passively collected, instruction-style corpora that assume a complete, fixed patient records~\cite{ singhal2025toward, mcduff2025towards, wu2024pmc, chen2023meditron, wang2025baichuan, dubey2024llama} without dynamic evolving. This static paradigm collapses the multi-turn nature of real diagnosis into a single shot, eliminating the interaction with an external environment or revise hypotheses as dynamic evidence accumulates and updates. As a result, state-of-the-art models often fail to plan full diagnostic trajectories~\cite{qiu2025quantifying, hager2024evaluation, liao2024automatic, johri2025evaluation, agentclinic}, including which tests to recommend, when to stop, and when to final diagnose.

\begin{figure}[!t]
    \centering
    \includegraphics[width=\linewidth]{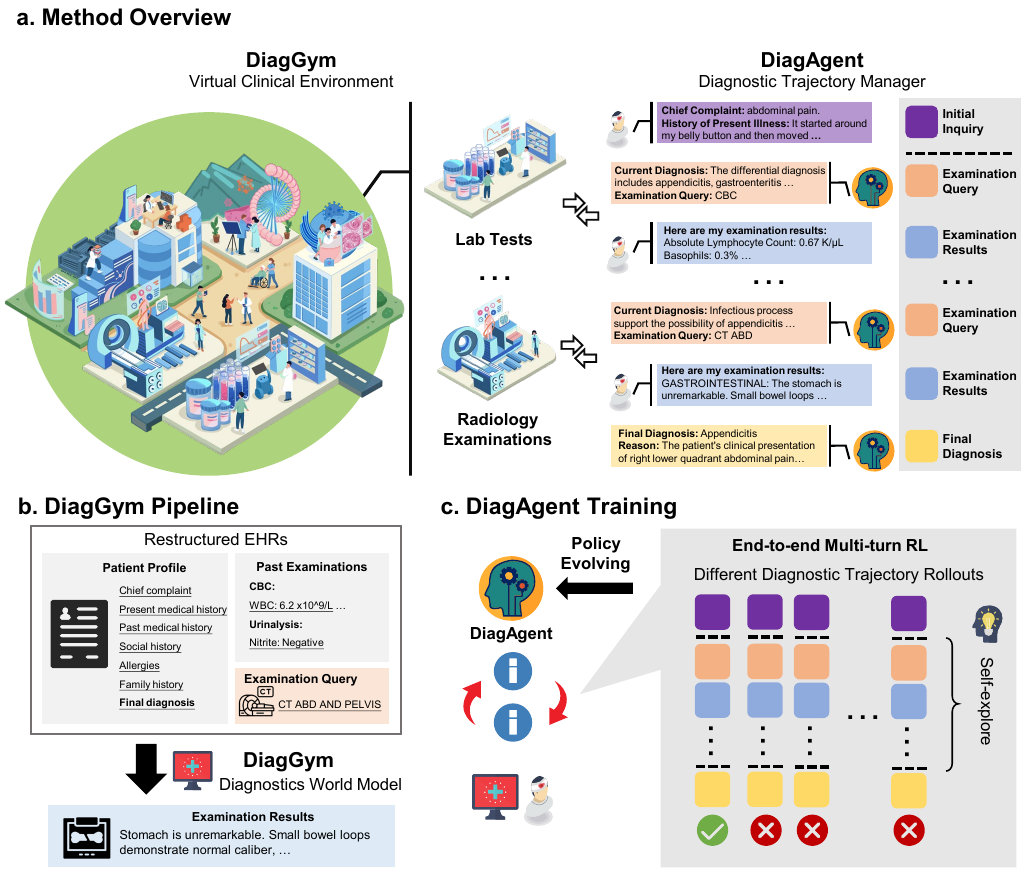}
    \caption{\textbf{Overview of our method.} \textbf{a} illustrates the overview of our method, we establish a virtual clinical environment, DiagGym, that can simulate examination results in real time. Then, within it, we train a diagnostic agent capable of managing multi-turn diagnostic trajectories in a long-term manner, recommendations, recommending diverse examinations until sufficient evidence is gathered for a final diagnosis. \textbf{b} presents the diagnostics world model we constructed based on EHRs, representing a virtual clinical environment. It receives the patient’s basic profile and performed past examinations, and the next examination query as condition input, then simulates the related results as feedback. \textbf{c} depicts the DiagAgent end-to-end multi-turn Reinforcement Learning(RL) training, where the agent interacts with the virtual environment, rolls out different possible diagnostic trajectories, self-explores suitable examination recommendation chains, and iteratively evolves its decision-making policy through end-to-end reinforcement rewards.}
    \label{fig:teaser}
\end{figure}

In this paper, we present an reinforcement learning(RL) based framework for establishing LLMs as \change{interactive} diagnostic agents capable of dynamically suggesting examinations and rendering final diagnoses (Figure~\ref{fig:teaser}a). This advances clinical LLMs from isolated, point-in-time consultations to active patient management within a continually evolving diagnostic window.

Central to our agent-training framework is a diagnostics world model, \textbf{DiagGym}, built on the EHRs with generative model~(Figure~\ref{fig:teaser}b). \textbf{DiagGym} is able to emit examination results conditioned on the former patient state. This enables the safe, closed-loop simulation of real patient reaction, allowing the diagnostic agents to order tests and immediately observe their consequences. Within this environment, we train an \change{interactive} diagnostic agent, \textbf{DiagAgent}~(Figure~\ref{fig:teaser}c) through end-to-end, multi-turn RL. The agent’s policy network maps the current patient state to the next optimal action: either recommending the most informative examination or finalizing the diagnosis. Reward signals are derived from the informativeness of queried examinations, diagnostic accuracy, and efficiency. This approach enables the agent to explore diverse patient trajectories, refine interactive diagnostic strategies through RL without relying on risky, time-intensive real-world implementation. 




\change{To comprehensively assess the final interactive diagnostic agent across diverse clinical settings, we constructed \textbf{DiagBench}, a multi-center benchmark comprising 2.2K physician-validated cases. DiagBench integrates data from four distinct sources: {MIMIC-IV}~\cite{johnson2023mimic}, {PMC-OA case reports}~\cite{pmc_open_access_subset}, {MTSamples}~\cite{mtsamples}, and {DDXPlus}~\cite{ddxplus}. Each case is linked with a physician-verified reference trajectory. Furthermore, to enable fine-grained qualitative evaluation, we engaged senior physicians to author 3.3K weighted evaluation rubrics for a subset of 399 cases. These rubrics specify critical interactive diagnostic steps and rules,  each assigned a weighted score, providing a rigorous standard for measuring the validity of the agent’s interactive decision-making process.}

\change{In experiments, we first evaluate \textbf{DiagGym} as a diagnostics world model for examination result generation. \textbf{DiagGym} significantly outperforms the DeepSeek-v3-based simulator, achieving higher instance-wise consistency (96.90\% vs. 88.81\%) and a lower examination-wise Wasserstein distance (0.128 vs. 1.336), indicating its fidelity. Additionally, its normalized variance aligns with the ground truth ({4.65} vs. 5.31), highlighting robust simulation diversity with reduced mode collapse~\cite{lucic2018gans}.}




\change{We further evaluate \textbf{DiagAgent} against 11 state-of-the-art LLMs (\emph{e.g.}, DeepSeek-v3, Claude-4-Sonnet) and two recent agentic systems under two complementary evaluation settings: single-turn evaluations on real cases and end-to-end evaluations on simulated cases. In the former, the model executes a single interactive diagnostic decision based on real-world observations. In the latter, the agent engages in multi-step interactions with \textbf{DiagGym} to complete full interactive diagnostic workflows.}
\change{On the \textbf{in-domain MIMIC-IV test set} of DiagBench, \textbf{DiagAgent} shows superior performance, outperforming the nearest competitor by {8.94\%} in single-turn diagnostic accuracy and {43.99\%} in examination recommendation hit rate. In end-to-end evaluations, it leads by {11.20\%} in accuracy and 23.09 points {17.58\%} in F1 score, surpassing the runner-up (Claude-4-Sonnet) by 7.1\% in weighted rubric scores. \textbf{DiagAgent} also exhibits robust generalization on the left \textbf{multi-center out-of-distribution test set}. In the single-turn setting, it achieves a 65.30\% hit rate and 92.57\% accuracy, significantly outpacing GPT-4o (52.09\% and 87.06\%). Furthermore, in end-to-end scenarios, \textbf{DiagAgent} attains 63.84\% accuracy compared to Claude-4-Sonnet's 60.12\%, while achieving a significant higher weighted rubric score of 50.27 against 39.07.} These gains highlight our method can effectively equip LLMs with dynamic, clinically meaningful interactive diagnostic capabilities.


%% file: content_npj/02_Results.tex
\section{Problem Formulation}
\label{sec:problem_formulation}

We first formalize the functionality of \textbf{DiagGym} and its role in training \textbf{DiagAgent}.

\textbf{DiagGym.} 
As illustrated in Figure~\ref{fig:teaser}b, we define a diagnostics world model as a conditional textual EHR generator, $\Phi_{\text{env}}$, that generates synthetic examination results conditioned on a dynamically evolving patient state. At step $t$, the patient EHR state is $(\mathcal{B}, E_t)$, where $\mathcal{B}$ is the background patient profile—including chief complaint, present medical history, and the final diagnosis.
The set $E_t = \{(a_1, e_1), (a_2, e_2), \cdots, (a_t, e_t)\}$ represents the past examination records, where each $a_i$ denotes a specific examination item, \emph{e.g.}, ``complete blood count'' or ``CT abdomen examination,'' and $e_i$ is the result. 

DiagGym is designed to generate the potential examination result for the patient based on a specific examination query $a_{t+1}$, as follows:
\begin{equation}
    e_{t+1} = \Phi_{\text{env}}(a_{t+1} \mid E_t, \mathcal{B}),
\end{equation}
where initially $E_0 = \emptyset$, and $e_{t+1}$ denotes the synthetic examination result. 

We frame the training process as a conditional generation task, 
that minimizes the negative log‑likelihood of the ground‑truth examination results.
These results are treated as free text, regardless of whether they are numerical (with numbers directly embedded as text) or textual. The training objective is formalized as:

\begin{equation}
    \mathcal{L}_\text{sim} = -\sum_{t=0}^{T-1} \log \Phi_{\text{env}}(\hat{e}_{t+1} \mid a_{t+1}, E_t, \mathcal{B}),
\end{equation}
where $\hat{e}_t$ represents the ground truth examination result at step $t$ and $T$ denotes the total examination length recorded in a certain EHR. 
More details can be found in Method Section~\ref{sec:simulator_data_construction}.

Once trained, $\Phi_{\text{env}}$ can generate plausible results for any examination and patient state, capturing conditional dependencies across diseases, histories, and prior tests. This capability enables safe and repeatable reinforcement learning (RL) training of diagnostic agents without direct access to real patient records; in other words, it serves as a virtual clinical environment for RL.

\textbf{DiagAgent.} As illustrated in Figure~\ref{fig:teaser}c, 
within DiagGym we train an \textbf{interactive} diagnostic agent, \textbf{DiagAgent}, 
using reinforcement learning. 
Formally, at time step $t$, the agent’s state is defined as $s_t = (\mathcal{I}, E_t)$, 
where $\mathcal{I}$ is the patient’s initial inquiry—including chief complaint, history of present illness, and other relevant presentation details—but, unlike $\mathcal{B}$ in the environment model, contains no information about the final diagnosis. 
The set $E_t = \{(a_1, e_1), \dots, (a_t, e_t)\}$ records the examinations performed so far and their observed results.

For the agent, its action space is defined as $\mathcal{A} = \{a_1, a_2, \cdots, a_N \}$, representing all available clinical examination items and the final diagnosis action. In response to the current state $s_t$, the agent selects an action to recommend the next examination for the patient, based on its policy:
\begin{equation}
    a_{t+1} \sim \pi_\theta(a \mid s_t),
\end{equation}
where $\pi_\theta$ is the learnable policy function parameterized by a large language model $\Phi_{\text{diag}}$:
\begin{equation}
    \pi_\theta = \Phi_{\text{diag}}(s_t).
\end{equation}

\textbf{DiagGym} then returns reasonable examination results, serving as the external environment feedback:
\begin{align}
    &e_{t+1} = \Phi_{\text{env}}(a_{t+1} \mid E_t, \mathcal{B}), \\
    &s_{t+1} = s_t \cup (a_{t+1}, e_{t+1}),
\end{align}
where $s_{t+1}$ is the next state.

The diagnostic trajectory proceeds until the agent selects a final diagnosis action, after which DiagAgent outputs the predicted diagnosis $D$. 
The ultimate training objective is to optimize the interactive policy function $\Phi_{\text{diag}}$ to maximize the expected cumulative reward:
\begin{equation}
    \max_{\Phi_{\text{diag}}} \ \mathbb{E}_{\Phi_{\text{diag}}} \left[ \sum_{t=1}^T \gamma^t \mathcal{R}(s_t, a_t) \right],
\end{equation}
where $\gamma \in [0,1]$ is the discount factor, $T$ is the trajectory length, and $\mathcal{R}$ is the reward function, defined as the sum of three sub-rewards:
\begin{equation}
    \mathcal{R} = \lambda_1 r_\text{diag} + \lambda_2 r_\text{exam} + \lambda_3 r_\text{turn},
\end{equation}
with $\lambda_1, \lambda_2, \lambda_3$ as hyperparameters. Here, $r_\text{diag}$ encourages accurate diagnoses, $r_\text{exam}$ promotes relevant examination recommendations, and $r_\text{turn}$ rewards fewer used turns. The detailed design of the reward function is provided in the Method Section~\ref{sec:Agent_training}.

The final trained DiagAgent can actively manage multi-turn diagnostic trajectories by iteratively interacting with patients, selecting relevant examinations, and ultimately arriving at an accurate final diagnosis.

\section{Results}

In this section, we first evaluate the two key components: 
\textbf{DiagGym}, the high-fidelity diagnostics world model, 
and \textbf{DiagAgent}, the reinforcement-trained diagnostic agent. 
Then, we carry out ablation studies to investigate the effectiveness of our approach design. 
Lastly, we present detailed case studies. \change{Notably, for simplicity, throughout all the following~(including the Method section), we denote all prompts as \texttt{prompt~x}. Further details regarding them are provided in Supplementary Section~\ref{sec:prompt_collection}.}


\subsection{Evaluation for DiagGym}
We first assess the fidelity and reliability of DiagGym against other strong open-source LLMs with the same simulation prompt~(Section~\ref{sec:baselines}).
The evaluation aims to verify whether the world model can generate clinically consistent, context-appropriate examination results that faithfully reflect real-world patterns in EHRs. 

\subsubsection{Evaluation Settings}

We construct an evaluation set of 863 patient cases from MIMIC-IV using the process in Section~\ref{sec:simulator_data_construction}.
These cases span 863 distinct diseases, categorized based on the original ICD codes in MIMIC-IV corresponding to patient admissions, representing a combined 35,548 examination records: on average, 8.77 physical exams, 28.37 laboratory events, 2.04 microbiology events, and 2.01 radiology events per case.

Each case comprises two components: (i) \textbf{patient profile}: baseline information including chief complaint, present and past medical history, social history, allergies, family history, and the final diagnosis; 
(ii) \textbf{examination chain}: the chronological sequence of actual examination results, serving as ground truth.
The simulator’s task is to reconstruct each examination result in sequence, conditioned on the patient profile, all prior examination data, and the current examination query. Thus, during RL training, the simulator can dynamically respond to arbitrary examination queries.


We quantify the generation quality using instance-wise and examination-wise metrics (Figure~\ref{fig:DigGym_results}), the detailed metric calculations are provided in Supplementary Section~\ref{sec:simulator_evaluation}.
These metrics jointly assess whether simulated results match real diagnostic examination results at both the instance-wise and the examination-wise.


\textbf{Instance-wise metrics.} These metrics assess the quality of generated examination sequences at the level of individual patient cases. This evaluation utilizes \change{two} methods: \change{\textbf{heuristic metrics~(BLEU~\cite{papineni2002bleu} \& normalised MAE} and \textbf{LLM-as-a-judge} (GPT-4o, version gpt-4o-2024-08-06).}
The LLM-as-a-judge metrics are defined as below:


\vspace{-6pt}
\begin{itemize}\setlength\itemsep{3pt}
    \item \textbf{Step-level similarity} measures how closely the simulator’s output for each examination step matches the corresponding real-world record, given the patient profile and all prior ground-truth results. Similarity is scored on a 0-5 scale, where 5 indicates perfect medical equivalence or high similarity to the reference. Both the automated evaluator and physician raters assign an independent score from 0 (no similarity) to 5 (perfect equivalence).
    
    \item \textbf{Full-chain consistency} evaluates the coherence of an entire generated sequence where each step depends on the simulator’s \textbf{previously generated outputs}. This setting mirrors the use of reinforcement learning, 
    prioritising internal clinical consistency over word-for-word agreement with ground truth. 
    Consistency is judged using a binary score (1 = consistent, 0 = inconsistent). The automated evaluator provides a binary judgment using \texttt{prompt~\ref{prompt:eval_full_chain_consistency}}. In the physician rating, raters must make a judgment ensuring the sequence maintains adherence to medical common sense, features appropriate calibration of severity, and is free of internal contradictions or conflicts across all reported findings. 
\end{itemize}

\textbf{Examination-wise metrics.}  
\change{To ensure DiagGym captures the full spectrum of real-world distributions rather than just specific instances, we adopt distribution-level metrics for different examination items, which are widely adopted in EHR generation~\cite{yoon2023ehr, li2023generating, yan2022multifaceted}. These metrics assess global fidelity and diversity to guard against mode collapse~\cite{lucic2018gans}, which might otherwise circumvent instance-level evaluations.}

\vspace{-6pt}
\begin{itemize}\setlength\itemsep{3pt}
    \item \textbf{Numerical fidelity~\&~diversity} 
    compares generated numerical values ({\em e.g.}, red blood cell counts) to real distributions. Fidelity is quantified via the 1-Wasserstein distance (lower is better), while diversity is measured as the normalized variance of the generated distribution (higher reflects broader coverage and less mode collapse). Metrics are averaged across all selected numerical tests (Supplementary Section~\ref{sec:selected_examinations_collection}).

    \item \textbf{Free-text fidelity~\&~diversity:} 
    compares the generated narrative reports~({\em e.g.}, CT abdomen findings) in embedding space using BioLORD~\cite{remy2024biolord}. Fidelity is measured by the Fréchet Inception Distance~\cite{yu2021frechet}~(lower is better) and diversity by Intra-LPIPS~\cite{ojha2021few}~(higher indicates more inter-case variation). 
    Metrics are averaged over the selected free-text examinations (Supplementary Section~\ref{sec:selected_examinations_collection}).

\end{itemize}

\begin{figure}[!t]
    \centering
    \includegraphics[width=\linewidth]{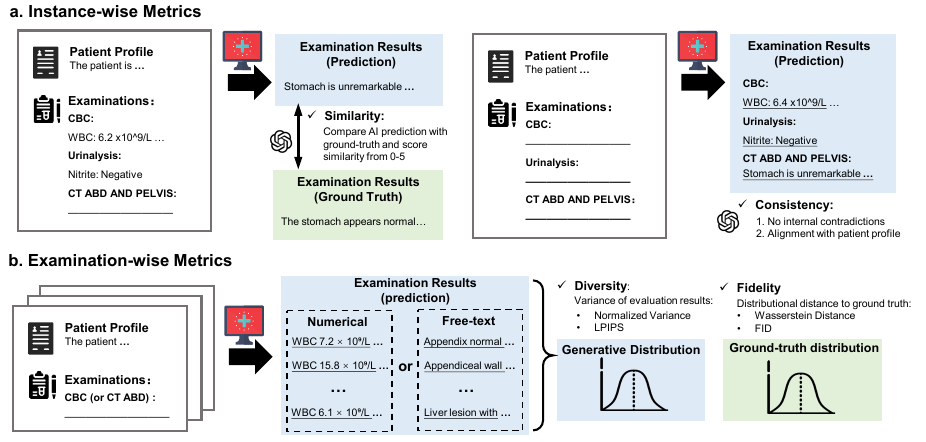}
    \caption{Overview of simulator evaluation settings. \textbf{a} Instance-wise metrics: GPT‑4o assesses the quality of generated examination results on an individual patient case level. \textbf{b} Examination-wise metrics: fidelity and diversity are evaluated by comparing the statistical distributions of generated examination results against those from real cases.}
    \label{fig:DigGym_results}
\end{figure}

During the calculation of examination-wise metrics, we fit the generative distribution by sampling from the first examination step, \emph{i.e.}, directly conditioned on the `patient profile' without giving extra past examination results. This aims to preserve enough sampling randomness, thus better reflect the underlying distribution. Instead, providing too much context, \emph{e.g.}, detailed past examinations, could overly constrain the generation and make it excessively deterministic. For each examination item, we only use test cases where that specific examination was actually performed, ensuring that both the generative and real distributions are calculated using the same patient set.



\textbf{Computational metrics.} 
In addition to the generation quality, we assess computational efficiency, as the simulator must respond rapidly to support interactive training. Two metrics are reported: \textbf{minimal GPU}, the lowest number of GPU cards required for deployment. \textbf{Time~(GPU$\cdot$s)}, the average wall-clock time to generate a single examination result, multiplied by the minimal GPU count. All measurements were obtained on NVIDIA A100 80GB GPUs.

\change{\textbf{Physician Evaluation.} Beyond former automated metrics, we also involve three independent physicians, each with over 10 years of experience, to further rate the generated cases. Due to cost constraints, the physician evaluation utilized a random sample of 100 instances.
The rating rules are the same as the instance-wise LLM-as-a-judge metrics, \emph{i.e.}, \textbf{step-wise similarity} and \textbf{full-chain consistency}, straightforwardly replacing LLMs with human experts.}

\subsubsection{Results Analysis} 

In this part, we analyze the main quantitative results under automated evaluation, computational efficiency and physician evaluation.

\textbf{Automated Evaluation}

The automated evaluation results are provided in Table~\ref{tab:ehrgenerator_eval_results}. Across nearly all metrics, DiagGym delivers state-of‑the-art performance, combining high‑fidelity with substantially greater computational efficiency.


\change{For instance-wise quality, \textbf{DiagGym} demonstrates superior performance across both heuristic and LLM-as-a-judge metrics. It significantly outperforms competitive baselines, for example, achieving a similarity score of 3.565 compared to 2.576 for DeepSeek-v3 and 2.495 for Qwen2.5-72B. While large-scale models maintain reasonable consistency, they struggle to align closely with the ground truth. Conversely, smaller models (e.g., Qwen2.5-7B) show marked degradation in both consistency and alignment.}

\change{Regarding examination-wise metrics, DiagGym effectively balances fidelity and diversity, achieving distributions closest to real-world data, \emph{e.g.}, lowest Wasserstein Distance~(0.128) and FID~(0.747). In contrast, baselines exhibit polarized behaviors: models like DeepSeek-v3-671B and MedGemma-27B show high diversity but suffer from large distribution gaps (poor fidelity), whereas Qwen2.5-72B produces overly deterministic outputs with minimal diversity.}


\textbf{Computational Efficiency}

Existing large-scale baselines trade performance for heavy resource demands. For instance, a single simulation with DeepSeek-v3-671B necessitates a multi-GPU setup (16$\times$A100) and exceeds one minute of computation time. In sharp contrast, DiagGym operates on a single A100 GPU with sub-second latency ($\approx$0.5s). This order-of-magnitude acceleration, achieved without compromising generative quality, is critical for the rapid interactions required in diagnostic agent reinforcement learning.

\change{\textbf{Physician Evaluation}}


\change{As shown in Table~\ref{tab:diaggym_human_eval_results}, DiagGym consistently outperforms baselines, achieving the highest average similarity score~(4.49) and a dominant majority-vote consistency of 95\%, whereas baselines fluctuate between 44\% and 74\%. Qualitatively, physicians noted that DiagGym effectively avoids common baseline failure modes, such as over-extrapolation and logical inconsistencies, maintaining balanced coverage and strict alignment with the case context.
}

Collectively, these results demonstrate that DiagGym is a high-fidelity, diverse, and computationally efficient world model. It is reliable and well-suited to serve as a virtual clinical environment for dynamic diagnostic agent training with Reinforcement Learning, substantially outperforming current open-source baselines. \change{More detailed qualitative case study are shown in Supplementary Section~\ref{sec:diaggym_case_study}}.

\begin{table}[!t]
\renewcommand{\arraystretch}{1.3} 
\footnotesize
\centering
\caption{Quantitative comparison of different models as diagnostics world model for generating synthetic patient examination results. We report the mean value with the \textbf{95\% Confidence Interval} in brackets or range where applicable. Metrics include Similarity, Consistency, \change{BLEU, normalised MAE(NMAE),} Fidelity (Wasserstein Distance, FID), and Diversity (Normalized Variance, LPIPS).}
\label{tab:ehrgenerator_eval_results}
\resizebox{\textwidth}{!}{
\begin{tabular}{l|cc|cccc|cc|cc}
\toprule
\multicolumn{1}{c|}{\multirow{3}{*}{Model}} & \multicolumn{2}{c|}{\multirow{2}{*}{Computational Metrics}} & \multicolumn{4}{c|}{\multirow{2}{*}{Instance-wise Metrics}} & \multicolumn{4}{c}{Examination-wise Metrics} \\ 
\cline{8-11}
 & & & & & & & \multicolumn{2}{c|}{Numerical} & \multicolumn{2}{c}{Free-text}\\
 \cline{2-11} 
 & \makecell{Minimal \\ GPUs$\downarrow$} & \makecell{Time \\ (GPU$\cdot$s)$\downarrow$} & Similarity$\uparrow$ & Consistency(\%)$\uparrow$ & \change{BLEU$\uparrow$} & \change{NMAE$\downarrow$} & \makecell{Normalized \\Variance$\uparrow$} & \makecell{Wasserstein\\ Distance$\downarrow$} & LPIPS$\uparrow$ & FID$\downarrow$   \\
\midrule
GT & - & - & - & - & - & - & 
\makecell{5.31} & - & 
\makecell{\textbf{0.427} \\ \scriptsize{[0.31-0.51]}} & - \\
\midrule
DeepSeek-v3-671B & 16 & 62.72 & 
\makecell{2.576 \\ \scriptsize{[2.56-2.59]}} & 
\makecell{88.81 \\ \scriptsize{[86.6-90.8]}} & 
\makecell{4.37 \\ \scriptsize{[4.21-4.54]}} & 
\makecell{0.691 \\ \scriptsize{[0.643-0.740]}} & 
\makecell{\textbf{24.56} \\ \scriptsize{[1.37-70.0]}} & 
\makecell{1.336 \\ \scriptsize{[0.39-3.02]}} & 
\makecell{0.237 \\ \scriptsize{[0.20-0.27]}} & 
\makecell{4.158 \\ \scriptsize{[3.68-4.63]}} \\

Qwen2.5-7B & \textbf{1} & 0.54 & 
\makecell{2.181 \\ \scriptsize{[2.16-2.20]}} & 
\makecell{81.64 \\ \scriptsize{[79.0-84.4]}} & 
\makecell{2.54 \\ \scriptsize{[2.43-2.65]}} & 
\makecell{0.831 \\ \scriptsize{[0.776-0.889]}} & 
\makecell{20.18 \\ \scriptsize{[6.25-42.9]}} & 
\makecell{9.680 \\ \scriptsize{[2.37-19.4]}} & 
\makecell{0.256 \\ \scriptsize{[0.22-0.28]}} & 
\makecell{4.800 \\ \scriptsize{[4.27-5.42]}} \\

Qwen2.5-72B & 4 & 18.68 & 
\makecell{2.495 \\ \scriptsize{[2.48-2.51]}} & 
\makecell{92.40 \\ \scriptsize{[90.5-94.1]}} & 
\makecell{4.43 \\ \scriptsize{[4.28-4.59]}} & 
\makecell{0.962 \\ \scriptsize{[0.901-1.026]}} & 
\makecell{1.21 \\ \scriptsize{[0.45-2.13]}} & 
\makecell{1.839 \\ \scriptsize{[0.38-4.28]}} & 
\makecell{0.183 \\ \scriptsize{[0.14-0.21]}} & 
\makecell{4.905 \\ \scriptsize{[4.26-5.53]}} \\

MedGemma-27B & 2 & 9.1 & 
\makecell{2.438 \\ \scriptsize{[2.42-2.46]}} & 
\makecell{89.87 \\ \scriptsize{[87.8-91.8]}} & 
\makecell{3.78 \\ \scriptsize{[3.66-3.93]}} & 
\makecell{0.767 \\ \scriptsize{[0.714-0.824]}} & 
\makecell{18.70 \\ \scriptsize{[2.66-39.7]}} & 
\makecell{16.936 \\ \scriptsize{[0.86-38.8]}} & 
\makecell{0.341 \\ \scriptsize{[0.30-0.37]}} & 
\makecell{4.158 \\ \scriptsize{[3.80-4.47]}} \\
\midrule
DiagGym & \textbf{1} & \textbf{0.52} & 
\makecell{\textbf{3.565} \\ \scriptsize{[3.55-3.58]}} & 
\makecell{\textbf{96.90} \\ \scriptsize{[95.6-98.0]}} & 
\makecell{\textbf{45.97} \\ \scriptsize{[45.36-46.54]}} & 
\makecell{\textbf{0.442} \\ \scriptsize{[0.413-0.474]}} & 
\makecell{4.65 \\ \scriptsize{[1.39-8.80]}} & 
\makecell{\textbf{0.128} \\ \scriptsize{[0.11-0.15]}} & 
\makecell{0.379 \\ \scriptsize{[0.24-0.47]}} & 
\makecell{\textbf{0.747} \\ \scriptsize{[0.61-0.91]}} \\
\bottomrule
\end{tabular}}
\end{table}




\begin{table}[!t]
\renewcommand{\arraystretch}{1.2} 
\footnotesize
\centering
\caption{Human ratings comparing DiagGym with baseline models. Similarity (0-5) is reported for each of the three physicians and as the mean across physicians. Consistency is a binary judgment, we report the per-physician percentage of cases judged clinically coherent and the majority-vote consistency rate (percentage of cases deemed coherent by at least two of three physicians).}
\label{tab:diaggym_human_eval_results}
\resizebox{\textwidth}{!}{
\begin{tabular}{l|ccc|c|ccc|c}
\toprule
\multicolumn{1}{c|}{\multirow{2}{*}{Model}}&  \multicolumn{4}{c|}{Similarity} & \multicolumn{4}{c}{Consistency(\%)}  \\ 
\cline{2-9}
 & Physician 1 & Physician 2 & Physician 3 & Avg. Score & Physician 1 & Physician 2 & Physician 3 & Majority Vote \\
\midrule
DeepSeek-v3-671B & 4.66 & 4.49 & 3.11 & 4.09 & 54.00 & 58.00 & 42.00 & 54.00 \\
Qwen2.5-72B      & 4.50 & 4.37 & 3.04 & 3.97 & 46.00 & 44.00 & 32.00 & 44.00 \\
MedGemma-27B     & 4.56 & 4.28 & 2.82 & 3.89 & 73.00 & 75.00 & 56.00 & 74.00  \\
\midrule
DiagGym          & \textbf{4.71} & \textbf{4.70} & \textbf{4.05} & \textbf{4.49} & \textbf{96.00} & \textbf{94.00} & \textbf{92.00} &\textbf{ 95.00} \\
\bottomrule
\end{tabular}}
\end{table}

\subsection{Evaluation for DiagAgent}
We next evaluate our diagnostic model, \textbf{DiagAgent}, against leading LLMs and agentic systems~(baselines described in Section~\ref{sec:baselines}), focusing on its ability to manage complete multi-turn diagnostic trajectories.

\subsubsection{Evaluation Settings}
We evaluate \textbf{DiagAgent} on \textbf{DiagBench}, \change{a multi-center benchmark designed to rigorously assess multi-turn diagnostic capabilities across diverse clinical settings. As shown in Table~\ref{tab:diagbench_stats}, the benchmark comprises a total of \textbf{2,257 physician-validated patient cases}. Considering the training distribution of DiagAgent, we stratify the benchmark into: (1) a MIMIC-IV \textbf{In-Domain~(ID)} test set (750 cases), representing critical care and emergency medicine; and (2) a multi-center \textbf{Out-of-Domain (OOD)} test set (1,507 cases) aggregated from PMC-OA, MTSamples, and DDXPlus, covering multi-source patient distributions, including global case reports, outpatient records, and synthesized differential diagnosis cases based on statistic prior.} \change{To enable fine-grained evaluation, the benchmark is further annotated with \textbf{3,318 physician-authored rubrics} (973 for MIMIC-IV and 2,345 for the OOD subsets). These rubrics delineate critical diagnostic interaction logic and necessary examinations within the diagnostic process.}
The detailed DiagBench construction pipeline can be found in Section~\ref{sec:method_diagagent_rubrics}.

All cases are standardized into a unified format containing three key element:
(i) \textbf{initial inquiry}: patient initialized presentation details (chief complaint, current and past medical history, and other relevant information), forming the starting point for the diagnostic process;
(ii) \textbf{referenced multi-turn diagnostic trajectory}: a physician-curated sequence extracted from real EHR records, serving as the ground-truth reference;
(iii) \textbf{final diagnosis}: the final confirmed clinical diagnosis outcome.
Notably, cases are structured following the same simulation case-construction pipeline used earlier, ensuring each includes a patient profile compatible with the simulator.


\begin{table}[!t]
    \centering
    \footnotesize
    \renewcommand{\arraystretch}{1.25}
    \caption{\textbf{Data Statistics of DiagBench.} The benchmark spans one in-domain center and three out-of-domain (OOD) centers. The OOD subsets focus on diverse clinical scenarios beyond the ICU setting. All cases underwent rigorous human validation and rubric construction.}
    \label{tab:diagbench_stats}
    \begin{tabular}{lll l rr}
        \toprule
        \textbf{Domain} & \textbf{Source Dataset} & \textbf{Access} & \textbf{Clinical Setting} & \textbf{\# Cases} & \textbf{\# Rubrics} \\
        \midrule
        \textbf{In-Domain} & MIMIC-IV~\cite{johnson2023mimic} & \href{https://physionet.org/content/mimiciv/3.1/}{[Link]} & Critical Care and Emergency Medicine & 750 & 973 \\
        \midrule
        \multirow{3}{*}{\textbf{Out-of-Domain}} & PMC-OA~\cite{pmc_open_access_subset} & \href{https://ftp.ncbi.nlm.nih.gov/pub/pmc/}{[Link]} & Open Access Biomedical Case Reports & 631 & 1075 \\
         & MTSamples~\cite{mtsamples} & \href{https://www.mtsamples.com/}{[Link]} & Transcribed Outpatient Records & 379 & 620 \\
         & DDXPlus~\cite{ddxplus} & \href{https://figshare.com/articles/dataset/DDXPlus_Dataset_English_/22687585}{[Link]} & Synthesized Differential Diagnosis & 497 & 650 \\
         \cmidrule(l){2-6} 
         & \textit{OOD Total} & - & -& \textit{1,507} & \textit{2,345} \\

        \midrule
        \textbf{Total} & \textbf{All 4 Centers} & \textbf{-} & \textbf{Comprehensive Dynamic Benchmark} & \textbf{2,257} & \textbf{3,318} \\
        \bottomrule
    \end{tabular}
\end{table}

We consider two complementary evaluation scenarios with corresponding metrics, namely, single-turn and end-to-end evaluation, as detailed below.

\textbf{Single-turn Evaluation}

In this setting, we evaluate the DiagAgent in the single-turn setting. As shown in Figure~\ref{fig:static_results}a,b, here, we leverage both the ground truth `initial inquiry' and partial `referenced multi-turn diagnostic trajectory' as input. The DiagAgent is directly forced~(prompt details can be found in Section~\ref{sec:baselines}) to provide an examination recommendation or make a final diagnosis based on the preceding oracle diagnostic trajectory, extending the process by one additional turn, without self-deciding which action to perform next.
Such single-turn evaluations are conducted at each agent response turn recorded in the referenced multi-turn trajectory. For example, for the 750 cases sourced from MIMIC-IV, this expanding into 4,485 turns for evaluation, consisting 3,735 intermediate turns for examination recommendation and 750 final turns for diagnosis.

For examination recommendation turns, we calculate the hit ratio based on whether the suggested examination is in the key examination list. Specifically, for MIMIC-IV, the relevant list is derived directly from the historical records of the current admission; for other datasets, key examinations are extracted from the raw text. To assess final diagnosis, we employ Accuracy. To avoid issues with synonyms and the inclusion relationships between examinations, we utilize GPT-4o to judge whether an examination recommendation appears within the remaining part in the `referenced multi-turn diagnostic trajectory' with \texttt{prompt~\ref{prompt:instruction_to_check_if_exam_in_the_list}}.

\textbf{End-to-End Evaluation} 

In this setting, we adopt an end-to-end evaluation approach. The diagnostic trajectory is initialized with the `initial inquiry', after which diagnostic agents continually interact with the environment and sequentially propose examination queries until they determine that a final diagnostic decision can be made. Throughout the trajectory, all returned examination results are simulated by the DiagGym, conditioned on the background `patient profile,' ensuring that all queried information is available. 

This evaluation setting more closely reflects real‑world clinical practice, highlighting the model’s ability to dynamically construct a complete diagnostic trajectory, autonomously determining both the timing and type of actions based on the patient’s evolving condition. While the assessment necessarily relies on the external diagnostics world model to simulate examination results, this stems from a fundamental limitation of real‑world EHRs: they only contain examinations that were actually performed. Consequently, when the diagnostic agent suggests an examination that was not carried out for a given patient, there is no corresponding result in the EHR, making it impossible to evaluate that decision or its downstream effects. These inherent gaps prevent the use of real‑world EHRs for fully interactive, end‑to‑end evaluation, as the diagnostic trajectory would be repeatedly interrupted by missing information.

After obtaining the complete predicted diagnostic trajectory, we evaluate performance from two perspectives.

\textbf{First}, as shown in Figure~\ref{fig:dynamic_results}a, for the all cases in DiagBench we employ automatic metrics to assess the efficacy of examination recommendations and the accuracy of the final diagnosis.
(i) \textbf{Examination recommendation} compares the examination items proposed by the model with those in the reference multi‑turn diagnostic trajectory, and computing precision, recall, and F1-score to measure recommendation quality; 
(ii) \textbf{Final diagnosis} assesses the accuracy of the model’s ultimate diagnosis after completing the multi‑turn interaction, by directly comparing it to the ground‑truth diagnosis.


\change{\textbf{Second}, as illustrated in Figure~\ref{fig:results_rubrics}a, for cases annotated with physician-authored rubrics, we derived a weighted rubric score to more accurately reflect expert satisfaction with the generated diagnostic interactions. Following the methodology of HealthBench~\cite{arora2025healthbench}, we adopt an LLM-as-a-judge approach, prompting GPT-4o to verify whether the diagnostic process adheres to specific criteria and computing a final score based on the assigned clinical weights. This metric serves as a necessary complement to standard automated metrics. Detailed rubrics can capture the temporal logic~(which examinations should be recommended first) and critical rule-out steps essential to rigorous clinical reasoning~(which examinations must be done). More implementation details and human validation of these metrics are provided in Supplementary Section~\ref{sec:diagagent_evaluation} and~\ref{sec:human_validation} respectively.}


\subsubsection{Single-turn Evaluation Analysis}

\begin{figure}[!t]
    \centering
    \includegraphics[width=\linewidth]{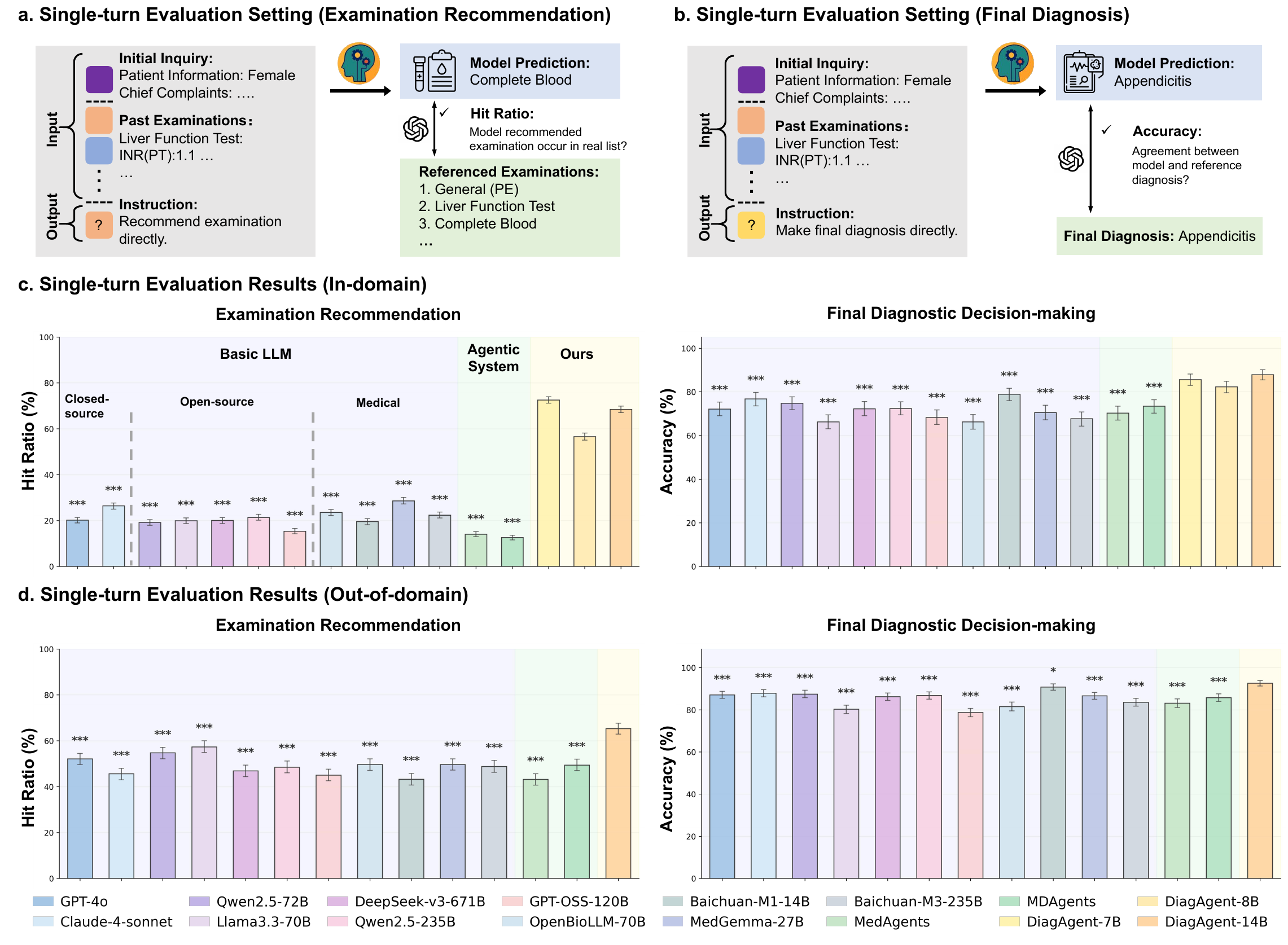}
    \caption{Overview of single-turn evaluation settings and results. \textbf{a} shows the single-turn evaluation setting for examination recommendation measured with the hit ratio. \textbf{b} illustrates the single-turn evaluation setting for final diagnosis measured with the accuracy. \textbf{c} benchmarks our DiagAgent variants against 11 leading LLMs and 2 agentic systems on the MIMIC-IV dataset. The detailed numbers can be found in Supplementary Table~\ref{tab:diagnoser_result_under_static}. 
    \textbf{d} evaluates the model's performance on out-of-domain datasets sourced from PMC-OA, MTsamples, and DDXPlus. The detailed number can be found in Supplementary Table~\ref{tab:diagnoser_result_under_static} and~\ref{tab:diagnoser_result_combined_single_turn}.
    Error bars show 95\% confidence intervals.     
    Significance levels relative to the DiagAgent-14B are marked as: $^{*}p < 0.05$, $^{**}p < 0.01$, and $^{***}p < 0.001$. }
    \label{fig:static_results}
\end{figure}

In this section, we analyze the performance of \textbf{DiagAgent} under single-turn evaluation. The main results are shown in Figure~\ref{fig:static_results}. 

\change{\textbf{In-Domain Analysis}} 

Figure~\ref{fig:static_results}c summarize the performance of all models under single-turn evaluation on the MIMIC-IV test set. Models are grouped into three categories: basic LLMs, agentic systems, and our DiagAgent variants.


\change{{DiagAgent} variants deliver substantial gains across all metrics. 
The best-performing model, DiagAgent-14B, achieves a Hit Ratio of 68.49\% and diagnosis accuracy of 87.87\%. 
This performance establishes a massive margin over baselines: it improves examination recommendation by over 40\% compared to the strongest medical LLM (MedGemma) and exceeds general-purpose models like DeepSeek-v3 by nearly 48\%. 
These results confirm that post-training within DiagGym fundamentally boosts clinical reasoning capabilities beyond current state-of-the-art~(SOTA) levels.}


\change{Among \textbf{Basic LLMs}, close-sourced models (\emph{e.g.}, Claude-4-sonnet, GPT-4o) generally outperform open-source counterparts but still struggle with examination recommendation, remaining below 30\%. 
Notably, scaling parameters, as seen in DeepSeek-v3 (671B), does not yield proportional gains in medical recommendation without targeted adaptation. 
While medical-specialized models, like MedGemma, Baichuan-M1, Baichuan-M3 show slight advantages over general LLMs, they still fall significantly short of the DiagAgent series.}


\change{Regarding \textbf{Agentic Systems}, frameworks like MedAgents and MDAgents~(based on DeepSeek-V3) fail to deliver meaningful improvements over their base models. 
This suggests that without a well-aligned foundation, multi-agent coordination alone cannot overcome the complexity of interactive diagnostic reasoning and may even exacerbate hallucination-driven errors.}


\change{\textbf{Out-of-Domain Analysis}}

\change{As shown in Figure~\ref{fig:static_results}d, DiagAgent-14B maintains robust leadership in out-of-domain settings, achieving a Hit Ratio of 65.30\% and Diagnosis Accuracy of 92.57\%. 
This significantly outperforms both the strongest open-source baseline (Llama3.3-70B) and proprietary SOTA models (GPT-4o). 
Crucially, while some baselines (\emph{e.g.}, MedGemma, Baichuan-M1) demonstrate reasonable adaptability in diagnosis accuracy, DiagAgent is the only model to simultaneously achieve high precision in {both} examination retrieval and diagnostic decision-making. 
This confirms that the policy learned from DiagGym captures generalized medical logic rather than merely overfitting to MIMIC-IV distributions.}

\subsubsection{End-to-end Evaluation Analysis on Automatic Metrics} 

\begin{figure}[!t]
    \centering
    \includegraphics[width=\linewidth]{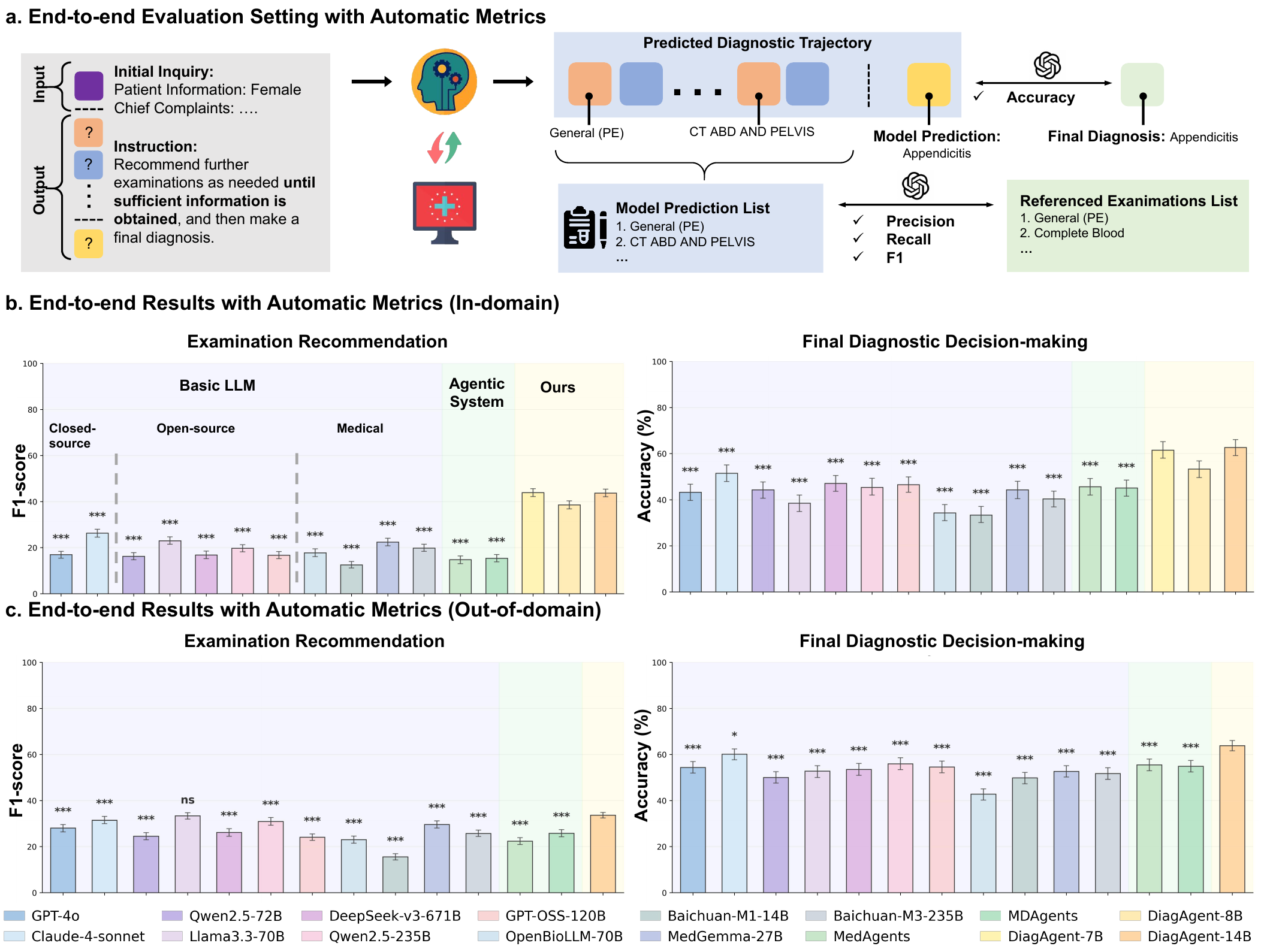}
    \caption{Overview of end-to-end evaluation. In this setting, diagnostic agents are evaluated through end-to-end finishing the entire diagnostic trajectory by interaction with the external diagnostics world model. \textbf{a} illustrates the end-to-end evaluation pipeline with automatic metrics to assess examination recommendation efficacy and diagnostic accuracy. \textbf{b} benchmarks our DiagAgent with 11 LLMs and 2 more agentic systems under end-to-end evaluation settings with automatic metrics on the MIMIC-IV dataset. \textbf{c} evaluates the model's e performance on out-of-domain datasets sourced from PMC-OA, MTsamples, and DDXPlus. The detailed number can be found in Supplementary Table~\ref{tab:diagnoser_result_under_ehrgenerator_env},~\ref{tab:diagnoser_result_combined}. Error bars show 95\% confidence intervals. Significance levels relative to the DiagAgent-14B are marked as: $^{*}p < 0.05$, $^{**}p < 0.01$, and $^{***}p < 0.001$. }
    \label{fig:dynamic_results}
\end{figure}

In this section, we analyze the performance of \textbf{DiagAgent} under end-to-end evaluation on automatic metrics. The main results are shown in Figure~\ref{fig:dynamic_results}.

\change{\textbf{In-Domain Analysis}}

\change{As shown in Figure~\ref{fig:dynamic_results}b, DiagAgent models achieve the highest scores across all evaluation metrics. 
Crucially, additionally reported in Supplementary Table~\ref{tab:diagnoser_result_under_ehrgenerator_env}, DiagAgent-14B engages in substantially longer diagnostic dialogues (average 6.66 turns) compared to standard LLMs like DeepSeek-v3 ($\approx$2.5 turns). 
This sustained interaction facilitates comprehensive evidence gathering, yielding a recall of 52.14\%, which is more than four times that of the strongest baselines, without compromising precision. 
Consequently, DiagAgent-14B achieves a dominant diagnostic accuracy of 62.63\%, outperforming the nearest competitor by over 10\%. 
These findings confirm that the active exploration capability acquired through DiagGym directly translates into more informed and accurate decision-making.}


\change{Among \textbf{Basic LLMs}, advanced general models (\emph{e.g.}, Claude-4-sonnet, DeepSeek-V3) tend to outperform medical LLMs, like MedGemma and Baichuan series. 
Despite their domain knowledge, medical LLMs struggle in this dynamic setting, indicating that static medical knowledge does not automatically confer the ability to actively query and integrate evidence across multi-turn interactions.}


\change{Regarding \textbf{Agentic Systems}, frameworks like MedAgents and MDAgents~(based on DeepSeek-V3) fail to deliver meaningful improvements over their base models. 
Comparison of turn lengths reveals that these multi-agent systems suffer from \textit{premature closure}, averaging fewer than 2.5 turns and terminating inquiry before gathering sufficient information. 
This results in low recall and stagnation in diagnostic accuracy, further validating that ``expert discussion'' prompts cannot substitute for a well-aligned underlying policy.}

\change{\textbf{Out-of-Domain Analysis}} 

\change{We further validate robustness on the out-of-domain data in DiagBench (Figure~\ref{fig:dynamic_results}c). 
DiagAgent-14B retains its leadership, achieving the highest F1-score~({33.63\%}) and Diagnosis Accuracy~(63.84\%) among all 13 evaluated systems, significantly surpassing both proprietary SOTA models (Claude-4-Sonnet) and open-source giants (Qwen3-235B). 
Consistent with in-domain behavior, DiagAgent maintains a longer average turn duration~(6.88 turns) compared to baselines, as detailed in Supplementary Table~\ref{tab:diagnoser_result_combined}.
This confirms that the model's active inquiry strategy is a fundamental, generalized behavioral trait rather than an artifact overfitting the MIMIC-IV dataset.}

In summary, these results highlight the importance of post‑training LLMs within clinically realistic, interactive environments. By endowing models with the capacity to determine the timing and content of their multi‑turn actions, DiagGym substantially improves both the quality of examination recommendations and the accuracy of final diagnoses, narrowing the gap towards deployable, decision‑capable clinical AI systems.

\begin{figure}[!t]
    \centering
    \includegraphics[width=\linewidth]{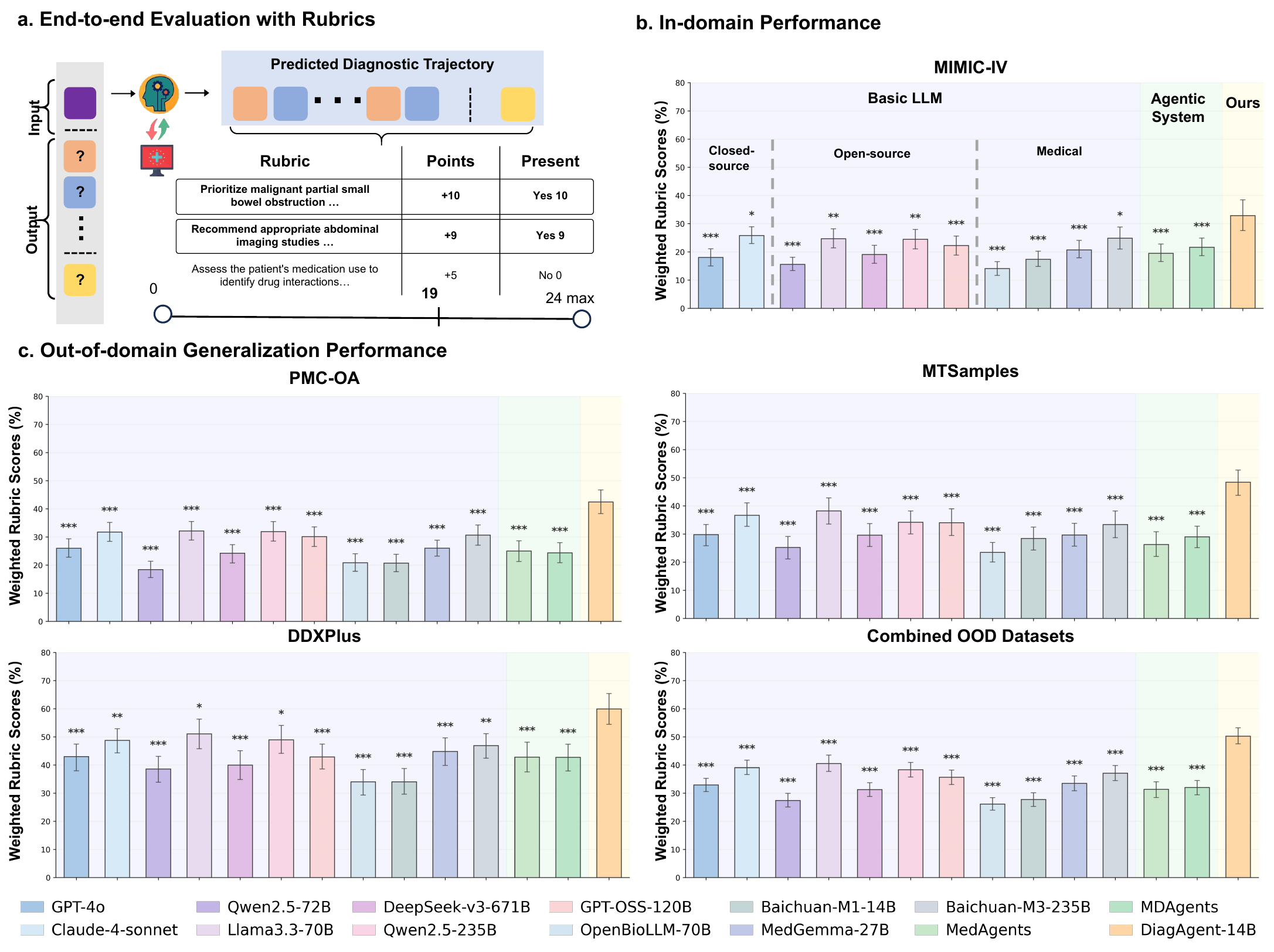}
    \caption{Overview of end-to-end evaluation settings and results. In this setting, diagnostic agents are evaluated through end-to-end finishing the entire diagnostic trajectory by interaction with the external diagnostics world model. \textbf{a} illustrates our end-to-end evaluation pipeline with rubric-based metrics. A judge model evaluates the full diagnostic trajectory of a diagnostic model against physician-curated rubrics, which specify criteria, clinical importance weights (Points), and whether the criterion was satisfied (Present). \textbf{b} compares the aggregate weighted proportion of satisfied rubrics(\%) across different LLMs in different model sizes on the MIMIC-IV dataset. \textbf{c} evaluates the model's e performance on three out-of-domain~(OOD) sources separately: PMC-OA, MTsamples, and DDXPlus. Error bars show 95\% confidence intervals. Significance levels relative to the DiagAgent-14B are marked as: $^{*}p < 0.05$, $^{**}p < 0.01$, and $^{***}p < 0.001$.}
    \label{fig:results_rubrics}
\end{figure}

\subsubsection{End-to-end Evaluation Analysis on Rubric-based Metrics} 

In this section, we analyze the performance of \textbf{DiagAgent} under end-to-end evaluation on rubric-based metrics. The main results are shown in Figure~\ref{fig:results_rubrics}. 


\textbf{In-Domain Analysis}

\change{Figure~\ref{fig:results_rubrics}b indicates that our dynamic training paradigm significantly enhances alignment with high-importance clinical protocols. 
DiagAgent-14B achieves a weighted rubric score of 32.86\%, outperforming the strongest basic LLM (Qwen3-235B) by over 8\% and the best agentic baseline (MDAgent) by over 10\%. 
These results suggest that DiagAgent does not merely guess the final diagnosis but demonstrates superior procedural interactive reasoning, satisfying rigorously weighted criteria for history taking and examination selection.}



\change{Among baselines, \textbf{general LLMs} (\emph{e.g.}, Claude-4-sonnet, Qwen3-235B) consistently outscore smaller medical-specialized models. While medical models may arrive at correct diagnoses, they frequently fail to satisfy process-focused rubrics due to weaker reasoning capabilities, suggesting that prior clinical LLM adaptation methods have neglected the enhancement of multi-turn interactive capabilities. \textbf{Agentic systems} like MDAgent provide only marginal gains ($\approx$2.5\%) over their base models. This indicates that inference-time discussion alone is insufficient to ensure procedural quality without fundamental policy alignment.}

\change{\textbf{Out-of-Domain Analysis}}

\change{We further assess alignment with clinical standards on the out-of-domain data in the benchmark (Figure~\ref{fig:results_rubrics}c). 
DiagAgent-14B establishes a decisive lead with an aggregate weighted score of \textbf{50.27\%}, creating a substantial gap of over 10\% compared to both the strongest proprietary (Claude-4-sonnet) and open-source (Qwen3-235B) baselines. 
This superiority is consistent across all individual data sources (PMC-OA, MTSamples, DDXPlus). DiagAgent outperforms second competitors by wide margins, \emph{i.e.}, surpassing Claude-4-Sonnet by 10.65\% on PMC-OA, exceeding GPT-4o by 18.62\% on MTSamples, and outperforming DeepSeek-V3 by 19.98\% on DDXPlus. 
Statistical hypothesis testing confirms that DiagAgent's improvements are significant ($p < 0.05$) against all baselines across these diverse clinical scenarios.}

Overall, the results on DiagBench reinforce that dynamic training focused on intermediate reasoning processes enables significantly improved satisfaction of high-value clinical procedures. DiagAgent consistently surpasses both large generic LLMs and current agentic frameworks, underscoring the essential role of fine-grained, process-aware training for safe and effective clinical decision support.


\subsection{Ablation Study}
We conducted ablation experiments under the end-to-end evaluation setting to assess three aspects of the proposed framework: 
(i) whether the reinforcement learning in virtual environment outperforms supervised fine-tuning (SFT) at the same model scale; 
(ii) the impact of reward design, comparing diagnosis-only rewards with dual rewards incorporating both diagnosis accuracy and examination-recommendation quality; and (iii) the generality of \textbf{DiagAgent} across different model sizes and families.

\noindent \textbf{Experimental Design}

As outlined in Table~\ref{tab:ablation_study}, for each base model, we first establish a zero-shot baseline in which the LLM answers without fine-tuning. We then apply full supervised finetuning, in which all cases are converted into multi-turn diagnostic dialogues for supervised training~(Supplementary Section~\ref{sec:supervised_finetuning}), bypassing the simulator and reinforcement learning pipeline. 
Finally, we test three DiagAgent configurations:
(i) \textbf{cold-start only}: supervised fine-tuning on a small subset to learn output format;
(ii) \textbf{cold-start + RL with diagnosis reward}: RL optimising diagnosis accuracy only;
(iii) \textbf{full DiagAgent}: diagnose reward plus examination-recommendation reward.

\noindent \textbf{Supervised Finetuning~(SFT) vs.~Reinforcement Learning~(RL)}

As shown in Table~\ref{tab:ablation_study}, 
zero-shot baselines perform poorly (diagnosis accuracy: 16.93\% for Qwen2.5-7B; 34.93\% for Qwen2.5-14B), highlighting the difficulty of managing interactive trajectories without domain adaptation. Full SFT improves diagnosis accuracy (45.33\%, 47.07\%, and 47.60\% for Qwen2.5-7B, Llama3.1-8B, and Qwen2.5-14B, respectively) and examination recommendation quality, but the gains are limited by static trajectory data extracted from MIMIC-IV discharge notes, which do not reflect the dynamic branching in interactive consultations. 
\change{In contrast, our RL-trained DiagAgent achieves a universal performance gain across all three base models, outperforming the SFT-only counterparts.} Specifically, accuracy surges to 61.47\% (Qwen2.5-7B), 53.33\% (Llama3.1-8B), and  62.67\% (Qwen2.5-14B). Furthermore, DiagAgent reduces both the average dialogue turns and token length, proving that the transition from static imitation~(SFT) to dynamic feedback learning~(RL) fosters more efficient and decisive diagnostic reasoning.

\change{To ensure the SFT baseline was not limited by insufficient compute, we analyzed the impact of training duration in Supplementary Section~\ref{sec:ablation_on_flops}. We found that extending SFT to 10 epochs resulted in a performance plateau after epoch 3. This confirms that our gains are attributable to the superior RL learning paradigm, not merely additional training steps.}

\noindent \textbf{Effect of Reward Design}

To demonstrate the effectiveness of reward design, we first conduct a cold start phase, where each model is supervised-finetuned on a subset of the training set to learn the output format. Adding RL with only the Diagnose Reward yields large gains over SFT, for example, Qwen2.5-7B improves from 44.40\% (Full SFT) to 59.41\%, Qwen2.5-14B from 45.98\% to 59.73\%, and Llama3.1-8B from 45.35\% to 54.12\%. However, but F1 scores for examination recommendation remain low ($\leq$ 36\%).
Introducing the Examination Recommendation Reward markedly improves F1 across all models~(Qwen2.5-7B: 32.76\% $\rightarrow$ 46.86\%; Qwen2.5-14B: 35.95\% $\rightarrow$ 47.89\%), with slight additional gains in accuracy. This confirms the importance of dual-reward shaping for balancing precision in diagnosis and quality in examination planning.

\noindent \textbf{Model Size and Family}

Reinforcement learning with virtual environment benefits all LLMs training,  for example, in diagnosis accuracy, from 16.38\% to 60.78\% for Qwen2.5-7B,  25.16\% to 53.85\% for Llama3.1-8B, and 33.83\% to 61.63\% for Qwen2.5-14B, proving the effectiveness of our method. Larger or intrinsically stronger base models achieve higher post-training ceilings: Qwen2.5-14B delivers the best overall performance (61.63\% accuracy, 47.89\% F1), followed by Qwen2.5-7B (60.78\%, 46.86\%), while Llama3.1-8B lags (53.85\%, 43.02\%). This suggests that while DiagAgent’s exploration-driven optimisation is broadly applicable, the quality of the base model constrains the attainable performance upper bound.

\begin{table}[!t]
\renewcommand{\arraystretch}{1.3} 
\footnotesize
\centering
\caption{Results of ablation study. We report the Avg. Turns, precision, recall and F1-score for examination recommendations and accuracy for diagnosis. All metrics are reported with \textbf{95\% Confidence Intervals}}
\label{tab:ablation_study}
\resizebox{\textwidth}{!}{
\begin{tabular}{c|cccc|c|ccc|c}
\toprule
\multirow{2}{*}{Method} & \multicolumn{2}{c|}{Instruction Tuning} & \multicolumn{2}{c|}{Reinforcement Learning} & \multirow{2}{*}{Avg. Turns} & \multicolumn{3}{c|}{Examination Recommendation} & Diagnosis \\ \cline{2-3} \cline{4-5} \cline{7-9}
 & Full SFT & Cold Start & \makecell{Recommend \\ Reward} & \makecell{Diagnosis \\ Reward} & & Precision & Recall & F1 & Accuracy(\%) \\ 
\midrule
\rowcolor{mygray} \multicolumn{10}{c}{Qwen2.5-7B} \\ 
\midrule
Baseline & \ding{55} & \ding{55} & \ding{55} & \ding{55} & 1.97 & 
\makecell{17.76 \\ \scriptsize[15.54-20.37]} & 
\makecell{8.62 \\ \scriptsize[7.37-9.94]} & 
\makecell{10.09 \\ \scriptsize[8.76-11.58]} & 
\makecell{16.93 \\ \scriptsize[14.40-19.60]} \\ 

Full SFT & \ding{52} & \ding{55} & \ding{55} & \ding{55} & 7.98 & 
\makecell{37.51 \\ \scriptsize[35.59-39.50]} & 
\makecell{50.99 \\ \scriptsize[49.02-52.86]} & 
\makecell{39.70 \\ \scriptsize[38.03-41.29]} & 
\makecell{45.33 \\ \scriptsize[41.87-49.07]} \\ 
\midrule
\multirow{3}{*}{DiagGym} & \ding{55} & \ding{52} & \ding{55} & \ding{55} & 8.36 & 
\makecell{33.03 \\ \scriptsize[31.26-34.85]} & 
\makecell{47.16 \\ \scriptsize[45.18-49.25]} & 
\makecell{35.35 \\ \scriptsize[33.73-36.89]} & 
\makecell{36.13 \\ \scriptsize[32.66-39.73]} \\ 

 & \ding{55} & \ding{52} & \ding{55} & \ding{52} & 4.46 & 
\makecell{37.07 \\ \scriptsize[35.11-39.03]} & 
\makecell{33.64 \\ \scriptsize[31.78-35.28]} & 
\makecell{32.75 \\ \scriptsize[31.00-34.37]} & 
\makecell{60.40 \\ \scriptsize[56.93-64.00]} \\ 

 & \ding{55} & \ding{52} & \ding{52} & \ding{52} & 5.45 & 
\makecell{\textbf{46.02} \\ \scriptsize[44.04-47.82]} & 
\makecell{47.33 \\ \scriptsize[45.45-49.27]} & 
\makecell{\textbf{43.90} \\ \scriptsize[42.29-45.53]} & 
\makecell{\textbf{61.47} \\ \scriptsize[58.27-64.67]} \\ 
\midrule
\rowcolor{mygray} \multicolumn{10}{c}{Llama3.1-8B} \\ 
\midrule
Baseline & \ding{55} & \ding{55} & \ding{55} & \ding{55} & 4.36 & 
\makecell{23.49 \\ \scriptsize[21.59-25.31]} & 
\makecell{19.15 \\ \scriptsize[17.52-20.79]} & 
\makecell{18.73 \\ \scriptsize[17.17-20.25]} & 
\makecell{25.60 \\ \scriptsize[22.67-28.67]} \\ 

Full SFT & \ding{52} & \ding{55} & \ding{55} & \ding{55} & 7.89 & 
\makecell{36.90 \\ \scriptsize[35.06-38.89]} & 
\makecell{\textbf{49.66} \\ \scriptsize[47.61-51.68]} & 
\makecell{\textbf{38.63} \\ \scriptsize[36.99-40.28]} & 
\makecell{47.07 \\ \scriptsize[43.33-50.67]} \\ 
\midrule
\multirow{3}{*}{DiagGym} & \ding{55} & \ding{52} & \ding{55} & \ding{55} & 9.03 & 
\makecell{31.80 \\ \scriptsize[30.00-33.73]} & 
\makecell{48.06 \\ \scriptsize[45.93-49.96]} & 
\makecell{34.37 \\ \scriptsize[32.86-36.02]} & 
\makecell{32.00 \\ \scriptsize[28.67-35.33]} \\ 

 & \ding{55} & \ding{52} & \ding{55} & \ding{52} & 5.15 & 
\makecell{36.91 \\ \scriptsize[35.13-38.86]} & 
\makecell{37.47 \\ \scriptsize[35.50-39.39]} & 
\makecell{34.82 \\ \scriptsize[33.07-36.52]} & 
\makecell{52.53 \\ \scriptsize[48.93-56.13]} \\ 

 & \ding{55} & \ding{52} & \ding{52} & \ding{52} & 5.73 & 
\makecell{\textbf{39.57} \\ \scriptsize[37.74-41.41]} & 
\makecell{43.13 \\ \scriptsize[41.29-45.12]} & 
\makecell{38.56 \\ \scriptsize[36.89-40.18]} & 
\makecell{\textbf{53.33} \\ \scriptsize[49.73-56.80]} \\ 
\midrule
\rowcolor{mygray} \multicolumn{10}{c}{Qwen2.5-14B} \\ 
\midrule
Baseline & \ding{55} & \ding{55} & \ding{55} & \ding{55} & 3.61 & 
\makecell{33.98 \\ \scriptsize[31.02-36.96]} & 
\makecell{17.14 \\ \scriptsize[15.61-18.61]} & 
\makecell{19.70 \\ \scriptsize[18.06-21.24]} & 
\makecell{34.93 \\ \scriptsize[31.47-38.53]} \\ 

Full SFT & \ding{52} & \ding{55} & \ding{55} & \ding{55} & 7.63 & 
\makecell{38.01 \\ \scriptsize[36.03-39.84]} & 
\makecell{50.52 \\ \scriptsize[48.52-52.55]} & 
\makecell{39.74 \\ \scriptsize[38.15-41.38]} & 
\makecell{47.60 \\ \scriptsize[43.87-51.20]} \\ 
\midrule
\multirow{3}{*}{DiagGym} & \ding{55} & \ding{52} & \ding{55} & \ding{55} & 8.56 & 
\makecell{32.99 \\ \scriptsize[31.21-34.83]} & 
\makecell{48.49 \\ \scriptsize[46.36-50.40]} & 
\makecell{35.62 \\ \scriptsize[34.01-37.25]} & 
\makecell{35.47 \\ \scriptsize[32.27-38.67]} \\ 

 & \ding{55} & \ding{52} & \ding{55} & \ding{52} & 5.51 & 
\makecell{37.90 \\ \scriptsize[35.97-39.78]} & 
\makecell{37.72 \\ \scriptsize[35.92-39.68]} & 
\makecell{35.43 \\ \scriptsize[33.71-37.12]} & 
\makecell{58.27 \\ \scriptsize[54.67-61.73]} \\ 

 & \ding{55} & \ding{52} & \ding{52} & \ding{52} & 6.66 & 
\makecell{\textbf{42.04} \\ \scriptsize[40.13-43.89]} & 
\makecell{\textbf{52.14} \\ \scriptsize[50.20-54.09]} & 
\makecell{\textbf{43.72} \\ \scriptsize[42.09-45.56]} & 
\makecell{\textbf{62.67} \\ \scriptsize[59.20-66.00]} \\ 
\bottomrule
\end{tabular}}
\end{table}

\subsection{Case Study}

The following case study illustrates the qualitative performance of DiagAgent, offering insights into its ability to navigate reasoning and maintain high fidelity in complex clinical scenarios.





\textbf{A Case of Interactive Diagnostic Trajectory}

As shown in Supplementary Figure~\ref{fig:case_study_diagnoser}, we present a case study illustrating \textbf{DiagAgent-14B's} dynamic interactive diagnostic trajectory within the DiagGym environment, modeling a typical appendicitis work-up. Each case record includes the initial inquiry, final diagnosis, interactive exchanges between diagnostic agent and simulator, and a reference ground-truth diagnostic timeline drawn from clinical records.

\textit{First}, the model’s decisions follow standard reasoning. Upon receiving initial symptoms—abdominal pain migrating to the right lower quadrant, nausea, diarrhoea, and anorexia—the diagnostic agent prioritises appendicitis in the differential and orders a complete blood count (CBC). When the CBC reveals an abnormally high neutrophil count, further pointing toward infection or inflammation, the diagnostic agent appropriately requests a CT scan of the abdomen and pelvis with contrast, which display a dilated, fluid-filled appendix with periappendiceal fat stranding, confirming acute appendicitis. 
Throughout the process, each diagnostic step and rationale aligns closely with the reference timeline, demonstrating reliable differential diagnosis.

\textit{Second}, the dynamic environment's responses remain consistent with the patient's case summary and expected clinical progression. For instance, when the CBC is requested, it provides results consistent with an acute inflammatory, including elevated white blood cell and neutrophil counts; when the CT is ordered, it returns hallmark imaging features of appendicitis. This realistic feedback ensures that the agent’s decision-making unfolds in a manner faithful to real clinical workflows.

\textbf{Cases of Rubric-based Evaluation}

To further scrutinize the procedural quality of DiagAgent's intermediate diagnostic steps, we employed an evaluation based on physician-curated rubrics that assess the integrity of multi-turn clinical interactions.

A typical \textbf{successful case} (left lower extremity infection) is presented in Supplementary Figure~\ref{fig:rubric_success_mode}, illustrating the agent's robust procedural performance. The agent exhibits high coherence and dynamic strategy adjustment: it orders a CBC and, upon receiving a result showing an elevated but non-critical neutrophil count, appropriately requests a wound culture to identify the causative organism. Following the positive culture for Staphylococcus aureus, the agent orders a blood culture to rule out bacteremia and then efficiently terminates further investigation after receiving a negative result, avoiding over-testing. The high scores on the procedural rubrics confirm that the agent's decision-making process is both efficient and clinically sound, successfully meeting criteria for prioritizing tests, interpreting results, and achieving evidence-driven closure.

We also provide a illustrative \textbf{failure case} (ruptured ectopic pregnancy with hemodynamic instability) in Supplementary Figure~\ref{fig:rubric_fail_mode} to showcase model's fail mode. While the agent's diagnostic reasoning is highly effective—it correctly orders hCG and subsequent pelvic ultrasound based on the patient's unstable presentation, rapidly confirming the diagnosis—the evaluation reveals critical omissions in immediate emergency care. Specifically, the agent fails to satisfy the highest-weighted rubrics concerning emergency resuscitation and surgical team notification. It is important to emphasize that DiagAgent is primarily designed as a diagnostic reasoning model. Its core capability of accurate differential diagnosis and sequential information gathering remains intact, validating its utility for diagnostic quality enhancement despite the observed gap in acute therapeutic and stabilization management, which falls outside its initial scope of contribution.



%% file: content_npj/03_Discussion.tex
\section{Discussion}

Large language models (LLMs) have achieved notable success across a range of clinical tasks~\cite{singhal2023towards, wu2024pmc, qiu2024towards, singhal2025toward, qiu2025quantifying, sandmann2025benchmark, renc2025foundation}, yet they remain fundamentally limited in dynamic, multi-turn decision‑making. Even state‑of‑the‑art systems often struggle with real‑time interactive diagnostic reasoning, deciding which examinations to order, when to order them, and how to coordinate an efficient, end‑to‑end diagnostic process~\cite{qiu2025quantifying, hager2024evaluation, johri2025evaluation}. Unlike human physicians, who adaptively update decisions as new information emerges, current LLMs frequently fail to manage trajectories effectively under uncertainty.



\change{To address this gap, recent works enhance instruction‑tuning data by utilizing synthetic dialogues or extracting data from patient records~\cite{mcduff2025towards,tu2025towards}. Frameworks like DoctorAgent-RL~\cite{dockeragentrl} and MedAgentSim~\cite{medagentslim} further employ RL and role-play strategy to optimize interactive diagnosis. 
However, they primarily target online consultation scenarios where patient information remains static. In real-world practice, the physician's role extends to recommending examinations, facing a long-term dynamic environment with evolving patient conditions. We bridge this by training agents within a virtual environment, fostering dynamic ability to order test.} 





\textbf{Main Contribution}

\textbf{A diagnostics world model serving as a virtual clinical environment for end-to-end agentic RL.} At the core of our method is the world model, DiagGym, a fine‑tuned LLM that goes beyond replaying historical records, and generates dynamic results for any requested examination. Unlike prior role‑playing simulators such as AgentClinic~\cite{agentclinic}, AgentHospital~\cite{agenthospital}, and SDBench~\cite{microsoftSDBench}, which are constrained by static, pre‑collected data, DiagGym can generate novel, clinically plausible trajectories beyond the original patient records, serving as a dynamic virtual clinical environment for clinical agent evolving.
 
\textbf{First end-to-end RL platform for interactive diagnostic agents.} This paper provides an end-to-end, multi‑turn RL framework in which agents interact iteratively with the clinical environment until a final diagnosis is reached. Agents actively explore diverse diagnostic trajectories, optimizing for both diagnostic accuracy and efficiency in examination recommendation. This allows them to handle complex, uncommon, or evolving patient scenarios that static supervised approaches cannot cover. We believe our approach provides the community with a robust resource for developing, testing, and comparing interactive diagnostic agents in a dynamic long-term patient management manner, moving beyond the more commonly considered yet less clinically challenging static consultation scenarios for clinical LLMs.

\textbf{Interactive and exploratory dynamic diagnostic reasoning.} 
Our final agent, DiagAgent, achieves substantial improvements over all evaluated LLMs and agentic systems. \change{Reinforcement learning consistently improves LLMs' competence in planning and managing interactive diagnostic trajectories, addressing the shortcomings of static fine-tuning. This dynamic, end-to-end training paradigm cultivates robust interactive competencies that are essential for real-world clinical deployment.}

\change{\textbf{DiagBench, a comprehensive multi-center benchmark focusing on interactive trajectories with fine-grained physician-written rubrics.} Breaking away from single-center limitations, DiagBench integrates data from \textbf{four distinct clinical sources}: MIMIC-IV, PMC-OA case reports, MTSamples, and DDXPlus. The benchmark comprises a total of \textbf{2.2K physician-validated cases}. Furthermore, to enable granular, process-oriented assessment, 399 sampled cases are annotated with \textbf{3.3K physician-written rubrics} that assign weighted points to critical steps in the reasoning process, offering deeper insights into how a diagnosis is progressively reached beyond final diagnosis accuracy.}

\textbf{Key Findings}

\textbf{Better alignment with dynamic decision‑making.} Reinforcement learning consistently improves LLMs’ competence in planning and managing interactive diagnostic trajectories, addressing shortcomings of static fine‑tuning. This dynamic, end-to-end training paradigm cultivates robust interactive competencies that are essential for real-world clinical deployment.

\textbf{Superior to supervised fine-tuning (SFT).} Across scales (7B-14B) and families (Qwen2.5, Llama3.1), diagnostic agent trained with reinforcement learning outperforms SFT by significant margins.
This superiority is evident in \change{interactive diagnosis tasks}, where DiagAgent's self-exploration fosters adaptability to evolving patient scenarios, leading to higher accuracy, efficiency, and robustness compared to SFT's limitations in handling incomplete or atypical information.

\textbf{Dependence on base model quality.} The intrinsic capability of the foundation model strongly shapes DiagGym’s upper performance bound. While DiagGym delivers robust gains even for moderate‑scale (7B–14B) models, continuing to scale up to larger foundation models could unlock even greater advancements, suggesting a promising path for enhancing diagnostic agents.

\textbf{Limitations and Future Work}

\change{First, we acknowledge that DiagGym holds potential for further optimization to achieve higher fidelity. Currently, the system relies primarily on MIMIC-IV due to the scarcity of high-quality longitudinal records, which introduces potential biases specific to ED and ICU settings. Additionally, DiagGym operates as a text-based probabilistic model rather than a clearly verified disease mechanism simulator. While our extensive out-of-domain evaluation demonstrates that DiagGym, as a simulated RL environment, already enables the trained DiagAgent to achieve generalized, state-of-the-art diagnostic capabilities, future iterations could be significantly enhanced by incorporating cross-center data and and clinical knowledge priors, thereby facilitating the training of more robust interactive diagnostic agents.}


Secondly, the models evaluated in this work are relatively modest in scale (up to 14B parameters), which may constrain the framework's full potential. Larger foundation models, such as DeepSeek-v3, GPT-OSS-120B, could yield qualitative leaps in performance by enhancing inherent reasoning capabilities and exploratory depth. With sufficient model capacity, both DiagGym and DiagAgent could be further enhanced, culminating in higher-fidelity clinical environment simulations and superior interactive diagnostic capabilities.

Thirdly, we note that DiagAgent's absolute scores on the rubric-based benchmark is modest. This is because the physician-authored rubrics reflect real-world clinical practice by awarding points for immediate treatment and patient management actions (\emph{e.g.}, ``prepare for emergency transfer''). As our framework deliberately focuses on the diagnostic task, our agent was not trained to perform these out-of-scope therapeutic interventions. Extending the agent’s capabilities to integrate long-term management with timely treatment planning, interleaved with the diagnostic procedure, represents a clear direction for aligning clinical LLMs with practical usage demands in future work.

\change{Finally, a key advantage of RL is its ability to learn without direct process supervision, enabling the emergence of strategies that may transcend current human capabilities~\cite{alphago,deepseekv3}. This is particularly critical in healthcare, where interactive diagnosis remains a complex, subjective challenge. By employing a diagnostics world model to facilitate RL training, our framework allows agents to explore optimal diagnostic pathways from extensive data, thereby offering the potential to refine physician decision-making trajectories. While constraints on training scale and human evaluation resources prevented a demonstration of this capability in the current study, we highlight this as a promising direction for future work to further validate.}




%% file: content_npj/04_Methodology.tex
\section{Methods}
In this section, we provide additional details on the training process of DiagGym and DiagAgent, as well as the baseline methods used for comparison.

\subsection{DiagGym Training}
\label{sec:simulator_data_construction}

In this section, we present the detailed training of DiagGym.

\begin{figure}[!t]
    \centering
    \includegraphics[width=\linewidth]{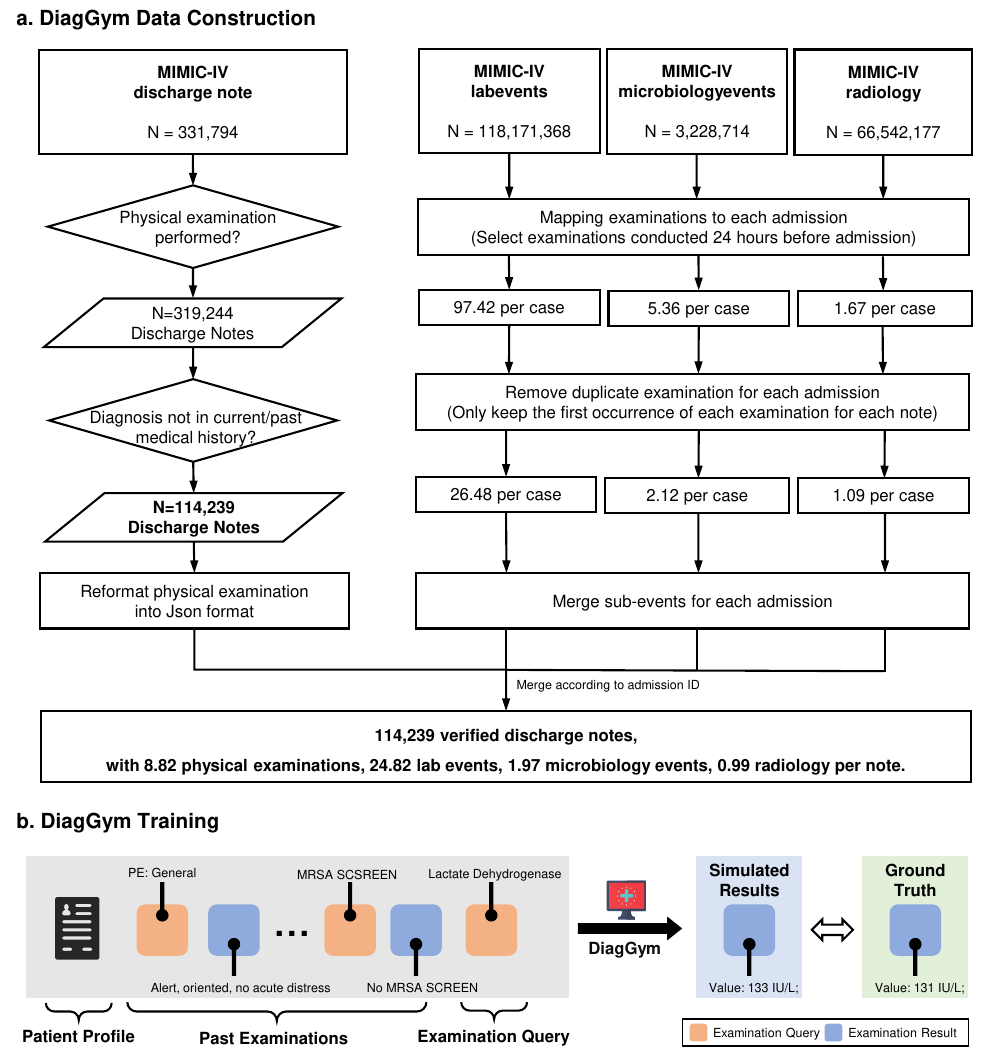}
    \caption{Overview of DiagGym data construction and training pipeline. \textbf{a} shows the process for constructing the DiagGym training dataset. \textbf{b} illustrates the pipeline for DiagGym training.}
    \label{fig:method_simulator}
\end{figure}

\textbf{Data Construction}

As shown in Figure~\ref{fig:method_simulator}a, to train DiagGym, we constructed a dataset of patient EHRs derived from MIMIC-IV. Each patient’s EHR was reorganized into two components: 
(i) patient profile, (ii) time-ordered examination set. The pipeline was based on MIMIC-CDM~\cite{hager2024evaluation}, but extended to cover a broader range of diseases.

We first process the MIMIC-IV discharge notes. Leveraging their structured format, we applied heuristic string matching to extract the patient profile. Specifically, a patient profile was composed of content under the headings `physical examination', `chief complaint', `current medical history', `past medical history', `social history', `family history', and, most critically, the final diagnosis listed under the `discharge diagnosis' heading.

Next, we apply a two-step filtering process: 
(i) cases without physical examination records are excluded; 
(ii) DeepSeek-V3, instructed with \texttt{prompt~\ref{prompt:check_if_diagnosis_in_past_medical_history}}, 
is used to remove cases where the discharge diagnosis appeared in either the past medical history or the current medical history. Such cases often involve transfers with established diagnoses and typically lack diagnostically relevant examinations.

We then construct a time-ordered examination set for each patient, with each examination in the set comprising the queried examination item and its corresponding results. 
First, the previously extracted physical examination text from discharge notes is reformatted into a structured tabular format using DeepSeek-v3 with \texttt{prompt~\ref{prompt:reformat_physical_exam_into_json}}, ensuring consistency with other MIMIC-IV examination records. 
Following MIMIC-CDM, this physical examination is designated as the initial test in the examination set. 
Next, we append laboratory results, microbiological examinations, and radiological records conducted within one day prior to admission. The one-day time interval is selected because examinations performed earlier generally have limited diagnostic relevance. Laboratory data is obtained from \texttt{labevents.csv}, microbiology data from \texttt{microbiologyevents.csv}, and radiology from \texttt{radiology.csv}.

For laboratory and microbiology data, we use the original structured records but standardized examination item names using the MIMIC-CDM mapping table to group the linked items~(e.g., ``red blood cell count'' under the broader ``complete blood count''). Radiology entries are supplemented with missing names extracted by string-matching from the `EXAMINATION' section of reports. All examination entries in MIMIC-IV contain timestamps, enabling accurate chronological ordering. \change{For repeated pre-admission examinations, we follow~\cite{hager2024evaluation} and prioritize the earliest record to preserve the original clinical state at the time of first assessment, thereby avoiding the influence of subsequent interventions.}

Finally, we split the restructured EHR dataset into training and testing sets. The resulting dataset consists of 118,478 patient EHRs, with each case containing the patient profile and a time-ordered set of examinations. Of these, 114,239 EHRs are used for the world model training, where the model is tasked with autoregressively reconstructing the examination results recorded in the examination set. These training cases span 4,897 distinct diseases. On average, each training patient underwent 29 examinations, including 26 laboratory tests, 2 microbiological tests, and 1 radiological test.
The remaining 4,239 cases are reserved for evaluation, covering 863 distinct diseases. However, given the high cost of evaluation --- due to the use of various commercial models -- we adopted a disease-wise sampling strategy. Specifically, we selected one representative case for each disease, resulting in a balanced test set of 863 cases, with each case corresponding to a unique disease.

\textbf{Training Details} 

Leveraging the constructed data, we train a diagnostics world model, DiagGym, with text generation loss, as introduced in the former Section~\ref{sec:problem_formulation}.
We frame its training as an auto-regressive text generation task, straightforwardly viewing all examination results as free text, regardless of whether they are numerical or textual. The loss function, a standard token-wise auto-regressive objective inspired by GPT-series models~\cite{radford2019language}, minimizes the negative log-likelihood of the ground-truth examination result $\hat{e}_t$ tokens:
\begin{equation}
    \mathcal{L}_\text{recon} = -\sum_{t=1}^{T} \sum_i \log \Phi_{\text{env}}(\hat{e}_t^i \mid a_{t+1}, E_{t}, \mathcal{B}),
\end{equation}
where $\hat{e}_t^i$ denotes the $i$-th token of the examination result $e_t$ and $E_{t}$ denotes the former examination records in the examination set.

\textbf{Implementation Details.} For our experiments, we initialized $\Phi_{\text{env}}$ from Qwen2.5-Instruct-7B. Training was performed on eight NVIDIA A100 GPUs using the \texttt{Transformers}\footnote{\url{https://github.com/huggingface/transformers}} library with DeepSpeed ZeRO Stage 2 for efficient distributed optimization. Models were trained for 15 epochs, with convergence achieved within this period. The learning rate was $4 \times 10^{-5}$, and the maximum input length was 8,192 tokens.

\subsection{DiagAgent Training} 
In this section, we provide a detailed procedure for DiagAgent training.

\begin{figure}[!t]
    \centering
    \includegraphics[width=\linewidth]{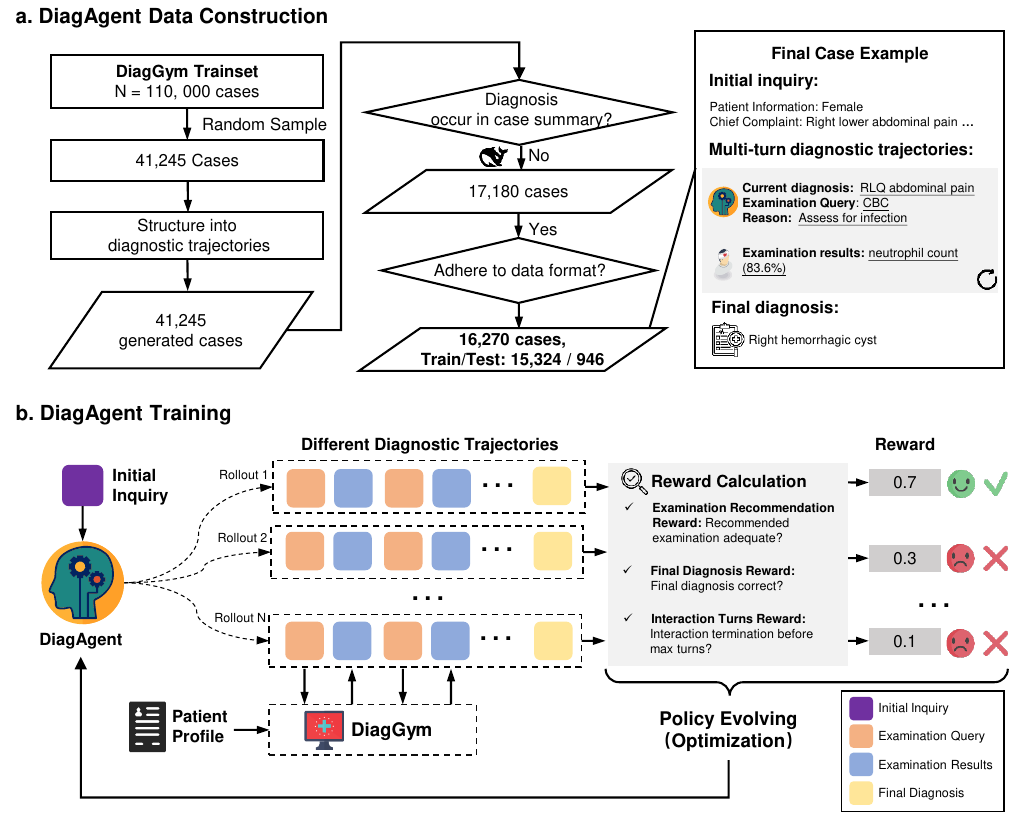}
    \caption{Overview of DiaAgent data construction and training pipeline. \textbf{a} shows the data construction process for the DiagAgent. \textbf{b} illustrates the training pipeline for the diagnostic agent, where the agent interacts with the virtual clinical environment, DiagGym, explores different diagnostic trajectories, and evolves its policy based on reward scores.}
    \label{fig:method_diagnoser}
\end{figure}

\label{sec:Agent_training}
\textbf{Data Construction}
\label{sec:diagnoser_construction}

The diagnostic agent $\Phi_{\text{diag}}$ is trained using a reformatted version of the DiagGym training dataset, organized into a multi-turn diagnostic trajectory format, as shown in Figure~\ref{fig:method_diagnoser}a.

We use DeepSeek-v3, guided by \texttt{prompt~\ref{prompt:generate_differential_diagnosis_data}}, to generate three tightly connected elements for each case based on the existing restructured EHR dataset and the original discharge notes:

\begin{itemize}\setlength\itemsep{3pt}
    \item \textbf{Initial inquiry}: A refined, structured summary of the patient’s medical history before admission, covering the chief complaint, history of present illness, past medical history, family history, and other relevant information. 
    While this is similar to the previously constructed patient profile, critically, it does not include the final diagnosis information. The inquiry is further refined by LLMs to align with real inquiry formats. This serves as the starting point for the dialogue.

    \item \textbf{Referenced multi-turn diagnostic trajectory}: 
    A step-by-step diagnostic trajectory is reformed from the time-ordered examination chain, with DeepSeek-v3 prioritizing the most informative tests related to the final diagnosis and omitting non-essential routine ones. Each step in the trajectory consists of: (1) Current Preliminary Diagnosis: An initial hypothesis based on prior data. (2)Next Recommended Examination with Rationale: A suggested test accompanied by a detailed explanation for its necessity. (3) Corresponding Test Results: The outcome of the recommended examination. The first two components are considered the agent's response, while the third represents the clinical environment's feedback — effectively, the user's input in the multi-turn dialogue. The trajectory concludes with the final diagnostic decision at the end-turn. Although the preliminary diagnosis and the rationale for recommending examinations are generated with the assistance of LLMs, the order of examination items and their corresponding results are directly extracted from real EHRs. Therefore, this trajectory is viewed as a referenced diagnostic pathway, grounded in the recommendations of real physicians.

    \item \textbf{Final diagnosis}: The final diagnostic decision for the case. Since the original final diagnosis recorded in the patient profile may sometimes include multiple conditions, the LLM is prompted to construct a self-contained process focused on a single primary condition. This process also includes selecting relevant examinations from the entire chain of examinations in the previous multi-turn diagnostic trajectory construction.

\end{itemize}

All three components are generated in a single pass to maintain contextual consistency.

To ensure data quality and prevent leakage of the final diagnosis into the initial inquiry, we applied a two-stage filtering process using \texttt{prompt~\ref{prompt:filter_no_data_leakage_differential_diagnosis_data}}.
This step removes instances where the model may have inadvertently introduced the final diagnosis into earlier parts of the text due to hallucinations.

Finally, we converted the multi-turn diagnostic trajectories into a structured dialogue format following former LLM multi-turn datasets~\cite{touvron2023llama}. In subsequent turns, each step in the trajectory is structured as an assistant message that includes the preliminary diagnosis and the next recommended examination with its rationale, along with a user message that provides the results of the recommended examinations. The last assistant turn contains only the final diagnosis and its rationale. The resulting dataset comprised 15,324 interactive diagnostic trajectories used for training.

\textbf{Training Details}

As introduced in Section~\ref{sec:problem_formulation}, we train DiagAgent  $\Phi_{\text{diag}}$ with end-to-end multi-turn RL, with a classical two-stage paradigm~\cite{liu2024deepseek}: an initial cold-start phase and a main RL phase.

\textbf{Cold start.} This phase mirrors standard instruction tuning~\cite{ouyang2022training}. 
The model is optimized with an auto-regressive text generation loss computed only over tokens labeled as \textbf{assistant} response in the dialogues:
\begin{equation}
    \mathcal{L}_{\text{cold}} = -\sum_{y_i \in \texttt{assitant}} \log \Phi_{\text{diag}}(y_i \mid y_{\leq i-1}),
\end{equation}
where $y_i$ is a token, $y_{\leq i-1}$ is the preceding context, and the loss is restricted to assistant response tokens. The goal of cold start is to initialize the LLM to produce well-formatted, contextually appropriate responses before interacting with the environment with RL. For this stage, we use 1,000 manually selected high-quality cases from the training set, free from formatting issues or diagnostic reasoning errors.

\textbf{Reinforcement learning.} After cold start, we optimize $\Phi_{\text{diag}}$ with the GRPO algorithm~\cite{liu2024deepseek} over the full training set. 
At rollout start, the agent receives the initial inquiry as the initial state, $s_0 = \mathcal{I}$ and iteratively interacts with the virtual environment $\Phi_{\text{env}}$ until it decides sufficient information has been gathered for a final diagnosis (Section~\ref{sec:problem_formulation}
). The policy is trained to maximize the following reward:
\begin{equation}
    R = \lambda_1 r_\text{diag} + \lambda_2 r_\text{exam} + \lambda_3 r_\text{turn},
\end{equation}
with $\lambda_1, \lambda_2, \lambda_3$ as hyper-parameter weights.
The final diagnosis reward evaluates the accuracy of the predicted diagnosis results, formulated as:
\begin{equation}
    r_\text{diag} =
    \begin{cases}
        1, & \text{if } \hat{d} = d  \\
        0, & \text{otherwise}
    \end{cases}, 
\end{equation}
where $\hat{d}$ is the predicted disease. We adopt Qwen2.5-72B with \texttt{prompt~\ref{prompt:accuracy_in_diagnose_result}} to measure 
semantic equivalence of $\hat{d}$ and $d$. The examination recommendation reward measures the alignment between the agent’s recommended examinations $\hat{E}$ and the reference set $E$ from the curated multi-turn trajectory based on real EHRs. We adopt the F1 score to measure their similarity, formulated as:
\begin{equation}
    r_\text{exam} = \text{F1}(\hat{E},E) =2 \cdot \frac{|\hat{E} \cap E|}{|\hat{E}| + |B|}.
\end{equation}
We calculate the union set of the two by instructing Qwen2.5-72B with \texttt{prompt~\ref{prompt:precision_in_exam_recommendation}} and \texttt{prompt~\ref{prompt:recall_in_exam_recommendation}} to search through the $\hat{E}$ and $E$ respectively. The last interaction turn penalty reward penalizes excessive rounds of dialogue without termination. 
To prevent unnecessarily long dialogues, we impose a maximum number of iterative turns, $T_{max}$. If the model cannot finish the diagnosis within this limitation, it will achieve a lower reward as:
\begin{equation}
    r_\text{turn} =
    \begin{cases}
        0.1, & \text{if } T \leq T_{max} \\
        0, & \text{otherwise}
    \end{cases}, 
\end{equation}
where $T$ denotes the total number of turns used in a certain rollout. 

Through iteratively optimizing this reward, $\Phi_{\text{diag}}$ learns to manage accurate, efficient diagnostic trajectory reasoning while minimizing unnecessary examination steps.

\textbf{Implementation details.}
In our experiments, we selected Qwen2.5-Instruct-7B, Qwen2.5-Instruct-14B, and Llama3.1-Instruct-8B as initialization models for further training. The same training strategy was applied to all models. 
For cold start settings, we utilize the \texttt{Transformers}\footnote{\url{https://github.com/huggingface/transformers}} framework and employed DeepSpeed ZeRO Stage 2 for efficient multi-GPU training. All models share the same hyperparameters: a maximum sequence length of 8192 tokens, a learning rate of $1\times10^{-5}$, and training on eight NVIDIA A100 GPUs. Training is conducted for three epochs, \textbf{at which point the training loss plateaued, confirming convergence for each model.}

For the reinforcement learning setting, we modify \texttt{Verl}\footnote{\url{https://github.com/volcengine/verl}} to enable interactive training. During training, the \textbf{DiagGym} was deployed on two nodes using \texttt{vLLM}\footnote{\url{https://github.com/vllm-project/vllm}}. The Qwen2.5-Instruct-72B model serves as the judge and is deployed on a separate node. 
\change{The weighted hyperparameters $\lambda_1,\lambda_2,\lambda_3 $ are set as 1, 0.5, 1, respectively. These values were selected based on the magnitude of reward components and preliminary experiments to ensure stability, as exhaustive grid search was computationally infeasible. The maximum interactive turns are set to 12, a threshold empirically derived from the MIMIC-IV diagnostic trajectory distribution to cover realistic workflows while preventing redundant loops.}
RL training is performed across four nodes in total, with each node equipped with eight NVIDIA A100 GPUs. We set the training batch size to 512, the maximum response length to 8192 tokens, the learning rate to $1\times10^{-6}$, and the rollout number to 5. Each model is trained for 200 steps, \change{where convergence was confirmed by the stabilization of the average cumulative reward per rollout.}

\subsection{DiagBench Construction}
\label{sec:method_diagagent_rubrics}

\begin{figure}
    \centering
    \includegraphics[width=0.9\linewidth]{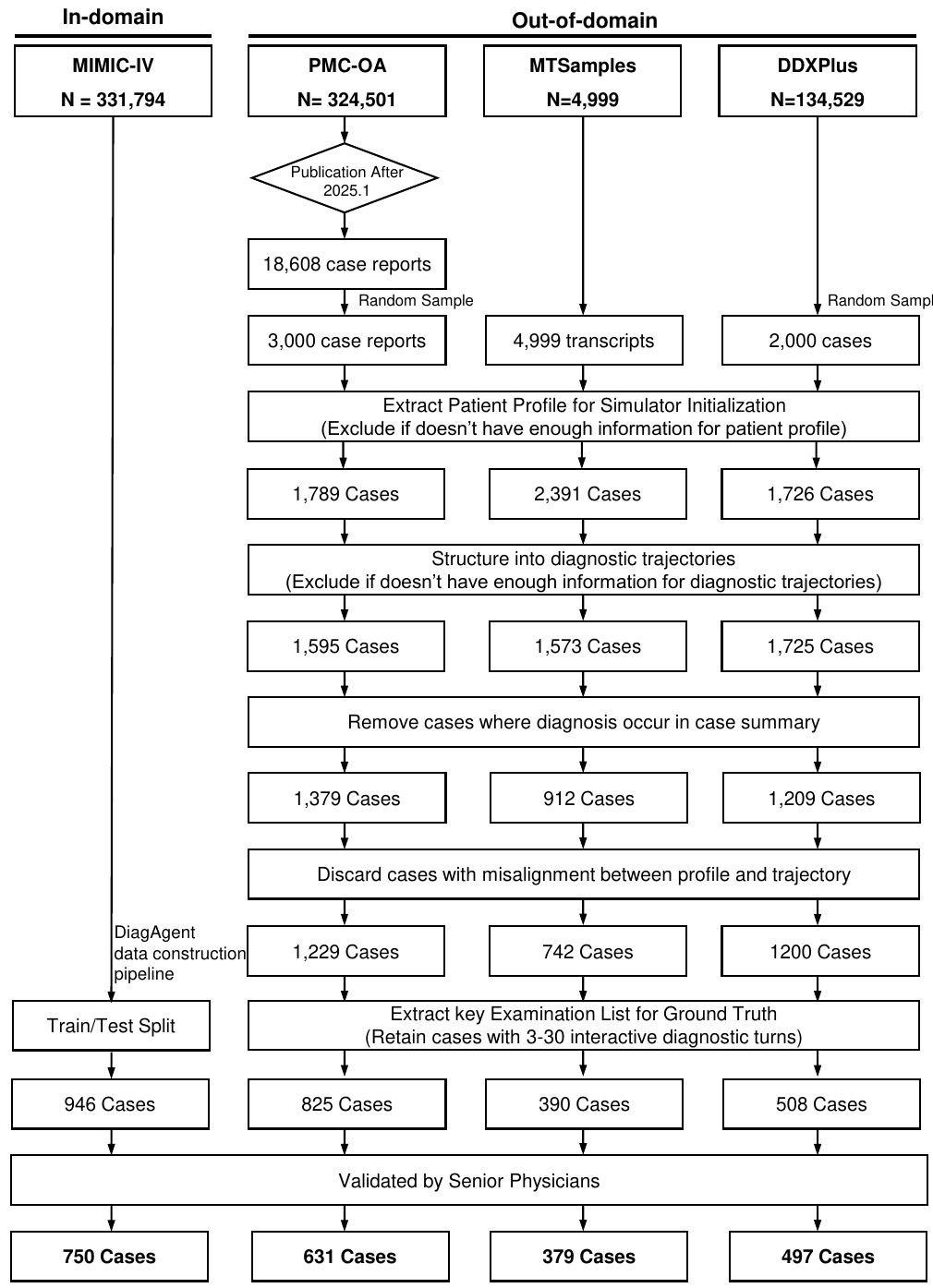}
    \vspace{6pt}
    \caption{\textbf{Overview of the DiagBench construction pipeline.} To ensure robust evaluation across diverse clinical settings, the benchmark integrates data from four distinct sources: MIMIC-IV, PMC-OA, MTSamples, and DDXPlus. As our training data originates from MIMIC-IV, the benchmark can be further divided into in-domain and out-of-domain subsets for the final DiagAgent evaluation. The pipeline consists of three key phases: (1) \textbf{Structured Extraction}, where LLMs synthesize patient profiles and reference trajectories from raw text; (2) \textbf{Automated Filtration}, which rigorously removes cases with data leakage or logical inconsistencies; and (3) \textbf{Human Validation}, where senior physicians verify the clinical plausibility of the final dataset. The counts in the boxes indicate the number of cases retained at each stage.}
    
    \label{fig:method_diagbench_construction}
\end{figure}

In this section, we describe the construction of \textbf{DiagBench}, \change{a comprehensive multi-center benchmark designed to evaluate multi-turn diagnostic interaction trajectories across diverse clinical settings. As illustrated in Figure~\ref{fig:method_diagbench_construction}, DiagBench integrates data from four distinct sources to ensure robust assessment of generalization:}
\change{\begin{itemize}
    \setlength \itemsep{3pt}
    \item \textbf{MIMIC-IV}: Derived from the same source as the training set, representing critical care and emergency medicine scenarios.
    \item \textbf{PMC-OA}: Curated from open-access biomedical case reports, representing complex and rare disease presentations found in medical literature. To rigorously prevent data leakage and assess generalization on unseen data, we strictly selected case reports published after January 2025.
    \item \textbf{MTSamples}: Sourced from transcribed medical reports, covering a wide range of outpatient and specialty clinic encounters.
    \item \textbf{DDXPlus}: A dataset focused on differential diagnosis logic, providing structured cases for reasoning evaluation.
\end{itemize}}

The process involved two main stages: initial dataset curation and the development of a rubric-based evaluation framework.

\textbf{Initial Dataset Curation: } 
The construction pipeline is illustrated in Figure~\ref{fig:method_diagbench_construction}. 
For MIMIC-IV, the test cases are generated using the same pipeline as our training data. We applied a standard train/test split to obtain them, forming an in-domain evaluation for our DiagAgent. 
\change{For the other out-of-domain datasets, the construction pipeline aligns generally with MIMIC-IV, but required specific adaptations to accommodate their heterogeneity:}
\change{\begin{itemize}
    \setlength \itemsep{3pt}
    \item \textbf{Patient Profile Generation:} Unlike MIMIC-IV, other datasets often lack structured patient profiles. We utilize DeepSeek-v3 with \texttt{prompt~\ref{prompt:generate_patient_profile}} to extract and synthesize a comprehensive profile from raw text to initialize the simulator. 
    \item \textbf{Diagnostic Trajectory Generation:} We generate multi-turn diagnostic trajectories utilizing the same prompt structure as the MIMIC-IV pipeline (\texttt{prompt~\ref{prompt:generate_differential_diagnosis_data}}). However, these datasets frequently contain rare specialized examinations. To address this, we append a strict constraint: ``You can only state Lab Events, Microbiology Events, or Radiology.'' This limits recommendations to conventional diagnostic modalities, thereby preventing clinically outlier cases.
    \item \textbf{Key Examinations Extraction for Hit Ratio:} To evaluate examination recommendation performance under single-turn evaluation setting, we extract a list of all relevant examinations mentioned in the raw text using \texttt{prompt~\ref{prompt:extract_all_exams}}.
    \item \textbf{Logical Consistency Check:} We implement an additional filtering step with \texttt{prompt~\ref{prompt:consistency_patient_profile_and_diag}} to discard cases where the synthesized patient profile conflicted with the diagnostic trajectory, ensuring the logical flow of the case.
\end{itemize}}
\change{To ensure clinical validity, each case undergoes a rigorous review by a physician. The reviewer is provided with the patient's initial inquiry and the complete reference diagnostic trajectory. They are tasked with evaluating: (1) the clinical appropriateness of each analysis and recommended examination at every step, and (2) the overall plausibility of the case. For the out-of-domain datasets, they are additionally required to review the patient profile and the extracted key examinations to confirm relevance to the patient's situation. This curation process results in a final, validated set of cases for automated evaluation.
}

\change{Figure~\ref{fig:method_diagbench_construction} illustrates the exclusion rates at each step during data construction, highlighting the distinct characteristics of each source.
\textbf{PMC-OA} case reports are clinically complex and often detail severe comorbidities. While we sample 3,000 recent reports, this complexity makes structured extraction challenging. Consequently, a significant portion is filtered during the profile and trajectory extraction phases or rejected during strict physician review, resulting in \textbf{631 final cases}.
\textbf{MTSamples} transcripts are often concise or contain incomplete information. This leads to a high filtration rate at the initial ``Patient Profile Extraction'' stage, where nearly 50\% of the raw inputs are discarded due to insufficient details for simulation initialization. Ultimately, \textbf{379 high-quality cases} are retained.
\textbf{DDXPlus}, designed for differential diagnosis, features structured and clear logic. Consequently, the construction is highly stable, with the ``Reference Trajectory Extraction'' step retaining almost all candidates. \textbf{497 cases} are selected for the final benchmark. Combined with the \textbf{750 cases} from MIMIC-IV, DiagBench provides a diverse evaluation suite.}

\textbf{Rubric-based Evaluation Framework}
In addition to automated metrics, and inspired by HealthBench~\cite{arora2025healthbench}, we develop a \textbf{rubric-based framework} to comprehensively assess the quality of the diagnostic inference process.

\change{First, to ensure balanced representation, we perform stratified random sampling across all four data sources in DiagBench, selecting 100 clinical cases from each center. Two physicians are independently provided with each case's initial patient inquiry and complete diagnostic trajectory. Each physician then independently authors a set of process-oriented rubrics they consider critical for evaluating the quality of the diagnostic journey. The rubric design explicitly prioritizes the reasoning process, including history taking, hypothesis generation, and test ordering, rather than focusing solely on the accuracy of the final diagnosis.}


\change{Subsequently, a third physician conducts a secondary review of the rubrics. This reviewer's role is to ensure the framework holistically covers the end-to-end diagnostic pipeline and that individual rubrics are coherent and well-defined. Following this screening, cases with insufficient rubric coverage are excluded, yielding a final consolidated set of \textbf{399 cases} (99 from MIMIC-IV and 100 from each of the three OOD centers) annotated with a total of \textbf{3,318 physician-authored rubrics}.}

Finally, a fourth physician is tasked with assigning an importance weight to each rubric on a scale from 0 to 10 (where 10 signifies an essential, non-negotiable criterion and 0 indicates a non-informative one). As a result, rubrics targeting high-impact clinical steps, such as appropriate test ordering and effective diagnostic narrowing, receive higher weights. Conversely, ancillary actions, like scheduling a follow-up without a clear clinical rationale, are assigned lower weights. Illustrative examples of this rubric-based evaluation are provided in Supplementary Figure~\ref{fig:rubric_success_mode} and Supplementary Figure~\ref{fig:rubric_fail_mode}.

\subsection{Baselines}
\label{sec:baselines}
Here, we introduce the baseline LLMs involved in our experiments:

\vspace{-6pt}
\begin{itemize}\setlength\itemsep{3pt}
    \item \textbf{Qwen2.5}~\cite{qwen2.5} and \textbf{Qwen3}~\cite{qwen3} are series of high-performance open-source language models developed by the Qwen team, available in variants ranging from 0.5 to 72 billion parameters. In this paper, we use \texttt{Qwen2.5-7B-Instruct} and \texttt{Qwen2.5-14B-Instruct} for training, and deploy \texttt{Qwen2.5-72B-Instruct} and \texttt{Qwen3-235B-A22B} locally for inference. 
    
    \item \textbf{Llama3.1} and \textbf{Llama3.3}~\cite{llama3} are series of language models developed by Meta AI and are among the most popular open-source large language models. In this paper, we utilize \texttt{Llama-3.1-8B-Instruct} for training and deploy \texttt{Llama-3.3-70B-Instruct} locally for inference.
    
    \item \textbf{OpenBioLLM}~\cite{OpenBioLLMs} is an advanced open-source language model specifically designed for the biomedical domain, developed based on Llama3. In this paper, we utilize \texttt{Llama3-OpenBioLLM-70B} and deploy it locally.
    
    \item \textbf{Baichuan-M1}~\cite{baichuan-m1} is an advanced open-source medical language model developed by Baichuan Intelligence. It is the first language model in the industry designed and developed from scratch specifically for the medical field, demonstrating strong performance in medical applications. We utilize \texttt{Baichuan-M1-14B-Instruct} and deploy it locally.

    \item \change{\textbf{Baichuan-M3}~\cite{baichuan-m3} is a next-generation medical-enhanced large language model developed by Baichuan Intelligence. Unlike traditional models focusing on static Q\&A, Baichuan-M3 achieves native medical enhancement primarily through advanced reinforcement learning frameworks. These strategies enable the model to autonomously collect key information, build rigorous clinical reasoning paths, and dynamically suppress hallucinations in real-world scenarios. We utilize \texttt{Baichuan-M3-235B} and deploy it locally.}
    
    \item \textbf{DeepSeek-V3}~\cite{deepseekv3}  is one of the most powerful open-source language models, developed by DeepSeek, with 671 billion parameters. In this paper, we use \texttt{DeepSeek-V3-0324} and deploy it locally.
    
    \item \textbf{MedGemma}~\cite{sellergren2025medgemma} is a recent medical LLM developed by Google as a variant of the Gemma3 collection. Based on Gemma3, this model is specially optimized for the medical field and possesses multi-modal capabilities. In this paper, we utilize \texttt{medgemma-27b-text-it} and deploy it locally.
    
    \item \textbf{GPT-OSS}~\cite{gptoss} is an open-source large language model released by OpenAI, with strong reasoning ability. In this paper, we use the \texttt{gpt-oss-120b} and deploy it locally.

    \item \textbf{GPT-4o}~\cite{openai_gpt4o} is one of the most commonly used close-sourced language model developed by OpenAI. It is good at handling most daily tasks. In this paper, we utilize \texttt{gpt-4o-2024-08-06} via API.

    \item \textbf{Claude-4}~\cite{claude4} is a high-performance large language model developed by Anthropic, optimized for coding and reasoning. In this paper, we use the \texttt{claude-sonnet-4} via API.

\end{itemize}

Next, we will introduce the agentic systems involved in our experiments. Note that, in our experiment, all the agentic systems use DeepSeek-V3 as its base model.
\begin{itemize}\setlength\itemsep{3pt}
    \item \textbf{MedAgents}\cite{medagents} is a multidisciplinary collaboration framework that requires large language models to assume the roles of medical experts from various specialties. The framework aggregates the experts’ opinions and summarizes them into a final report. This approach reveals the model’s knowledge across different domains and broadens its reasoning capabilities. In our implementation, we use DeepSeek-V3 as the base model.
    \item \textbf{MDAgents}\cite{mdagentsadaptivecollaborationllms} is a framework designed to automatically assign collaboration structures. Different from MedAgents, MDAgents effectively assesses medical complexity and adapt to varing tasks accordingly. In our implementation, we use DeepSeek-V3 as the base model.
\end{itemize}

In the \textbf{DiagGym evaluation}, we consider prompting the open-source models as EHR wold model baselines to simulate the examination results, including DeepSeek-V3-671B, MedGemma-27B, Qwen2.5-7B, and Qwen2.5-72B.
We exclude closed‑source API-based LLMs due to their high latency and cost, which make them impractical for scalable RL training. 
For instance-wise metrics, these models adopt \texttt{prompt~\ref{prompt:instruction_to_generate_as_simulator}} to generate examination results for comparison.
For examination-wise metrics, which measures the distribution of given examinations, the models use \texttt{prompt~\ref{prompt:instruction_to_generate_as_simulator_labevent}} for numerical examinations and \texttt{prompt~\ref{prompt:instruction_to_generate_as_simulator_radiology}} for free-text examinations. \change{We implement zero-shot inference for most settings, with the sole exception of \texttt{prompt~\ref{prompt:instruction_to_generate_as_simulator_labevent}}. Preliminary experiments indicated that under a zero-shot setting, baseline models frequently generated numerical results with inconsistent formatting or heterogeneous units, which severely impeded automated parsing. Consequently, we adopted a one-shot strategy specifically for this prompt to enforce syntactic constraints and ensure reliable quantitative evaluation. }

In the \textbf{DiagAgent evaluation}, we include all the aforementioned basic LLMs and more complex agentic systems for comparison.
\change{We consistently apply a zero-shot prompting strategy across all models. Each system is instructed using \texttt{prompt~\ref{prompt:instruction_for_llm_diagnose}}, which serves as a \textbf{standardized test interface} to unify the task definition and output format, ensuring fair comparison.}
For end-to-end evaluation on simulated cases, the models are required to decide whether to request an additional examination or to make a final diagnosis.
For single-turn evaluation on real cases, the models are directly instructed to either request an examination or make a final diagnosis. \change{In this setting, we modified \texttt{prompt~\ref{prompt:instruction_for_llm_diagnose}} by appending an explicit prompt to the end of the input in each turn, like ``Next step you should query examination'' or ``Next step you should make final diagnosis''.}

%% file: content_npj/05_Conclusion.tex
\section{Data Availability}
The data source for this work is MIMIC-IV. Due to licensing restrictions, we are unable to directly open-source the dataset. However, we are actively communicating with the relevant parties regarding the possibility of making the dataset publicly available on \url{https://physionet.org/}. 

\section{Code Availability}
All source codes of this paper have been released in \url{https://github.com/MAGIC-AI4Med/DiagGym}.


%% file: content_npj/06_Appendix.tex
\section{Supplementary}
\setcounter{table}{0}   
\setcounter{figure}{0}
\renewcommand{\tablename}{Supplementary Table}
\renewcommand{\figurename}{Supplementary Figure}

\subsection{Evaluation Metrics}

In this section, we provide a detailed calculation format for our adopted evaluation metrics.

\subsubsection{DiagGym Evaluation}
\label{sec:simulator_evaluation}
We evaluate the DiagGym ($\Phi_{\text{env}}$) from two perspectives, \emph{i.e.}, instance-wise and examination-wise.

\textbf{Instance-wise Metrics.} 
\change{We employ a hybrid evaluation strategy combining 2 traditional quantitative metrics with 2 LLM-based semantic assessments to evaluate the quality of generated examination sequences at the instance level: }

\begin{itemize}\setlength\itemsep{3pt}

    \item \change{\textbf{Normalized Mean Absolute Error (NMAE).} Designed specifically for \textbf{numeric laboratory results}, Designed specifically for \textbf{numeric laboratory results}, this metric quantifies the deviation between the simulator's predicted values and the ground truth. Since different laboratory tests operate on vastly different scales (e.g., pH $\approx$ 7.4 vs. Platelets $\approx$ 300), we normalize the absolute error relative to the global data range of each specific test type. Formally, for a specific examination type $k$ (e.g., Glucose), the NMAE is calculated as:
    \begin{equation}
    \text{NMAE}_k = \frac{1}{N_k} \sum_{i=1}^{N_k} \frac{|y_{i,k} - \hat{y}_{i,k}|}{V_{\max, k} - V_{\min, k}},
    \end{equation}
    where $y_{i,k}$ and $\hat{y}_{i,k}$ are the ground truth and predicted values for the $i$-th instance, and $V_{\max, k}$ and $V_{\min, k}$ represent the maximum and minimum values of this examination type across the entire dataset. The final reported metric is the macro-average of $\text{NMAE}_k$ across all examination types.}

    \item \change{\textbf{BLEU Score.} For \textbf{free-text radiology reports}, we utilize the standard BLEU metric to measure the n-gram overlap between the generated narratives and the reference reports. This provides a fundamental assessment of textual similarity and lexical coverage.}
    
    \item \textbf{Step-wise Similarity.} This metric evaluates the similarity between the model-predicted examination results and the ground truth at each step. During this evaluation, the simulator generates the current examination results conditioned on the queried examination names, the patient case summary, and all historical ground truth examination names and results. At each step, we apply Prompt~\ref{prompt:eval_similarity_step_wise} to compute similarity scores (ranging from 0 to 5) between the model-generated and ground truth examination results.
    \item \textbf{Full-chain Consistency.} This metric assesses whether the generated examination chain is internally consistent and aligns with the patient prfile. Given the generated chain, we use Prompt~\ref{prompt:eval_full_chain_consistency} to assess whether the sequence of generated examination results maintains logical and clinical consistency. The evaluation final score is binary 1/0, indicating yes/no.
\end{itemize}

\textbf{Examination-wise Metrics}
For the examination-wise metrics, we primarily assess the statistical distribution quality for both generative numerical and free-text results, covering fidelity and diversity.

\begin{itemize} \setlength\itemsep{3pt}
    \item \textbf{Numerical Fidelity \& Diversity.} For numerical examination results, we utilize the 1-Wasserstein distance to measure the generative distribution fidelity, where shorter distances indicate closer alignment.
    Formally, considering a certain numerical examination, its real distribution is characterized by the mean $ \mu $ and standard deviation $ \sigma $. Similarly, the generative distribution is characterized by $\hat{\mu}$ and $\hat{\sigma}$.
    Denoting a generative examination value on a certain test case as $x_i$, it can be normalized as $z_i = (x_i-\mu)/\sigma $ and similarly, for the ground-truth results, we have $\hat{z}_i = (\hat{x}_i-\hat{\mu})/\hat{\sigma}$. The 1-Wasserstein distance is formulated as: 
    \begin{equation}
    W_1(Z, \hat{Z}) = \inf_{\gamma \in \Gamma(Z, \hat{Z})} \int \|z_i - \hat{z}_i\| \, d\gamma(z_i, \hat{z}_i),
    \label{eq:wasserstein}
    \end{equation}
    where $ Z $ and $ \hat{Z} $ denote the generative and real distributions, and $ \Gamma(Z, \hat{Z}) $ is the joint distribution. 
    
    Then for diversity, we adopt normalized variance to evaluate its distribution diversity, with higher variance reflecting greater diversity. Formally, following the former notation, the normalized variance is defined as:
    \begin{equation}
    \sigma^2(X) = \frac{\text{Var}(X)}{\mu^2},
    \label{eq:norm_var}
    \end{equation}
    where $\text{Var}(\cdot)$ is the distribution variance. 
    Notably, for simplicity, all the above numerical metric formulations assume that the examination item contains only a single value item. In practice, some examinations may consist of multiple value items, such as a ``Complete Blood Count'', which may include multiple numerical values, and in these cases, the metrics are calculated by averaging the scores across all value items. Furthermore, to ensure consistency, all computations for the same value items are standardized to unified value units.

    \item \textbf{Free-text Fidelity \& Diversity.} For free-text results, we first encode the text into feature embeddings using BioLORD~\cite{remy2024biolord}, a biomedical text encoding model. Inspired by metrics commonly used in image generation, we then calculate the Fréchet Inception Distance (FID)\cite{yu2021frechet} in the embedding space to assess the fidelity. Lower FID values indicate better alignment with the ground truth. Specifically, considering a certain free-text-related examination, such as a chest CT examination, let the set of genrative free-text embeddings be $ F = \{f_1, f_2, \cdots, f_N\} $, where $ N $ denotes the total number of test cases. The corresponding ground truth text embeddings are denoted as $ \hat{F} = \{\hat{f}_1, \hat{f}_2, \cdots, \hat{f}_N\} $. The FID score can then be calculated as:
    \begin{equation}
        \text{FID}(F, \hat{F}) = \|\mu_F - \mu_{\hat{F}}\|_2^2 + \text{Tr}\left(\Sigma_F + \Sigma_{\hat{F}} - 2\left(\Sigma_F \Sigma_{\hat{F}}\right)^{\frac{1}{2}}\right),
    \end{equation}
    where $ \mu_F $ and $ \Sigma_F $ represent the mean and covariance of the generative embeddings $ F $, and $ \mu_{\hat{F}} $ and $ \Sigma_{\hat{F}} $ are for the groud turth embeddings $ \hat{F} $. The term $ \|\mu_F - \mu_{\hat{F}}\|_2^2 $ quantifies the difference between the means, while the trace term measures the distance between the covariance matrices. 
    To evaluate diversity, inspired by the Intra-LPIPS metric~\cite{ojha2021few} used in image generation, we propose using inter-case cosine similarity on the entire set of generated text embeddings. This metric reflects how well the embeddings distinguish from one another, defined as:
    \begin{equation}
        \text{Intra-LPIPS}(F) =1 - \frac{2}{N(N-1)}\sum_{i<j}\cos(f_i,f_j),
    \end{equation}
    where $ \cos(\cdot, \cdot) $ represents the cosine similarity function. A higher Intra-LPIPS score indicates that the generated free texts are more diverse in comparison to one another.
\end{itemize}

\subsubsection{DiagAgent Evaluation}
\label{sec:diagagent_evaluation}
We evaluate DiagAgent's performance across three key perspectives: final diagnosis accuracy, examination recommendation efficacy, and rubric-based diagnostic trajectory assessment.

\textbf{Final Diagnosis Accuracy.} For measure the final diagnosis, we adopt the straightforward \textbf{Accuracy} metric here. Similarly, considering diseases may have synonyms, we instruct GPT-4o utilizing Prompt~\ref{prompt:accuracy_in_diagnose_result} to compare the model’s diagnostic output with the reference standard.

\textbf{Examination Recommendation Metrics}  
In the end-to-end evaluation setting, we compare the predicted examination list against the referenced list and adopt \textbf{Precision}, \textbf{Recall}, and \textbf{F1} scores. Considering that the same examination may be expressed in synonyms, we instruct GPT-4o employing Prompt~\ref{prompt:precision_in_exam_recommendation} to count the number of examination names generated by the model that are covered by the true examination list, and Prompt~\ref{prompt:recall_in_exam_recommendation} to count the number of ground truth examination names that are present in the model’s output. Precision and recall are then computed accordingly, followed by the calculation of the F1 score. 
In the single-turn evaluation setting, we adopt the hit ratio metric, counting whether the recommended examination item appears in the referenced following referrenced list. We adopt GPT-4o with Prompt~\ref{prompt:instruction_to_check_if_exam_in_the_list} to determine whether this query appears among the examinations actually undergone by the patient. 

\textbf{Weighted Rubric Score}
To evaluate the clinical integrity of the multi-turn diagnostic interaction, we use the Weighted Rubric Score. This metric goes beyond the final outcome by qualitatively assessing the entire diagnostic trajectory against physician-authored, process-oriented rubrics. The score is calculated as the weighted proportion of satisfied rubrics, with weights reflecting the clinical significance of each step. The final score $\bar{s}$ is the average across all cases, calculated as:
$$\bar{s} = \frac{1}{|C|} \sum_{c \in C} \frac{\sum_{r \in A_c} w_r}{\sum_{r \in R_c} w_r}.$$
The $\text{GPT-4o}$ judge model determines rubric satisfaction for each case (Prompt $\text{\ref{prompt:check_each_rubric_hit}}$).

\subsubsection{\change{Human Validation of Automated Evaluation Metrics}}
\label{sec:human_validation}
\change{To ensure the reliability of our LLM-based automated evaluation framework, we conduct a rigorous human validation study.}

\change{We organized a large-scale expert panel comprising \textbf{12 senior physicians}, each with over 10 years of clinical experience. They performed the same evaluations as the automated LLM judge for the following metrics:
\begin{enumerate}
    \item \textbf{Diagnosis Accuracy:} A binary task determining whether the model's predicted diagnosis matches the ground truth label.
    \item \textbf{Rubric Hit Ratio:} A binary task assessing whether a model's diagnostic trajectory met specific physician-authored rubric criteria, which is the prerequisite step for calculating weighted rubric score.
    \item \textbf{Single-turn Hit Ratio:} A binary task verifying if a recommended examination appeared in the reference list.
    \item \textbf{Examination Recommendation Count:} A numerical task counting the number of correct matches (True Positives) between the predicted examination list and the reference list, which is the prerequisite step for calculating Precision, Recall, and F1 scores.
\end{enumerate}}

\change{Due to the cost limitations of physician annotation, we randomly sample 100 instances for each metric. To ensure assessment reliability, each instance is \textbf{independently annotated by three different physicians} from the panel. We further aggregate the annotations from the assigned physician triplet. For binary classification tasks, we define the GT using the majority vote (consensus of at least 2 out of 3 physicians). Consistency is measured using the percentage of agreement between the LLM and the GT. For the numerical task, we define the GT using the median of the three physicians' counts to mitigate outlier effects. Consistency is measured using the Mean Absolute Error (MAE) between the LLM's count and the human median.}

\change{As detailed in Supplementary Table~\ref{tab:human_eval}, the LLM judge demonstrated high reliability across all metrics. For binary tasks, the LLM achieved near-perfect alignment with the human consensus (e.g., 100.00\% for Diagnosis Accuracy and 96.00\% for Rubric Hit Ratio). Notably, the LLM's consistency with the consensus often exceeded the average consistency of individual physicians (e.g., 100.00\% vs. 92.67\% for Diagnosis Accuracy), suggesting that the LLM judge functions as a standardized evaluator, mitigating the subjective variance and fatigue often observed in human annotation. For the numerical examination counting task, the LLM yielded a low MAE of 0.25. This error margin is comparable to the internal disagreement among physicians themselves (MAE 0.15), confirming that the LLM judge can effectively handle the semantic ambiguity of medical terms during counting tasks. These results substantiate the validity of our automated evaluation framework.}

\begin{table}[!h]
\centering
\footnotesize
\caption{Consistency analysis between Human Physicians and LLM-as-judge across 100 sampled cases. The \textbf{Ground Truth (GT)} is derived from the consensus of three independent senior physicians: using \textbf{majority vote} for Diagnosis Accuracy, Rubric Hit Ratio, Single-turn Hit Ratio; and the \textbf{median} value for Exam Rec. Count. Results indicate high alignment between the LLM-as-judge and the physician-derived GT. `Phy' denotes physician and `Rec.' denotes recommendation.}
\label{tab:human_eval}
\resizebox{0.8\linewidth}{!}{%
\begin{tabular}{lccccc}
\toprule
\multirow{2}{*}{\textbf{Metric}} & \multicolumn{3}{c}{\textbf{Phy. vs. AI}} & \textbf{GT vs. Phy.} & \multirow{2}{*}{\textbf{GT vs. AI}} \\
\cmidrule(lr){2-4} \cmidrule(lr){5-5} 
 & \textbf{Phy. 1} & \textbf{Phy. 2} & \textbf{Phy. 3} & \textbf{Avg. of Three} &  \\ 
\midrule
Diagnosis Accuracy & 82.00\% & 100.00\% & 96.00\% & 92.67\% & \textbf{100.00\%} \\
Rubric Hit Ratio & 62.00\% & 91.00\% & 96.00\% & 85.00\% & \textbf{96.00\%} \\
Single-turn Hit Ratio & 94.00\% & 94.00\% & 98.00\% & 96.67\% & \textbf{96.67\%} \\ 
\midrule
 & \multicolumn{5}{c}{\textit{Mean Absolute Error (MAE) $\downarrow$}} \\
\cmidrule(lr){2-6}
Exam Rec. Count & 0.12 & 0.33 & 0.35 & 0.15 & \textbf{0.25} \\
\bottomrule
\end{tabular}%
}
\end{table}

\subsection{Supervised Finetuning}
\label{sec:supervised_finetuning}

In this section, we introduce the multi-turn supervised fine-tuning (SFT) paradigm.
In multi-turn SFT, each training sample consists of a dialogue history and the next expected response. Formally, let $\mathcal{D}_{\text{multi}} = \{(h^{(j)}_t, y^{(j)}_t)\}_{j=1}^M$, where $h^{(j)}_t = (u^{(j)}_1, r^{(j)}_1, \ldots, u^{(j)}_{t-1}, r^{(j)}_{t-1}, u^{(j)}_t)$ denotes the dialogue history up to turn $t$ (with $u$ as user inputs and $r$ as model responses), and $y^{(j)}_t$ is the supervised response. The loss becomes:

\begin{equation}    
\mathcal{L}_{\text{multi-turn SFT}} = - \sum_{j=1}^M \log P_\theta(y^{(j)}_t \mid h^{(j)}_t)
\end{equation}

\subsection{\change{Ablation on Training Compute (FLOPs)}}
\label{sec:ablation_on_flops}

\change{To investigate whether the performance limitations of Supervised Fine-tuning (SFT) baselines stem from insufficient training steps or computational budget, we conducted an ablation study focusing on training duration. Using \textbf{Qwen2.5-7B} as the representative base model, we extended the fine-tuning process to 10 epochs --- significantly beyond the standard convergence point ---and monitored the performance on the end-to-end diagnostic task.}

\change{As illustrated in Supplementary Figure~\ref{fig:flops_ablation}, the results reveal a clear performance ceiling inherent to the SFT paradigm. 
\begin{itemize} 
\item \textbf{Rapid Convergence:} The model learns the static data distribution rapidly. Both the Final Diagnostic Accuracy (Supplementary Figure~\ref{fig:flops_ablation}a) and the Examination Recommendation F1-score (Supplementary Figure~\ref{fig:flops_ablation}b) reach their peak performance approximately at the \textbf{3rd epoch}.
\item \textbf{Performance Plateau:} Extending training beyond this point (from epoch 3 to 10) yields negligible gains, with the performance curves flattening out. This plateau confirms that simply increasing training FLOPs or iterations does not enable the model to discover better diagnostic strategies. \end{itemize}}

\change{This finding reinforces our core motivation: the performance gap between baselines and DiagAgent is not due to under-fitting, but rather the limitation of supervised imitation. SFT is bounded by the amount of the static training data, whereas our RL-based approach enables the agent to break through this ceiling by actively exploring and optimizing for long-term diagnostic utility.}

\begin{figure}[h]
    \centering
    \includegraphics[width=\linewidth]{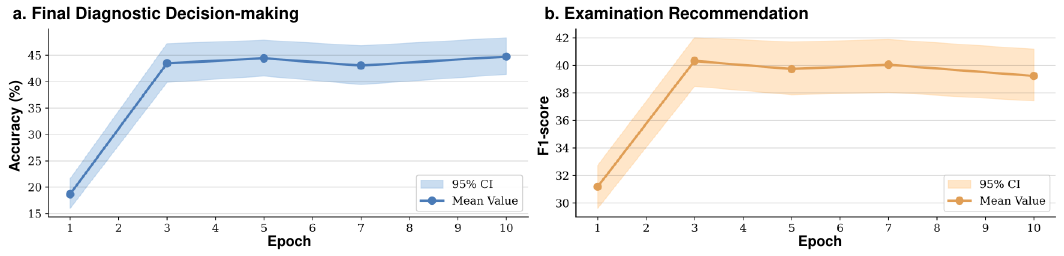}
    \caption{\textbf{Ablation study on the impact of training epochs (FLOPs) for Supervised Fine-tuning (SFT).} We conduct the experiment using Qwen2.5-7B under the end-to-end evaluation setting. \textbf{a} illustrates the Final Diagnostic Decision-making accuracy across 10 epochs. \textbf{b} presents the Examination Recommendation F1-score. The shaded regions indicate the 95\% confidence interval (CI). The results demonstrate that the model converges rapidly, with performance hitting a bottleneck and plateauing after the 3rd epoch, indicating that simply increasing training steps in SFT does not lead to continuous improvement.}
    \label{fig:flops_ablation}
\end{figure}

\subsection{Selected Examinations Collection}
\label{sec:selected_examinations_collection}
To ensure the reliability of the distributional analysis, we considered only examinations with relatively high occurrence frequencies.
For numerical-type examinations, only those with more than 500 occurrences were retained, resulting in 11 examinations comprising a total of 24 subevents. The distributional metrics were computed based on these 24 subevents, as detailed in Supplementary Table~\ref{tab:selected_lab_events}.
For textual-type examinations, only radiology examinations with more than 100 occurrences were considered, yielding five types: ``LIVER OR GALLBLADDER US (SINGLE ORGAN)'', ``CT HEAD W/O CONTRAST'', ``CT HEAD W/O CONTRAST Q111 CT HEAD'', ``CHEST (PA AND LAT)'', and ``CHEST (PORTABLE AP)''.

\begin{table}[t]
\centering
\caption{Selected Laboratory Events and Their Corresponding Sub-events}
\label{tab:selected_lab_events}
\begin{tabularx}{\textwidth}{>{\RaggedRight\arraybackslash}p{0.25\textwidth} >{\RaggedRight\arraybackslash}X}
\toprule
\textbf{Event} & \textbf{Sub-events} \\
\midrule
Complete Blood Count & MCH; White Blood Cells; Absolute Basophil Count; Absolute Monocyte Count; Absolute Eosinophil Count; Monocytes; Platelet Count; Hemoglobin; Hematocrit; Neutrophils; Absolute Neutrophil Count; RDW; Basophils; Eosinophils; Red Blood Cells; RDW-SD; Lymphocytes \\
Mean Corpuscular Volume & MCV \\
Liver Function Test & PT; Alanine Aminotransferase (ALT); Asparate Aminotransferase (AST); Albumin; Alkaline Phosphatase \\
Anion Gap & Anion Gap \\
Comprehensive Metabolic Panel & Bicarbonate \\
Total Bilirubin & Bilirubin, Total \\
Total Calcium & Calcium, Total \\
Kidney Function Tests & Urea Nitrogen; Glucose; Phosphate; Creatinine; Sodium; Potassium; Chloride \\
Lactate Dehydrogenase & Lactate Dehydrogenase (LD) \\
Magnesium & Magnesium \\
Coagulation Profile & PTT \\
Lactate & Lactate \\
Urine Analysis & RBC; pH; WBC \\
Total Calculated CO2 & Calculated Total CO2 \\
pCO2 & pCO2 \\
pH & pH \\
pO2 & pO2 \\
Lipase & Lipase \\
Creatine Kinase & Creatine Kinase (CK) \\
Creatine Kinase, MB Isoenzyme & Creatine Kinase, MB Isoenzyme \\
H & H \\
I & I \\
L & L \\
Thyroid Function Test & Thyroid Stimulating Hormone \\
\bottomrule
\end{tabularx}
\end{table}

\subsection{Detailed Results on DiagBench}

\subsubsection{Results on MIMIC-IV}

\begin{table}[!h]
\renewcommand{\arraystretch}{1.3} 
\footnotesize
\centering
\caption{Single-turn evaluation on the \textbf{MIMIC-IV} dataset. We report the \textbf{Hit Ratio} for examination recommendation and \textbf{Accuracy} for final diagnosis, with \textbf{95\% Confidence Intervals} in brackets. Agentic systems use DeepSeek-v3 as their base model.}
\label{tab:diagnoser_result_under_static}
\resizebox{.6\textwidth}{!}{
\begin{tabular}{c|cc|cc}
\toprule
Model & Size & Year & Hit Ratio(\%) & Diagnosis Accuracy(\%) \\
\midrule
\rowcolor{mygray} \multicolumn{5}{c}{Basic LLM} \\
\midrule
GPT-4o          & -    & 2024.8 & \makecell{20.21 \\ \scriptsize[18.98, 21.39]} & \makecell{72.13 \\ \scriptsize[69.07, 75.33]} \\
Claude-4-sonnet & -  & 2024.6 & \makecell{26.40 \\ \scriptsize[24.95, 27.66]} & \makecell{76.80 \\ \scriptsize[73.60, 79.73]} \\
\midrule
Qwen2.5         & 72B  & 2024.9 & \makecell{19.20 \\ \scriptsize[17.94, 20.43]} & \makecell{74.80 \\ \scriptsize[71.87, 77.73]} \\
Llama3.3        & 70B  & 2024.12& \makecell{19.97 \\ \scriptsize[18.77, 21.20]} & \makecell{66.27 \\ \scriptsize[63.07, 69.47]} \\
DeepSeek-v3     & 671B & 2024.12& \makecell{20.08 \\ \scriptsize[18.77, 21.37]} & \makecell{72.27 \\ \scriptsize[69.07, 75.60]} \\
Qwen2.5         & 235B & -      & \makecell{21.39 \\ \scriptsize[20.19, 22.70]} & \makecell{72.40 \\ \scriptsize[69.46, 75.47]} \\
GPT-OSS         & 120B & -      & \makecell{15.37 \\ \scriptsize[14.27, 16.60]} & \makecell{68.27 \\ \scriptsize[65.06, 71.73]} \\
\midrule
OpenBioLLM      & 70B  & 2024.4 & \makecell{23.53 \\ \scriptsize[22.22, 24.79]} & \makecell{66.27 \\ \scriptsize[62.93, 69.60]} \\
Baichuan-M1     & 14B  & 2025.2 & \makecell{19.60 \\ \scriptsize[18.23, 20.88]} & \makecell{78.93 \\ \scriptsize[76.00, 81.60]} \\
MedGemma        & 27B  & 2025.7 & \makecell{28.57 \\ \scriptsize[27.23, 30.07]} & \makecell{70.53 \\ \scriptsize[67.20, 73.87]} \\
Baichuan-M3     & 235B    & 2026.1 & \makecell{22.36 \\ \scriptsize[21.15, 23.67]} & \makecell{67.73 \\ \scriptsize[64.27, 70.80]} \\
\midrule
\rowcolor{mygray} \multicolumn{5}{c}{Agentic System} \\
\midrule
MedAgents       & -    & 2024.1 & \makecell{14.08 \\ \scriptsize[13.01, 15.21]} & \makecell{70.27 \\ \scriptsize[67.07, 73.47]} \\
MDAgents        & -    & 2024.10& \makecell{12.64 \\ \scriptsize[11.57, 13.63]} & \makecell{73.47 \\ \scriptsize[70.26, 76.40]} \\
\midrule
\rowcolor{mygray} \multicolumn{5}{c}{Our Method (DiagAgent)} \\
\midrule
\multirow{3}{*}{DiagAgent} & 7B   & - & \makecell{\textbf{72.56} \\ \scriptsize[71.19, 74.00]} & \makecell{85.60 \\ \scriptsize[82.93, 88.14]} \\ 
                           & 8B   & - & \makecell{56.57 \\ \scriptsize[55.02, 58.13]} & \makecell{82.27 \\ \scriptsize[79.60, 84.80]} \\
                           & 14B  & - & \makecell{68.49 \\ \scriptsize[67.07, 69.91]} & \makecell{\textbf{87.87} \\ \scriptsize[85.47, 90.13]} \\
\bottomrule
\end{tabular}
}
\end{table}

\begin{table}[!h]
\renewcommand{\arraystretch}{1.3} 
\footnotesize
\centering
\caption{End-to-end evaluation on the \textbf{MIMIC-IV} dataset. Metrics include average conversation turns, precision, recall, F1-score for examination recommendation, and final diagnostic accuracy. All metrics are reported with \textbf{95\% Confidence Intervals}}
\label{tab:diagnoser_result_under_ehrgenerator_env}
\resizebox{\textwidth}{!}{
\begin{tabular}{c|cc|c|ccc|c}
\toprule
Model & Size & Year & Avg. Turns & Precision & Recall & F1 & Accuracy(\%) \\
\midrule
\rowcolor{mygray} \multicolumn{8}{c}{Basic LLM} \\
\midrule
GPT-4o      & -   & 2024.8 & 3.30 & 
\makecell{31.08 \\ \scriptsize[28.22, 33.76]} & 
\makecell{14.76 \\ \scriptsize[13.48, 16.13]} & 
\makecell{16.96 \\ \scriptsize[15.37, 18.44]} & 
\makecell{43.20 \\ \scriptsize[39.73, 46.67]} \\

Claude-4-Sonnet & - & 2024.6 & 3.91 & 
\makecell{37.10 \\ \scriptsize[34.42, 39.67]} & 
\makecell{25.02 \\ \scriptsize[23.25, 26.78]} & 
\makecell{26.32 \\ \scriptsize[24.62, 28.03]} & 
\makecell{51.47 \\ \scriptsize[47.86, 55.07]} \\
\midrule
Qwen2.5     & 72B & 2024.9 & 2.47 & 
\makecell{35.19 \\ \scriptsize[31.98, 38.59]} & 
\makecell{12.13 \\ \scriptsize[10.93, 13.39]} & 
\makecell{16.23 \\ \scriptsize[14.66, 17.78]} & 
\makecell{44.27 \\ \scriptsize[40.67, 47.73]} \\

Llama3.3    & 70B & 2024.12& 4.25 & 
\makecell{28.99 \\ \scriptsize[27.08, 31.07]} & 
\makecell{22.33 \\ \scriptsize[20.66, 24.11]} & 
\makecell{23.01 \\ \scriptsize[21.52, 24.66]} & 
\makecell{38.53 \\ \scriptsize[34.93, 42.00]} \\

DeepSeek-v3 & 671B& 2024.12& 2.49 & 
\makecell{35.09 \\ \scriptsize[32.03, 37.94]} & 
\makecell{13.09 \\ \scriptsize[11.74, 14.50]} & 
\makecell{16.78 \\ \scriptsize[15.19, 18.48]} & 
\makecell{47.07 \\ \scriptsize[43.60, 50.53]} \\

Qwen2.5     & 235B& -      & 3.34 & 
\makecell{30.14 \\ \scriptsize[27.63, 32.63]} & 
\makecell{18.60 \\ \scriptsize[17.08, 20.20]} & 
\makecell{19.74 \\ \scriptsize[18.23, 21.30]} & 
\makecell{45.33 \\ \scriptsize[42.00, 49.34]} \\

GPT-OSS     & 120B& -      & 4.08 & 
\makecell{27.82 \\ \scriptsize[25.11, 30.48]} & 
\makecell{16.54 \\ \scriptsize[15.00, 18.02]} & 
\makecell{16.72 \\ \scriptsize[15.22, 18.26]} & 
\makecell{46.53 \\ \scriptsize[43.20, 49.87]} \\
\midrule
OpenBioLLM  & 70B & 2024.4 & 2.59 & 
\makecell{32.80 \\ \scriptsize[29.92, 35.64]} & 
\makecell{14.02 \\ \scriptsize[12.63, 15.45]} & 
\makecell{17.73 \\ \scriptsize[16.07, 19.49]} & 
\makecell{34.27 \\ \scriptsize[30.93, 37.87]} \\

Baichuan-M1 & 14B & 2025.2 & 2.30 & 
\makecell{26.05 \\ \scriptsize[23.13, 28.94]} & 
\makecell{9.66 \\ \scriptsize[8.50, 10.90]} & 
\makecell{12.55 \\ \scriptsize[11.16, 13.97]} & 
\makecell{33.33 \\ \scriptsize[30.13, 37.07]} \\

MedGemma    & 27B & 2024.7 & 4.10 & 
\makecell{35.20 \\ \scriptsize[32.73, 37.79]} & 
\makecell{21.17 \\ \scriptsize[19.44, 22.90]} & 
\makecell{22.44 \\ \scriptsize[20.76, 24.09]} & 
\makecell{44.27 \\ \scriptsize[40.53, 48.00]} \\

Baichuan-M3 & 235B & 2026.1 & 4.83 & 
\makecell{29.15 \\ \scriptsize[26.52, 31.83]} & 
\makecell{21.88 \\ \scriptsize[20.03, 23.70]} & 
\makecell{19.83 \\ \scriptsize[18.35, 21.48]} & 
\makecell{40.40 \\ \scriptsize[36.93, 43.73]} \\

\midrule
\rowcolor{mygray} \multicolumn{8}{c}{Agentic System} \\
\midrule
MedAgents   & -   & 2024.1 & 2.31 & 
\makecell{30.59 \\ \scriptsize[27.37, 33.76]} & 
\makecell{11.37 \\ \scriptsize[10.13, 12.68]} & 
\makecell{14.76 \\ \scriptsize[13.09, 16.39]} & 
\makecell{45.60 \\ \scriptsize[42.00, 49.20]} \\

MDAgents    & -   & 2024.10& 2.40 & 
\makecell{32.18 \\ \scriptsize[29.23, 35.40]} & 
\makecell{11.90 \\ \scriptsize[10.82, 13.13]} & 
\makecell{15.37 \\ \scriptsize[13.78, 17.02]} & 
\makecell{45.07 \\ \scriptsize[41.47, 48.53]} \\
\midrule
\rowcolor{mygray} \multicolumn{8}{c}{Our Method (DiagAgent)} \\
\midrule
\multirow{3}{*}{DiagAgent} & 7B &-& 5.45 & 
\makecell{\textbf{46.02} \\ \scriptsize[44.04, 47.97]} & 
\makecell{47.33 \\ \scriptsize[45.40, 49.21]} & 
\makecell{\textbf{43.90} \\ \scriptsize[42.21, 45.58]} & 
\makecell{61.47 \\ \scriptsize[58.00, 65.20]}\\ 

                           & 8B &-& 5.73 & 
\makecell{39.57 \\ \scriptsize[37.66, 41.35]} & 
\makecell{43.13 \\ \scriptsize[41.10, 44.95]} & 
\makecell{38.56 \\ \scriptsize[36.83, 40.34]} & 
\makecell{53.33 \\ \scriptsize[49.60, 56.80]} \\

                           & 14B&-& 6.66 & 
\makecell{42.04 \\ \scriptsize[40.19, 43.69]} & 
\makecell{\textbf{52.14} \\ \scriptsize[50.14, 54.05]} & 
\makecell{43.72 \\ \scriptsize[42.07, 45.41]} & 
\makecell{\textbf{62.67} \\ \scriptsize[59.07, 66.13]}  \\
\bottomrule
\end{tabular}
}
\end{table}

\clearpage
\subsubsection{Results on PMC-OA}

\begin{table}[!h]
\renewcommand{\arraystretch}{1.3} 
\footnotesize
\centering
\caption{Single-turn evaluation on the \textbf{PMC-OA} dataset. We report \textbf{Hit Ratio} for examination recommendation and \textbf{Accuracy} for final diagnosis, with \textbf{95\% Confidence Intervals}.}
\label{tab:diagnoser_result_pmcoa_single_turn}
\resizebox{0.75\textwidth}{!}{
\begin{tabular}{c|cc|cc}
\toprule
Model & Size & Year & Hit Ratio(\%) & Diagnosis Accuracy(\%) \\
\midrule
\rowcolor{mygray} \multicolumn{5}{c}{Basic LLM} \\
\midrule
GPT-4o          & -    & 2024.8 & \makecell{51.03 \\ \scriptsize[46.91, 54.68]} & \makecell{84.15 \\ \scriptsize[81.14, 86.85]} \\
Claude-4-sonnet & -  & 2024.6 & \makecell{46.12 \\ \scriptsize[42.15, 49.77]} & \makecell{84.63 \\ \scriptsize[81.93, 87.48]} \\
\midrule
Qwen2.5         & 72B  & 2024.9 & \makecell{50.08 \\ \scriptsize[46.11, 54.05]} & \makecell{82.88 \\ \scriptsize[80.03, 85.90]} \\
Llama3.3        & 70B  & 2024.12& \makecell{51.82 \\ \scriptsize[48.02, 55.63]} & \makecell{77.18 \\ \scriptsize[73.69, 80.51]} \\
DeepSeek-v3     & 671B & 2024.12& \makecell{47.86 \\ \scriptsize[44.06, 51.66]} & \makecell{83.84 \\ \scriptsize[80.98, 86.69]} \\
Qwen2.5         & 235B & -      & \makecell{48.65 \\ \scriptsize[44.69, 52.61]} & \makecell{83.68 \\ \scriptsize[80.51, 86.37]} \\
GPT-OSS         & 120B & -      & \makecell{43.42 \\ \scriptsize[39.62, 47.39]} & \makecell{79.08 \\ \scriptsize[76.22, 82.09]} \\
\midrule
OpenBioLLM      & 70B  & 2024.4 & \makecell{48.18 \\ \scriptsize[44.22, 51.98]} & \makecell{72.42 \\ \scriptsize[68.94, 75.59]} \\
Baichuan-M1     & 14B  & 2025.2 & \makecell{43.26 \\ \scriptsize[39.30, 47.23]} & \makecell{88.59 \\ \scriptsize[86.05, 90.97]} \\
MedGemma        & 27B  & 2024.7 & \makecell{48.81 \\ \scriptsize[44.85, 52.93]} & \makecell{83.20 \\ \scriptsize[80.35, 86.05]} \\
Baichuan-M3     & 235B & 2026.1 & \makecell{48.02 \\ \scriptsize[44.06, 51.66]} & \makecell{81.14 \\ \scriptsize[78.13, 84.31]} \\
\midrule
\rowcolor{mygray} \multicolumn{5}{c}{Agentic System} \\
\midrule
MedAgents       & -    & 2024.1 & \makecell{43.26 \\ \scriptsize[39.30, 47.39]} & \makecell{81.77 \\ \scriptsize[78.92, 84.63]} \\
MDAgents        & -    & 2024.10& \makecell{48.18 \\ \scriptsize[44.22, 52.14]} & \makecell{84.15 \\ \scriptsize[81.30, 87.00]} \\
\midrule
\rowcolor{mygray} \multicolumn{5}{c}{Our Method (DiagAgent)} \\
\midrule
\multirow{1}{*}{DiagAgent} & 14B & \multirow{1}{*}{-} & \makecell{\textbf{51.98} \\ \scriptsize[47.86, 55.94]} & \makecell{\textbf{88.91} \\ \scriptsize[86.37, 91.28]} \\
\bottomrule
\end{tabular}
}
\end{table}

\begin{table}[!h]
\renewcommand{\arraystretch}{1.3} 
\footnotesize
\centering
\caption{End-to-end evaluation on the \textbf{PMCOA} dataset. We report Precision, Recall, F1-score for examination recommendation, and final diagnostic accuracy with \textbf{95\% Confidence Intervals}.}
\label{tab:diagnoser_result_pmcoa}
\resizebox{0.95\textwidth}{!}{
\begin{tabular}{c|cc|c|ccc|c}
\toprule
Model & Size & Year & Avg. Turns & Precision & Recall & F1 & Accuracy(\%) \\
\midrule
\rowcolor{mygray} \multicolumn{8}{c}{Basic LLM} \\
\midrule
GPT-4o      & -   & 2024.8 & 3.04 & 
\makecell{37.32 \\ \scriptsize[33.9-40.7]} & 
\makecell{22.16 \\ \scriptsize[19.9-24.3]} & 
\makecell{24.20 \\ \scriptsize[21.8-26.4]} & 
\makecell{48.81 \\ \scriptsize[45.0-52.8]} \\

Claude-4-sonnet & - & 2024.6 & 4.17 & 
\makecell{34.70 \\ \scriptsize[32.0-37.5]} & 
\makecell{30.03 \\ \scriptsize[27.7-32.5]} & 
\makecell{28.05 \\ \scriptsize[26.0-30.2]} & 
\makecell{49.13 \\ \scriptsize[45.0-52.9]} \\
\midrule
Qwen2.5     & 72B & 2024.9 & 2.59 & 
\makecell{32.44 \\ \scriptsize[29.3-36.0]} & 
\makecell{16.97 \\ \scriptsize[15.1-18.9]} & 
\makecell{20.19 \\ \scriptsize[18.2-22.2]} & 
\makecell{39.14 \\ \scriptsize[35.3-42.8]} \\

Llama3.3    & 70B & 2024.12& 4.09 & 
\makecell{31.10 \\ \scriptsize[28.8-33.3]} & 
\makecell{31.46 \\ \scriptsize[28.9-33.9]} & 
\makecell{28.22 \\ \scriptsize[26.1-30.4]} & 
\makecell{45.48 \\ \scriptsize[41.7-49.1]} \\

DeepSeek-v3 & 671B& 2024.12& 2.52 & 
\makecell{37.63 \\ \scriptsize[34.3-41.0]} & 
\makecell{19.61 \\ \scriptsize[17.6-21.8]} & 
\makecell{23.46 \\ \scriptsize[21.3-25.9]} & 
\makecell{43.11 \\ \scriptsize[39.5-46.9]} \\

Qwen2.5     & 235B& -      & 3.45 & 
\makecell{34.96 \\ \scriptsize[32.2-37.9]} & 
\makecell{28.56 \\ \scriptsize[26.2-31.0]} & 
\makecell{27.43 \\ \scriptsize[25.2-29.7]} & 
\makecell{49.45 \\ \scriptsize[45.6-53.3]} \\

GPT-OSS     & 120B& -      & 4.30 & 
\makecell{28.19 \\ \scriptsize[25.4-31.0]} & 
\makecell{25.21 \\ \scriptsize[22.8-27.6]} & 
\makecell{22.20 \\ \scriptsize[20.1-24.4]} & 
\makecell{45.96 \\ \scriptsize[42.2-49.9]} \\
\midrule
OpenBioLLM  & 70B & 2024.4 & 2.49 & 
\makecell{31.03 \\ \scriptsize[27.9-34.6]} & 
\makecell{16.38 \\ \scriptsize[14.7-18.2]} & 
\makecell{19.23 \\ \scriptsize[17.2-21.4]} & 
\makecell{35.18 \\ \scriptsize[31.7-38.7]} \\

Baichuan-M1 & 14B & 2025.2 & 2.62 & 
\makecell{20.59 \\ \scriptsize[17.7-23.3]} & 
\makecell{11.89 \\ \scriptsize[10.1-13.6]} & 
\makecell{13.82 \\ \scriptsize[11.9-15.9]} & 
\makecell{39.46 \\ \scriptsize[35.8-43.3]} \\

MedGemma    & 27B & 2024.7 & 4.04 & 
\makecell{32.64 \\ \scriptsize[29.7-35.5]} & 
\makecell{26.79 \\ \scriptsize[24.4-29.2]} & 
\makecell{25.47 \\ \scriptsize[23.4-27.7]} & 
\makecell{44.37 \\ \scriptsize[40.7-48.3]} \\

Baichuan-M3 & 235B& 2026.1 & 4.40 & 
\makecell{29.42 \\ \scriptsize[26.5-32.4]} & 
\makecell{27.22 \\ \scriptsize[24.9-29.6]} & 
\makecell{23.38 \\ \scriptsize[21.4-25.7]} & 
\makecell{42.31 \\ \scriptsize[38.5-46.1]} \\

\midrule
\rowcolor{mygray} \multicolumn{8}{c}{Agentic System} \\
\midrule
MedAgents   & -   & 2024.1 & 2.26 & 
\makecell{33.26 \\ \scriptsize[29.9-36.6]} & 
\makecell{16.99 \\ \scriptsize[15.2-19.0]} & 
\makecell{20.31 \\ \scriptsize[18.1-22.5]} & 
\makecell{45.80 \\ \scriptsize[41.8-49.8]} \\

MDAgents    & -   & 2024.10& 2.40 & 
\makecell{\textbf{38.11} \\ \scriptsize[34.6-41.5]} & 
\makecell{18.75 \\ \scriptsize[16.7-20.7]} & 
\makecell{22.52 \\ \scriptsize[20.3-24.8]} & 
\makecell{45.48 \\ \scriptsize[41.7-49.6]} \\
\midrule
\rowcolor{mygray} \multicolumn{8}{c}{Our Method} \\
\midrule
DiagAgent   & 14B & -      & 6.88 & 
\makecell{24.33 \\ \scriptsize[22.9-25.9]} & 
\makecell{\textbf{46.21} \\ \scriptsize[43.7-49.0]} & 
\makecell{\textbf{29.72} \\ \scriptsize[27.9-31.5]} & 
\makecell{\textbf{53.88} \\ \scriptsize[50.2-58.0]} \\
\bottomrule
\end{tabular}
}
\end{table}

\clearpage
\subsubsection{Results on MTSamples}

\begin{table}[!h]
\renewcommand{\arraystretch}{1.3} 
\footnotesize
\centering
\caption{Single-turn evaluation on the \textbf{MTSamples} dataset. We report \textbf{Hit Ratio} for examination recommendation and \textbf{Accuracy} for final diagnosis, with \textbf{95\% Confidence Intervals}. }
\label{tab:diagnoser_result_mtsamples_single_turn}
\resizebox{0.75\textwidth}{!}{
\begin{tabular}{c|cc|cc}
\toprule
Model & Size & Year & Hit Ratio(\%) & Diagnosis Accuracy(\%) \\
\midrule
\rowcolor{mygray} \multicolumn{5}{c}{Basic LLM} \\
\midrule
GPT-4o          & -    & 2024.8 & \makecell{49.08 \\ \scriptsize[44.33, 54.35]} & \makecell{86.02 \\ \scriptsize[82.32, 89.19]} \\
Claude-4-sonnet & -  & 2024.6 & \makecell{43.54 \\ \scriptsize[38.52, 48.55]} & \makecell{84.43 \\ \scriptsize[81.00, 87.86]} \\
\midrule
Qwen2.5         & 72B  & 2024.9 & \makecell{53.03 \\ \scriptsize[48.02, 58.05]} & \makecell{86.81 \\ \scriptsize[83.38, 89.97]} \\
Llama3.3        & 70B  & 2024.12& \makecell{49.60 \\ \scriptsize[44.59, 54.62]} & \makecell{75.99 \\ \scriptsize[71.50, 80.21]} \\
DeepSeek-v3     & 671B & 2024.12& \makecell{44.33 \\ \scriptsize[39.05, 49.60]} & \makecell{83.64 \\ \scriptsize[79.42, 87.34]} \\
Qwen2.5         & 235B & -      & \makecell{47.76 \\ \scriptsize[42.48, 52.77]} & \makecell{83.38 \\ \scriptsize[79.42, 86.81]} \\
GPT-OSS         & 120B & -      & \makecell{44.85 \\ \scriptsize[39.84, 49.60]} & \makecell{75.73 \\ \scriptsize[70.98, 79.95]} \\
\midrule
OpenBioLLM      & 70B  & 2024.4 & \makecell{46.97 \\ \scriptsize[41.69, 51.72]} & \makecell{82.06 \\ \scriptsize[78.10, 86.02]} \\
Baichuan-M1     & 14B  & 2025.2 & \makecell{44.33 \\ \scriptsize[39.05, 49.08]} & \makecell{90.24 \\ \scriptsize[87.07, 92.88]} \\
MedGemma        & 27B  & 2024.7 & \makecell{50.40 \\ \scriptsize[45.38, 55.67]} & \makecell{83.91 \\ \scriptsize[79.95, 87.34]} \\
Baichuan-M3     & 235B & 2026.1 & \makecell{47.23 \\ \scriptsize[42.22, 52.51]} & \makecell{79.16 \\ \scriptsize[74.93, 83.11]} \\
\midrule
\rowcolor{mygray} \multicolumn{5}{c}{Agentic System} \\
\midrule
MedAgents       & -    & 2024.1 & \makecell{41.42 \\ \scriptsize[36.41, 45.91]} & \makecell{79.16 \\ \scriptsize[74.67, 83.11]} \\
MDAgents        & -    & 2024.10& \makecell{51.45 \\ \scriptsize[46.70, 56.20]} & \makecell{81.00 \\ \scriptsize[77.31, 84.70]} \\
\midrule
\rowcolor{mygray} \multicolumn{5}{c}{Our Method (DiagAgent)} \\
\midrule
\multirow{1}{*}{DiagAgent} & 14B & \multirow{1}{*}{-} & \makecell{\textbf{60.16} \\ \scriptsize[55.15, 65.17]} & \makecell{\textbf{93.67} \\ \scriptsize[91.29, 95.78]} \\
\bottomrule
\end{tabular}
}
\end{table}

\begin{table}[!h]
\renewcommand{\arraystretch}{1.3} 
\footnotesize
\centering
\caption{End-to-end evaluation on the \textbf{MTSamples} dataset. We report Precision, Recall, F1-score for examination recommendation, and final diagnostic accuracy with \textbf{95\% Confidence Intervals}.}
\label{tab:diagnoser_result_mtsamples}
\resizebox{0.95\textwidth}{!}{
\begin{tabular}{c|cc|c|ccc|c}
\toprule
Model & Size & Year & Avg. Turns & Precision & Recall & F1 & Accuracy(\%) \\
\midrule
\rowcolor{mygray} \multicolumn{8}{c}{Basic LLM} \\
\midrule
GPT-4o      & -   & 2024.8 & 3.13 & 
\makecell{41.60 \\ \scriptsize[37.13-46.00]} & 
\makecell{27.55 \\ \scriptsize[24.85-30.50]} & 
\makecell{30.34 \\ \scriptsize[27.64-33.47]} & 
\makecell{47.23 \\ \scriptsize[42.48-52.24]} \\

Claude-4-sonnet & - & 2024.6 & 3.54 & 
\makecell{43.71 \\ \scriptsize[39.92-47.56]} & 
\makecell{37.14 \\ \scriptsize[34.05-40.23]} & 
\makecell{36.59 \\ \scriptsize[33.63-39.57]} & 
\makecell{57.52 \\ \scriptsize[52.51-62.53]} \\
\midrule
Qwen2.5     & 72B & 2024.9 & 2.58 & 
\makecell{37.21 \\ \scriptsize[33.44-41.39]} & 
\makecell{22.82 \\ \scriptsize[20.19-25.67]} & 
\makecell{25.85 \\ \scriptsize[22.89-28.91]} & 
\makecell{46.17 \\ \scriptsize[41.16-51.19]} \\

Llama3.3    & 70B & 2024.12& 3.95 & 
\makecell{36.77 \\ \scriptsize[33.98-39.79]} & 
\makecell{39.25 \\ \scriptsize[36.08-42.69]} & 
\makecell{34.92 \\ \scriptsize[31.99-37.95]} & 
\makecell{50.40 \\ \scriptsize[45.12-55.15]} \\

DeepSeek-v3 & 671B& 2024.12& 2.42 & 
\makecell{\textbf{46.09} \\ \scriptsize[41.28-50.73]} & 
\makecell{24.38 \\ \scriptsize[21.78-27.18]} & 
\makecell{29.05 \\ \scriptsize[26.23-32.13]} & 
\makecell{52.24 \\ \scriptsize[47.76-56.99]} \\

Qwen2.5     & 235B& -      & 3.25 & 
\makecell{41.17 \\ \scriptsize[37.38-44.95]} & 
\makecell{33.31 \\ \scriptsize[30.30-36.28]} & 
\makecell{33.50 \\ \scriptsize[30.56-36.62]} & 
\makecell{50.40 \\ \scriptsize[45.12-55.15]} \\

GPT-OSS     & 120B& -      & 3.60 & 
\makecell{38.68 \\ \scriptsize[34.57-42.76]} & 
\makecell{29.74 \\ \scriptsize[26.91-33.06]} & 
\makecell{28.57 \\ \scriptsize[25.65-31.57]} & 
\makecell{52.51 \\ \scriptsize[47.23-57.26]} \\
\midrule
OpenBioLLM  & 70B & 2024.4 & 2.44 & 
\makecell{41.72 \\ \scriptsize[37.28-46.20]} & 
\makecell{24.23 \\ \scriptsize[21.53-26.98]} & 
\makecell{28.75 \\ \scriptsize[25.63-31.78]} & 
\makecell{41.16 \\ \scriptsize[36.15-46.70]} \\

Baichuan-M1 & 14B & 2025.2 & 2.50 & 
\makecell{32.92 \\ \scriptsize[28.53-37.58]} & 
\makecell{18.58 \\ \scriptsize[16.27-21.33]} & 
\makecell{22.08 \\ \scriptsize[19.26-25.03]} & 
\makecell{48.55 \\ \scriptsize[43.80-53.83]} \\

MedGemma    & 27B & 2024.7 & 3.84 & 
\makecell{41.23 \\ \scriptsize[37.62-45.29]} & 
\makecell{32.99 \\ \scriptsize[29.84-35.88]} & 
\makecell{32.42 \\ \scriptsize[29.57-35.57]} & 
\makecell{48.55 \\ \scriptsize[43.27-53.56]} \\

Baichuan-M3 & 235B& 2026.1 & 4.28 & 
\makecell{34.55 \\ \scriptsize[30.59-38.44]} & 
\makecell{32.59 \\ \scriptsize[29.67-35.72]} & 
\makecell{28.02 \\ \scriptsize[25.26-30.54]} & 
\makecell{46.44 \\ \scriptsize[41.69-51.19]} \\
\midrule
\rowcolor{mygray} \multicolumn{8}{c}{Agentic System} \\
\midrule
MedAgents   & -   & 2024.1 & 2.23 & 
\makecell{37.14 \\ \scriptsize[32.56-41.79]} & 
\makecell{19.56 \\ \scriptsize[16.91-22.31]} & 
\makecell{23.10 \\ \scriptsize[20.20-26.22]} & 
\makecell{48.28 \\ \scriptsize[43.01-53.04]} \\

MDAgents    & -   & 2024.10& 2.23 & 
\makecell{42.13 \\ \scriptsize[37.20-46.93]} & 
\makecell{21.55 \\ \scriptsize[19.10-24.24]} & 
\makecell{26.01 \\ \scriptsize[23.05-28.86]} & 
\makecell{50.13 \\ \scriptsize[45.12-54.88]} \\
\midrule
\rowcolor{mygray} \multicolumn{8}{c}{Our Method} \\
\midrule
DiagAgent   & 14B & -      & 6.64 & 
\makecell{30.64 \\ \scriptsize[28.90-32.53]} & 
\makecell{\textbf{63.24} \\ \scriptsize[59.71-66.67]} & 
\makecell{\textbf{39.17} \\ \scriptsize[37.02-41.37]} & 
\makecell{\textbf{64.91} \\ \scriptsize[60.16-69.39]} \\
\bottomrule
\end{tabular}
}
\end{table}

\clearpage
\subsubsection{Results on DDXPlus}

\begin{table}[!h]
\renewcommand{\arraystretch}{1.3} 
\footnotesize
\centering
\caption{Single-turn evaluation on the \textbf{DDXPlus} dataset. We report \textbf{Hit Ratio} for examination recommendation and \textbf{Accuracy} for final diagnosis, with \textbf{95\% Confidence Intervals}. }
\label{tab:diagnoser_result_ddxplus_single_turn}
\resizebox{0.75\textwidth}{!}{
\begin{tabular}{c|cc|cc}
\toprule
Model & Size & Year & Hit Ratio(\%) & Diagnosis Accuracy(\%) \\
\midrule
\rowcolor{mygray} \multicolumn{5}{c}{Basic LLM} \\
\midrule
GPT-4o          & -    & 2024.8 & \makecell{55.73 \\ \scriptsize[51.71, 60.16]} & \makecell{91.55 \\ \scriptsize[89.13, 93.96]} \\
Claude-4-sonnet & -  & 2024.6 & \makecell{46.68 \\ \scriptsize[42.25, 51.11]} & \makecell{94.57 \\ \scriptsize[92.35, 96.38]} \\
\midrule
Qwen2.5         & 72B  & 2024.9 & \makecell{61.97 \\ \scriptsize[58.14, 66.20]} & \makecell{93.56 \\ \scriptsize[91.35, 95.57]} \\
Llama3.3        & 70B  & 2024.12& \makecell{70.22 \\ \scriptsize[66.20, 74.04]} & \makecell{87.53 \\ \scriptsize[84.51, 90.54]} \\
DeepSeek-v3     & 671B & 2024.12& \makecell{47.69 \\ \scriptsize[43.26, 52.52]} & \makecell{91.15 \\ \scriptsize[88.53, 93.56]} \\
Qwen2.5         & 235B & -      & \makecell{48.89 \\ \scriptsize[44.66, 53.12]} & \makecell{93.36 \\ \scriptsize[91.15, 95.57]} \\
GPT-OSS         & 120B & -      & \makecell{47.28 \\ \scriptsize[42.86, 51.71]} & \makecell{80.68 \\ \scriptsize[77.26, 84.31]} \\
\midrule
OpenBioLLM      & 70B  & 2024.4 & \makecell{53.52 \\ \scriptsize[49.30, 57.95]} & \makecell{92.76 \\ \scriptsize[90.54, 94.97]} \\
Baichuan-M1     & 14B  & 2025.2 & \makecell{42.45 \\ \scriptsize[37.83, 46.68]} & \makecell{93.96 \\ \scriptsize[91.55, 95.98]} \\
MedGemma        & 27B  & 2024.7 & \makecell{50.10 \\ \scriptsize[45.67, 54.53]} & \makecell{92.96 \\ \scriptsize[90.74, 94.97]} \\
Baichuan-M3     & 235B & 2026.1 & \makecell{51.11 \\ \scriptsize[46.68, 55.14]} & \makecell{89.94 \\ \scriptsize[87.32, 92.56]} \\
\midrule
\rowcolor{mygray} \multicolumn{5}{c}{Agentic System} \\
\midrule
MedAgents       & -    & 2024.1 & \makecell{44.47 \\ \scriptsize[40.44, 48.70]} & \makecell{87.73 \\ \scriptsize[84.71, 90.34]} \\
MDAgents        & -    & 2024.10& \makecell{49.30 \\ \scriptsize[45.07, 53.92]} & \makecell{91.35 \\ \scriptsize[88.73, 93.57]} \\
\midrule
\rowcolor{mygray} \multicolumn{5}{c}{Our Method (DiagAgent)} \\
\midrule
\multirow{1}{*}{DiagAgent} & 14B & \multirow{1}{*}{-} & \makecell{\textbf{86.12} \\ \scriptsize[83.10, 88.93]} & \makecell{\textbf{96.38} \\ \scriptsize[94.77, 97.79]} \\
\bottomrule
\end{tabular}
}
\end{table}

\begin{table}[!h]
\renewcommand{\arraystretch}{1.3} 
\footnotesize
\centering
\caption{End-to-end evaluation on the \textbf{DDXPlus} dataset. We report Precision, Recall, F1-score for examination recommendation, and final diagnostic accuracy with \textbf{95\% Confidence Intervals}. }
\label{tab:diagnoser_result_ddxplus}
\resizebox{0.95\textwidth}{!}{
\begin{tabular}{c|cc|c|ccc|c}
\toprule
Model & Size & Year & Avg. Turns & Precision & Recall & F1 & Accuracy(\%) \\
\midrule
\rowcolor{mygray} \multicolumn{8}{c}{Basic LLM} \\
\midrule
GPT-4o      & -   & 2024.8 & 3.23 & 
\makecell{36.95 \\ \scriptsize[33.38-40.31]} & 
\makecell{33.68 \\ \scriptsize[30.41-36.77]} & 
\makecell{31.29 \\ \scriptsize[28.55-34.20]} & 
\makecell{66.80 \\ \scriptsize[62.98-71.03]} \\

Claude-4-sonnet & - & 2024.6 & 3.28 & 
\makecell{36.00 \\ \scriptsize[32.82-39.45]} & 
\makecell{33.30 \\ \scriptsize[30.21-36.62]} & 
\makecell{31.78 \\ \scriptsize[28.93-34.68]} & 
\makecell{76.06 \\ \scriptsize[72.43-79.68]} \\
\midrule
Qwen2.5     & 72B & 2024.9 & 2.58 & 
\makecell{37.32 \\ \scriptsize[33.60-41.20]} & 
\makecell{27.04 \\ \scriptsize[24.14-29.96]} & 
\makecell{29.04 \\ \scriptsize[26.04-32.03]} & 
\makecell{66.60 \\ \scriptsize[62.58-70.62]} \\

Llama3.3    & 70B & 2024.12& 4.00 & 
\makecell{36.64 \\ \scriptsize[34.05-39.10]} & 
\makecell{48.48 \\ \scriptsize[45.28-51.42]} & 
\makecell{\textbf{38.67} \\ \scriptsize[36.37-41.13]} & 
\makecell{63.78 \\ \scriptsize[59.76-68.01]} \\

DeepSeek-v3 & 671B& 2024.12& 2.45 & 
\makecell{37.32 \\ \scriptsize[33.64-41.38]} & 
\makecell{25.18 \\ \scriptsize[22.23-28.07]} & 
\makecell{27.49 \\ \scriptsize[24.65-30.38]} & 
\makecell{67.61 \\ \scriptsize[63.78-71.83]} \\

Qwen2.5     & 235B& -      & 3.31 & 
\makecell{35.30 \\ \scriptsize[32.07-38.58]} & 
\makecell{37.16 \\ \scriptsize[33.73-40.31]} & 
\makecell{33.49 \\ \scriptsize[30.65-36.17]} & 
\makecell{68.41 \\ \scriptsize[64.38-72.03]} \\

GPT-OSS     & 120B& -      & 3.62 & 
\makecell{27.19 \\ \scriptsize[23.75-30.43]} & 
\makecell{27.75 \\ \scriptsize[24.83-30.51]} & 
\makecell{23.10 \\ \scriptsize[20.71-25.78]} & 
\makecell{67.00 \\ \scriptsize[62.78-71.23]} \\
\midrule
OpenBioLLM  & 70B & 2024.4 & 2.50 & 
\makecell{30.72 \\ \scriptsize[27.35-34.14]} & 
\makecell{22.11 \\ \scriptsize[19.49-24.52]} & 
\makecell{23.46 \\ \scriptsize[20.82-25.96]} & 
\makecell{53.72 \\ \scriptsize[49.30-57.95]} \\

Baichuan-M1 & 14B & 2025.2 & 2.43 & 
\makecell{18.13 \\ \scriptsize[14.85-21.78]} & 
\makecell{11.37 \\ \scriptsize[9.36-13.58]} & 
\makecell{12.68 \\ \scriptsize[10.51-15.03]} & 
\makecell{63.98 \\ \scriptsize[59.36-68.21]} \\

MedGemma    & 27B & 2024.7 & 3.69 & 
\makecell{34.69 \\ \scriptsize[31.49-38.03]} & 
\makecell{39.31 \\ \scriptsize[35.93-42.78]} & 
\makecell{32.81 \\ \scriptsize[29.77-35.79]} & 
\makecell{66.20 \\ \scriptsize[61.97-70.23]} \\

Baichuan-M3 & 235B& 2026.1 & 3.86 & 
\makecell{30.78 \\ \scriptsize[27.66-34.17]} & 
\makecell{32.53 \\ \scriptsize[29.43-35.54]} & 
\makecell{26.99 \\ \scriptsize[24.39-29.64]} & 
\makecell{67.61 \\ \scriptsize[63.38-71.84]} \\
\midrule
\rowcolor{mygray} \multicolumn{8}{c}{Agentic System} \\
\midrule
MedAgents   & -   & 2024.1 & 2.24 & 
\makecell{35.00 \\ \scriptsize[31.26-39.00]} & 
\makecell{21.80 \\ \scriptsize[19.11-24.38]} & 
\makecell{24.44 \\ \scriptsize[21.47-27.26]} & 
\makecell{73.24 \\ \scriptsize[69.41-77.26]} \\

MDAgents    & -   & 2024.10& 2.53 & 
\makecell{\textbf{39.01} \\ \scriptsize[35.24-43.00]} & 
\makecell{27.41 \\ \scriptsize[24.58-30.22]} & 
\makecell{29.77 \\ \scriptsize[26.72-32.80]} & 
\makecell{70.42 \\ \scriptsize[66.40-74.25]} \\
\midrule
\rowcolor{mygray} \multicolumn{8}{c}{Our Method} \\
\midrule
DiagAgent   & 14B & -      & 7.07 & 
\makecell{24.74 \\ \scriptsize[23.1-26.2]} & 
\makecell{\textbf{66.74} \\ \scriptsize[63.6-70.1]} & 
\makecell{34.36 \\ \scriptsize[32.5-36.5]} & 
\makecell{\textbf{76.42} \\ \scriptsize[72.56-80.08]} \\
\bottomrule
\end{tabular}
}
\end{table}

\clearpage
\subsubsection{Results on combined out-of-domain datasets}

\begin{table}[!h]
\renewcommand{\arraystretch}{1.3} 
\footnotesize
\centering
\caption{Single-turn evaluation of out-of-domain performance across combined data sources (PMC-OA, MTSamples, and DDXPlus; $N=1,507$). We report \textbf{Hit Ratio} for examination recommendation and \textbf{Accuracy} for final diagnosis, with \textbf{95\% Confidence Intervals}.}
\label{tab:diagnoser_result_combined_single_turn}
\resizebox{0.75\textwidth}{!}{
\begin{tabular}{c|cc|cc}
\toprule
Model & Size & Year & Hit Ratio(\%) & Diagnosis Accuracy(\%) \\
\midrule
\rowcolor{mygray} \multicolumn{5}{c}{Basic LLM} \\
\midrule
GPT-4o          & -    & 2024.8 & \makecell{52.09 \\ \scriptsize[49.63, 54.48]} & \makecell{87.06 \\ \scriptsize[85.40, 88.72]} \\
Claude-4-sonnet & -  & 2024.6 & \makecell{45.65 \\ \scriptsize[43.00, 48.04]} & \makecell{87.86 \\ \scriptsize[86.13, 89.52]} \\
\midrule
Qwen2.5         & 72B  & 2024.9 & \makecell{54.74 \\ \scriptsize[52.15, 57.13]} & \makecell{87.39 \\ \scriptsize[85.73, 89.25]} \\
Llama3.3        & 70B  & 2024.12& \makecell{57.33 \\ \scriptsize[54.88, 59.99]} & \makecell{80.29 \\ \scriptsize[78.23, 82.22]} \\
DeepSeek-v3     & 671B & 2024.12& \makecell{46.91 \\ \scriptsize[44.39, 49.37]} & \makecell{86.20 \\ \scriptsize[84.41, 87.92]} \\
Qwen2.5         & 235B & -      & \makecell{48.51 \\ \scriptsize[46.05, 51.16]} & \makecell{86.79 \\ \scriptsize[85.00, 88.45]} \\
GPT-OSS         & 120B & -      & \makecell{45.06 \\ \scriptsize[42.60, 47.64]} & \makecell{78.77 \\ \scriptsize[76.71, 80.69]} \\
\midrule
OpenBioLLM      & 70B  & 2024.4 & \makecell{49.64 \\ \scriptsize[47.11, 52.09]} & \makecell{81.55 \\ \scriptsize[79.50, 83.61]} \\
Baichuan-M1     & 14B  & 2025.2 & \makecell{43.26 \\ \scriptsize[40.74, 45.79]} & \makecell{90.78 \\ \scriptsize[89.25, 92.30]} \\
MedGemma        & 27B  & 2024.7 & \makecell{49.64 \\ \scriptsize[47.24, 52.09]} & \makecell{86.60 \\ \scriptsize[84.94, 88.19]} \\
Baichuan-M3     & 235B & 2026.1 & \makecell{48.84 \\ \scriptsize[46.25, 51.43]} & \makecell{83.54 \\ \scriptsize[81.75, 85.40]} \\
\midrule
\rowcolor{mygray} \multicolumn{5}{c}{Agentic System} \\
\midrule
MedAgents       & -    & 2024.1 & \makecell{43.20 \\ \scriptsize[40.68, 45.65]} & \makecell{83.08 \\ \scriptsize[81.09, 85.14]} \\
MDAgents        & -    & 2024.10& \makecell{49.37 \\ \scriptsize[47.05, 51.96]} & \makecell{85.73 \\ \scriptsize[83.94, 87.53]} \\
\midrule
\rowcolor{mygray} \multicolumn{5}{c}{Our Method (DiagAgent)} \\
\midrule
\multirow{1}{*}{DiagAgent} & 14B & \multirow{1}{*}{-} & \makecell{\textbf{65.30} \\ \scriptsize[62.91, 67.69]} & \makecell{\textbf{92.57} \\ \scriptsize[91.24, 93.83]} \\
\bottomrule
\end{tabular}
}
\end{table}

\begin{table}[!h]
\renewcommand{\arraystretch}{1.3} 
\footnotesize
\centering
\caption{End-to-end evaluation of out-of-domain performance across combined data sources (PMC-OA, MTSamples, and DDXPlus; $N=1,507$). We report Precision, Recall, F1-score for examination recommendation, and final diagnostic accuracy with \textbf{95\% Confidence Intervals}. }
\label{tab:diagnoser_result_combined}
\resizebox{0.95\textwidth}{!}{
\begin{tabular}{c|cc|c|ccc|c}
\toprule
Model & Size & Year & Avg. Turns & Precision & Recall & F1 & Accuracy(\%) \\
\midrule
\rowcolor{mygray} \multicolumn{8}{c}{Basic LLM} \\
\midrule
GPT-4o      & -   & 2024.8 & 3.12 & 
\makecell{38.28 \\ \scriptsize[36.25-40.38]} & 
\makecell{27.32 \\ \scriptsize[25.81-28.83]} & 
\makecell{28.08 \\ \scriptsize[26.66-29.54]} & 
\makecell{54.35 \\ \scriptsize[52.09-56.80]} \\

Claude-4-sonnet & - & 2024.6 & 3.72 & 
\makecell{37.39 \\ \scriptsize[35.47-39.42]} & 
\makecell{32.89 \\ \scriptsize[31.18-34.48]} & 
\makecell{31.43 \\ \scriptsize[29.87-33.04]} & 
\makecell{60.12 \\ \scriptsize[57.66-62.51]} \\
\midrule
Qwen2.5     & 72B & 2024.9 & 2.58 & 
\makecell{35.25 \\ \scriptsize[33.15-37.51]} & 
\makecell{21.76 \\ \scriptsize[20.40-23.18]} & 
\makecell{24.53 \\ \scriptsize[23.04-26.15]} & 
\makecell{49.97 \\ \scriptsize[47.31-52.29]} \\

Llama3.3    & 70B & 2024.12& 4.03 & 
\makecell{34.35 \\ \scriptsize[32.83-35.88]} & 
\makecell{39.03 \\ \scriptsize[37.40-40.70]} & 
\makecell{33.35 \\ \scriptsize[31.87-34.79]} & 
\makecell{52.75 \\ \scriptsize[50.17-55.21]} \\

DeepSeek-v3 & 671B& 2024.12& 2.47 & 
\makecell{\textbf{39.66} \\ \scriptsize[37.48-41.57]} & 
\makecell{22.65 \\ \scriptsize[21.17-24.13]} & 
\makecell{26.19 \\ \scriptsize[24.64-27.83]} & 
\makecell{53.48 \\ \scriptsize[50.96-55.74]} \\

Qwen2.5     & 235B& -      & 3.35 & 
\makecell{36.63 \\ \scriptsize[34.72-38.56]} & 
\makecell{32.59 \\ \scriptsize[30.92-34.15]} & 
\makecell{30.96 \\ \scriptsize[29.29-32.50]} & 
\makecell{55.94 \\ \scriptsize[53.55-58.59]} \\

GPT-OSS     & 120B& -      & 3.90 & 
\makecell{30.50 \\ \scriptsize[28.69-32.47]} & 
\makecell{27.19 \\ \scriptsize[25.68-28.84]} & 
\makecell{24.10 \\ \scriptsize[22.70-25.63]} & 
\makecell{54.55 \\ \scriptsize[52.02-57.13]} \\
\midrule
OpenBioLLM  & 70B & 2024.4 & 2.48 & 
\makecell{33.62 \\ \scriptsize[31.42-35.91]} & 
\makecell{20.24 \\ \scriptsize[18.86-21.49]} & 
\makecell{23.02 \\ \scriptsize[21.64-24.43]} & 
\makecell{42.80 \\ \scriptsize[40.48-45.12]} \\

Baichuan-M1 & 14B & 2025.2 & 2.53 & 
\makecell{22.88 \\ \scriptsize[20.81-24.90]} & 
\makecell{13.40 \\ \scriptsize[12.11-14.64]} & 
\makecell{15.52 \\ \scriptsize[14.13-16.91]} & 
\makecell{49.83 \\ \scriptsize[47.31-52.29]} \\

MedGemma    & 27B & 2024.7 & 3.88 & 
\makecell{35.48 \\ \scriptsize[33.57-37.48]} & 
\makecell{32.48 \\ \scriptsize[30.85-34.10]} & 
\makecell{29.64 \\ \scriptsize[28.17-31.12]} & 
\makecell{52.62 \\ \scriptsize[50.23-55.28]} \\

Baichuan-M3 & 235B& 2026.1 & 4.19 & 
\makecell{31.15 \\ \scriptsize[29.41-32.93]} & 
\makecell{30.33 \\ \scriptsize[28.85-31.92]} & 
\makecell{25.74 \\ \scriptsize[24.36-27.22]} & 
\makecell{51.69 \\ \scriptsize[49.17-54.15]} \\
\midrule
\rowcolor{mygray} \multicolumn{8}{c}{Agentic System} \\
\midrule
MedAgents   & -   & 2024.1 & 2.24 & 
\makecell{34.81 \\ \scriptsize[32.54-37.18]} & 
\makecell{19.22 \\ \scriptsize[17.92-20.60]} & 
\makecell{22.38 \\ \scriptsize[20.83-23.90]} & 
\makecell{55.47 \\ \scriptsize[53.02-57.80]} \\

MDAgents    & -   & 2024.10& 2.40 & 
\makecell{39.42 \\ \scriptsize[37.13-41.85]} & 
\makecell{22.31 \\ \scriptsize[21.01-23.71]} & 
\makecell{25.79 \\ \scriptsize[24.44-27.31]} & 
\makecell{54.88 \\ \scriptsize[52.49-57.33]} \\
\midrule
\rowcolor{mygray} \multicolumn{8}{c}{Our Method} \\
\midrule
DiagAgent   & 14B & -      & 6.88 & 
\makecell{26.05 \\ \scriptsize[25.10-27.07]} & 
\makecell{\textbf{57.26} \\ \scriptsize[55.43-59.11]} & 
\makecell{\textbf{33.63} \\ \scriptsize[32.48-34.81]} & 
\makecell{\textbf{63.84} \\ \scriptsize[61.44-66.29]} \\
\bottomrule
\end{tabular}
}
\end{table}

\clearpage

\subsection{Case Study}
\label{sec:case_study}
\subsubsection{Case study for DiagGym}
\label{sec:diaggym_case_study}

During physician evaluation, they also provide qualitative insights on each methods' simulation failure modes.
DeepSeek-v3-671B produces detailed, information-rich reports with broad coverage, but it frequently over-extrapolate from the diagnosis and clinical history, yielding overly severe positive findings. Also, narrative organization is sometimes disjointed with topic switching and repetition. 
Qwen2.5-72B offers comprehensive coverage but sometimes introduces unsupported neutral or false-positive findings with tenuous links to the core diagnosis.
MedGemma-27B, as a medical-domain LLM, is more concise and structurally clear with fewer off-topic assertions, though it intermittently exhibited logical inconsistencies. 
In contrast, DiagGym maintains balanced converge and close alignment with case context, largely avoids unwarranted positives and material contradictions.  

\begin{figure}[H]
    \centering
    \includegraphics[width=0.8\linewidth]{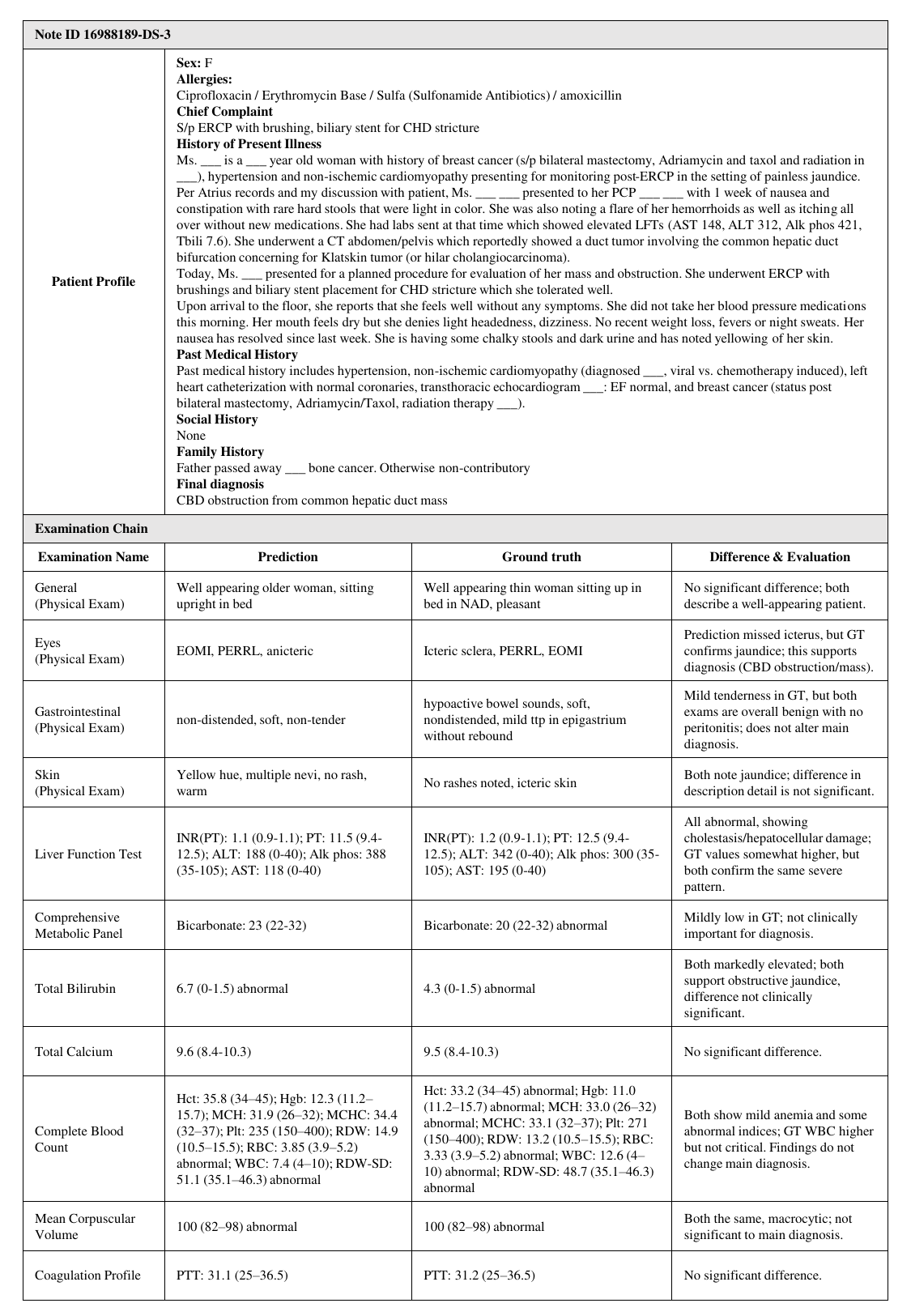}
    \caption{\textbf{Example Case Study from DiagGym: Comparison of Predicted and Ground Truth Examinations.} This case presents a single patient profile and final diagnosis, illustrating the step-wise evaluation setting. The core table compares the predicted examination results generated by DiagGym with the ground truth results in sequential order. The rightmost column analyzes key differences and discusses their clinical relevance in the diagnostic process.}
    \label{fig:case_study_simulator}
\end{figure}

Supplementary Figure~\ref{fig:case_study_simulator} compares outputs from the DiagGym with ground-truth examination results for a representative patient. The case involves a woman presenting with painless jaundice, with a history of breast cancer and non-ischemic cardiomyopathy. Upon admission, laboratory tests including liver function and bilirubin levels both supported a diagnosis of biliary obstruction; the final confirmed diagnosis was `CBD obstruction from common hepatic duct mass'.

The world model’s simulated predictions closely match the true clinical findings. A key indicator, \textbf{Total Bilirubin}, was predicted at 6.7 mg/dL versus the recorded value of 4.3 mg/dL (both well above the reference range of 0–1.5 mg/dL). This pronounced hyperbilirubinemia is consistent with the patient’s presentation (jaundice, scleral icterus) and strongly supports the diagnosis of 
``CBD obstruction from common hepatic duct mass''.
Minor numerical variations occur across other laboratory results, but none alter the clinical interpretation. Such variability illustrates the model’s ability to generate plausible but non-identical results, preserving both realism and diversity.

\subsubsection{Case study for DiagAgent}
\label{sec:diagagent_case_study}
\begin{figure}[H]
    \centering
    \includegraphics[width=0.95\linewidth]{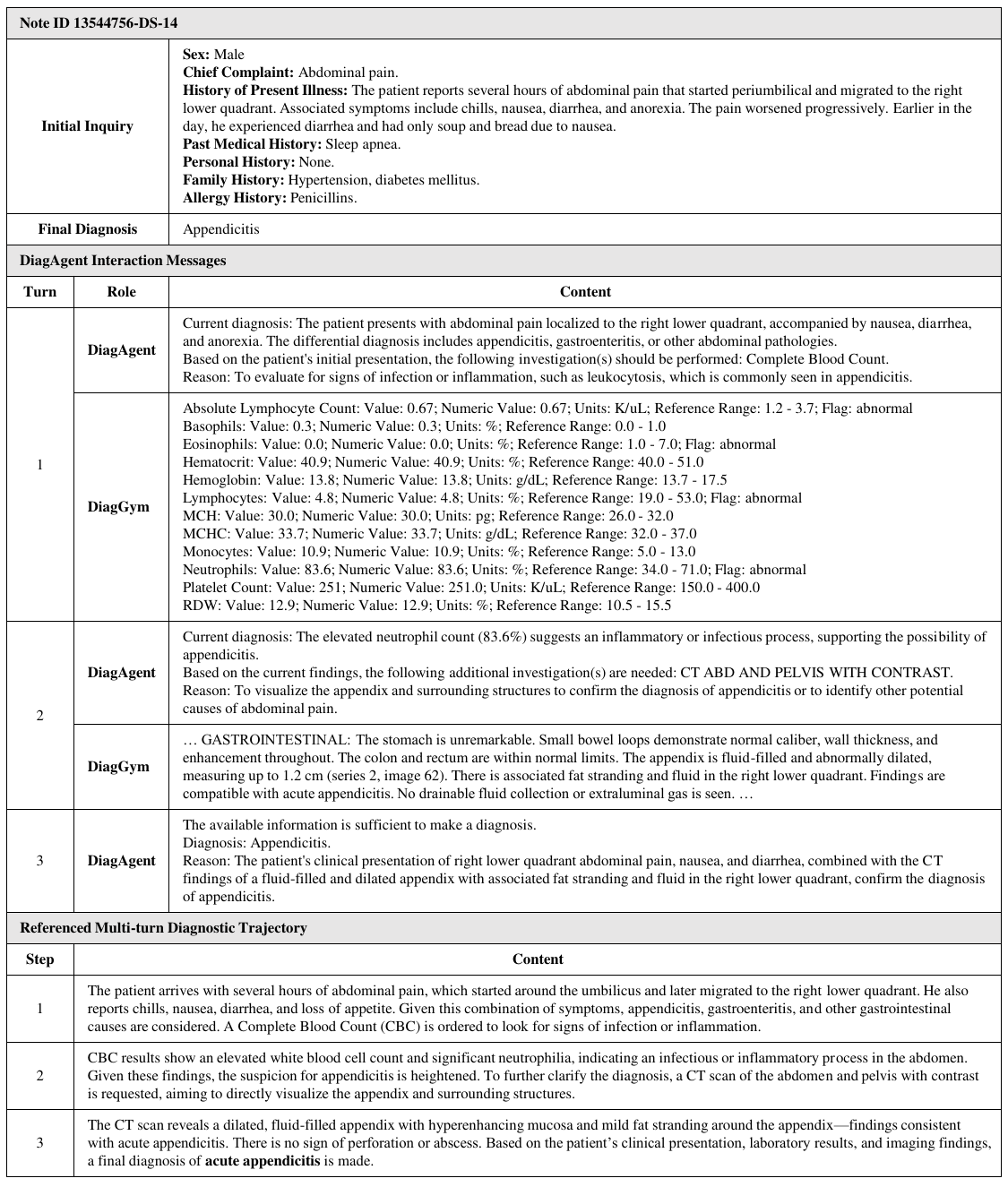}
    \caption{\textbf{Interactive Diagnostic Case Study with DiagAgent: Model Trajectory and Reference Timeline.} This figure illustrates a multi-turn interaction between the DiagAgent model and a simulator. The table details the agent's step-wise reasoning, differential diagnosis, and subsequent actions (e.g., ordering lab tests). The bottom Referenced Multi-Turn Trajectory provides a ground-truth clinical timeline for comparison, demonstrating the established diagnostic process leading to the final diagnosis.}
    \label{fig:case_study_diagnoser}
\end{figure}

\begin{figure}[H]
    \centering
    \includegraphics[width=0.95\linewidth]{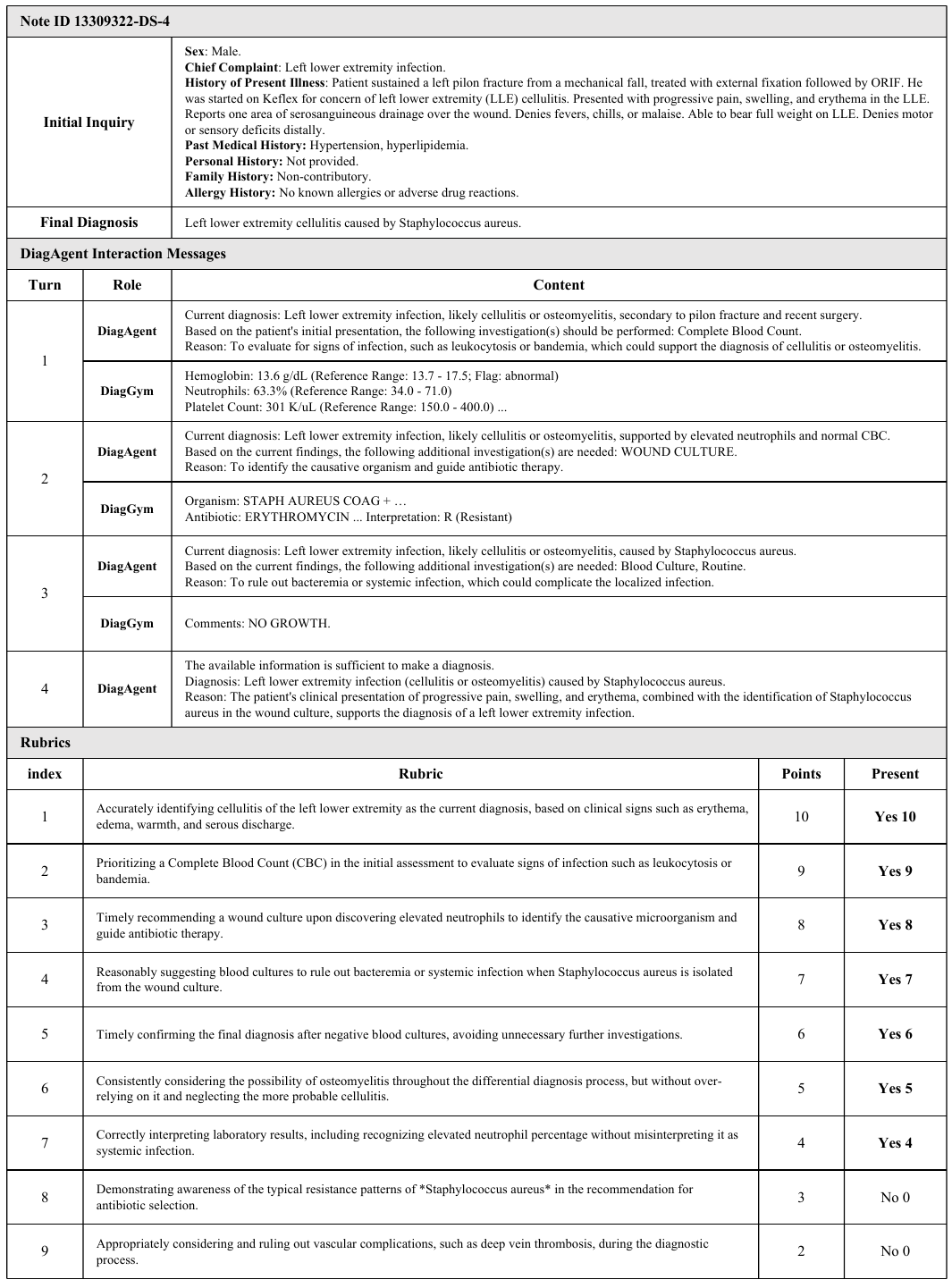}
    \caption{\textbf{Illustrative Success Case of DiagAgent Evaluated by Physician-Curated Rubrics.} This figure presents a case showcasing DiagAgent's high-quality multi-turn interaction for a lower extremity infection. The top section details the agent's step-wise dialogue. The bottom table of Procedural Evaluation Rubrics confirms that the agent successfully satisfied the majority of high-weighted, process-oriented criteria, demonstrating procedural integrity beyond merely achieving the correct final diagnosis.}
    \label{fig:rubric_success_mode}
\end{figure}

\begin{figure}[H]
    \centering
    \includegraphics[width=0.85\linewidth]{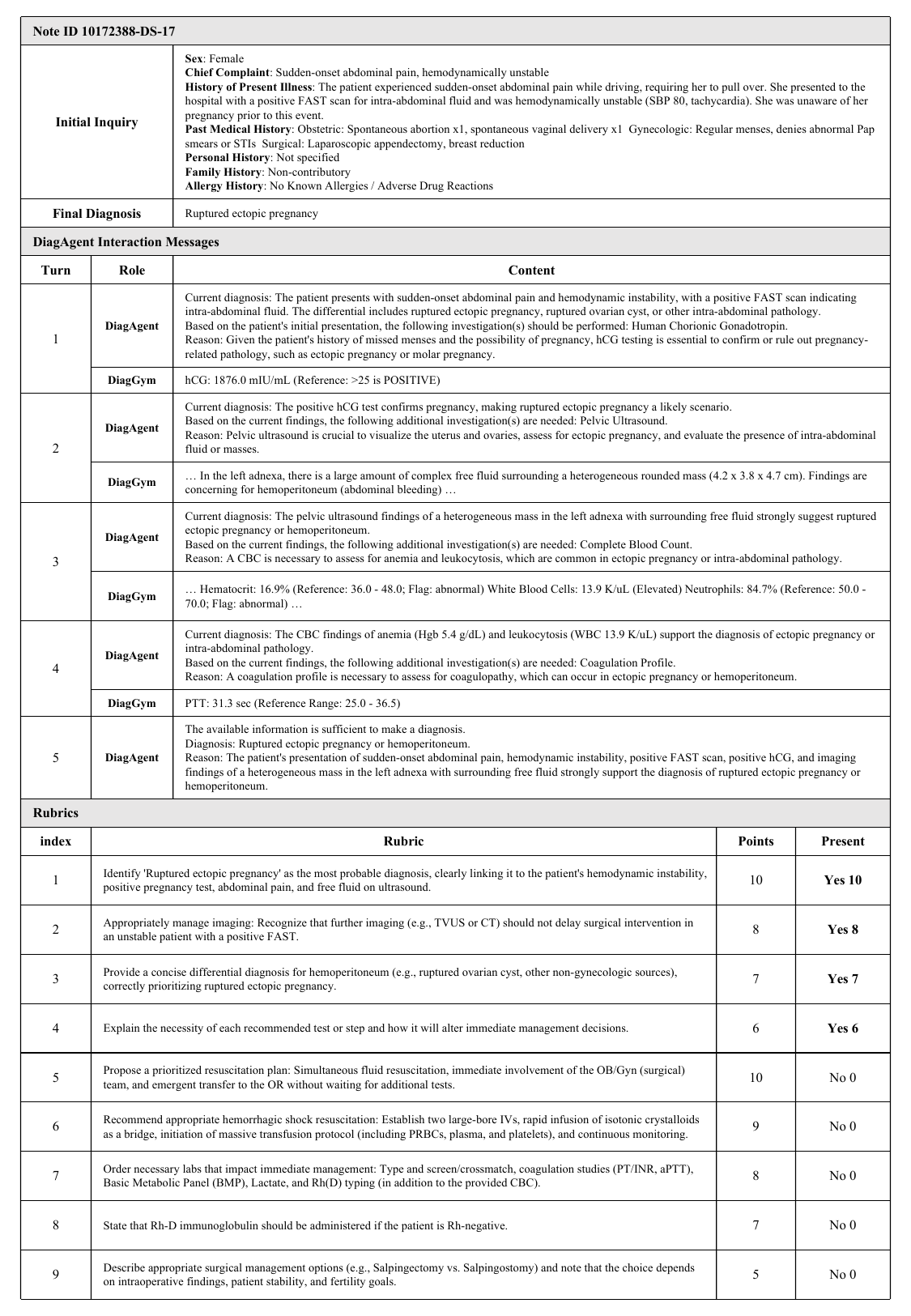}
    \caption{\textbf{Illustrative Fail Case of DiagAgent: Diagnostic Strength in Contrast to Management Deficit.} This case study (ruptured ectopic pregnancy) highlights a current limitation of DiagAgent. While the agent's Interactive Messages show robust step-wise diagnostic reasoning, successfully leading to the correct final diagnosis, the table of Procedural Evaluation Rubrics reveals critical omissions. Specifically, the agent fails to satisfy high-weighted criteria related to immediate emergency management (e.g., fluid resuscitation and surgical transfer). However, as a primary diagnostic model, the agent's core capability of accurate differential diagnosis and sequential information gathering remains intact, which is its main contribution.}
    \label{fig:rubric_fail_mode}
\end{figure}

Here we provide 2 cases for DiagAgent. Specifically, Supplementary Figure~\ref{fig:case_study_diagnoser} demonstrates the agent's ability to mirror real-world clinical timelines through logical, evidence-based information gathering, while Supplementary Figure~\ref{fig:rubric_success_mode} and Supplementary Figure~\ref{fig:rubric_fail_mode} showcases how its diagnostic quality is validated against rigorous, physician-curated rubrics.

\clearpage
\subsection{Prompt Collection}
\label{sec:prompt_collection}

\change{To facilitate reproducibility and provide transparency regarding our experimental setup, we detail our prompt engineering strategies and provide a full catalog of prompts used in this work.}

\subsubsection{\change{Prompt Design Principles}}
\change{Our prompt design is strictly \textbf{task-oriented}, categorized into three distinct phases based on their specific utility in the pipeline: Data Construction, Model Inference, and Evaluation. The prompting strategies (Zero-shot vs. One-shot) were selected to best serve the objective of each phase:}

\begin{itemize}
    \item \change{\textbf{Data Construction \& Filtering:} The objective of this phase is to convert raw EHRs into high-quality, structured datasets without hallucinations or parsing errors. Consequently, we employed a \textbf{hybrid strategy}. We utilized \textbf{One-shot} prompting for tasks requiring strict adherence to complex output schemas (to minimize formatting errors) and \textbf{Zero-shot} prompting for semantic extraction tasks. The validity of these prompts is guaranteed by our rigorous post-generation human verification process.}

    \item \change{\textbf{Model Inference:} For benchmarking baseline models, our primary goal is to assess their intrinsic capabilities fairly. We generally adopted a \textbf{Zero-shot} strategy to simulate realistic user behavior, where users typically interact with diagnostic agents without providing specific examples. 
    \textit{Exception:} The only exception is the simulation of numerical laboratory events (\textbf{Prompt~\ref{prompt:instruction_to_generate_as_simulator_labevent}}). Preliminary experiments showed that without a demonstration, baseline models generated heterogeneous units and formats, which unfairly penalized them during automated parsing. Thus, a \textbf{One-shot} example was introduced solely to constrain the output format for valid metric calculation.}

    \item \change{\textbf{Evaluation:} For the ``LLM-as-a-Judge'' metrics, we employed \textbf{Zero-shot} prompts. The tasks (e.g., binary classification of correctness, counting items) are straightforward and do not require few-shot learning. The reliability of these prompts was validated through a consistency study against human experts.}
\end{itemize}

\change{Supplementary Table~\ref{tab:prompt_catalog} provides a comprehensive catalog of all prompts, their specific utility, and the corresponding prompting strategy.}

\begin{table}[h!]
    \centering
    \scriptsize
    \renewcommand{\arraystretch}{1.3}
    \caption{Catalog of all prompts used in DiagGym and DiagAgent, categorized by their specific utility in the pipeline.}
    \label{tab:prompt_catalog}
    \begin{tabularx}{\textwidth}{@{}l >{\raggedright\arraybackslash}X l c@{}}
        \toprule
        \textbf{Usage Category} & \textbf{Design Purpose / Utility} & \textbf{Prompting Strategy} & \textbf{Ref.} \\
        \midrule
        \multicolumn{4}{@{}l}{\textit{\textbf{Data Construction \& Filtering}}} \\
        \multirow{2}{*}{MIMIC-IV Processing} 
         & Check for data leakage (diagnosis appearing in history) & Zero-shot & Prompt~\ref{prompt:check_if_diagnosis_in_past_medical_history} \\
         & Reformat physical examination text into structured JSON & Zero-shot & Prompt~\ref{prompt:reformat_physical_exam_into_json} \\
        \addlinespace
        \multirow{2}{*}{DiagAgent Training Data} 
         & Generate initial inquiry, trajectory, and diagnosis from EHRs & Zero-shot & Prompt~\ref{prompt:generate_differential_diagnosis_data} \\
         & Quality check to ensure no diagnosis leakage in generated data & Zero-shot & Prompt~\ref{prompt:filter_no_data_leakage_differential_diagnosis_data} \\
        \addlinespace
        \multirow{3}{*}{DiagBench (OOD) Curation} 
         & Extract structured patient profile from raw medical text & One-shot & Prompt~\ref{prompt:generate_patient_profile} \\
         & Extract all relevant examination names from text & One-shot & Prompt~\ref{prompt:extract_all_exams} \\
         & Consistency check between patient profile and diagnostic info & One-shot & Prompt~\ref{prompt:consistency_patient_profile_and_diag} \\
        
        \midrule
        \multicolumn{4}{@{}l}{\textit{\textbf{Model Inference (System Instructions)}}} \\
        \multirow{3}{*}{DiagGym (Simulator)} 
         & Simulate general examination results & Zero-shot & Prompt~\ref{prompt:instruction_to_generate_as_simulator} \\
         & Simulate numerical laboratory event results & One-shot & Prompt~\ref{prompt:instruction_to_generate_as_simulator_labevent} \\
         & Simulate free-text radiology reports & Zero-shot & Prompt~\ref{prompt:instruction_to_generate_as_simulator_radiology} \\
        \addlinespace
        DiagAgent (Agent) 
         & Instruct LLMs to perform interactive diagnosis & Zero-shot & Prompt~\ref{prompt:instruction_for_llm_diagnose} \\

        \midrule
        \multicolumn{4}{@{}l}{\textit{\textbf{Evaluation (LLM-as-a-Judge)}}} \\
        \multirow{2}{*}{Simulator Eval} 
         & Evaluate step-wise similarity against ground truth & Zero-shot & Prompt~\ref{prompt:eval_similarity_step_wise} \\
         & Evaluate full-chain internal clinical consistency & Zero-shot & Prompt~\ref{prompt:eval_full_chain_consistency} \\
        \addlinespace
        \multirow{4}{*}{Agent Eval (Auto Metrics)} 
         & Judge correctness of Final Diagnosis (Accuracy) & Zero-shot & Prompt~\ref{prompt:accuracy_in_diagnose_result} \\
         & Count matching exams for Precision calculation & Zero-shot & Prompt~\ref{prompt:precision_in_exam_recommendation} \\
         & Count matching exams for Recall calculation & Zero-shot & Prompt~\ref{prompt:recall_in_exam_recommendation} \\
         & Check if recommended exam is in valid list (Hit Ratio) & Zero-shot & Prompt~\ref{prompt:instruction_to_check_if_exam_in_the_list} \\
        \addlinespace
        Agent Eval (Rubrics) 
         & Assess if a specific rubric criterion is satisfied & Zero-shot & Prompt~\ref{prompt:check_each_rubric_hit} \\
        \bottomrule
    \end{tabularx}
\end{table}


\clearpage
\subsubsection{Full Prompt Content}
\label{sec:appendix_prompts}

\begin{prompt}
\label{prompt:check_if_diagnosis_in_past_medical_history}
\textbf{Prompt to check whether the final diagnosis appears in the patient's past medical history.}\newline
You are a highly knowledgeable and detail-oriented medical expert. Your task is to analyze and compare the provided discharge diagnoses with the patient's past medical history to determine whether any of the diagnoses in the discharge diagnoses are explicitly mentioned in the past medical history. Just output Yes or No without any other word.\newline

INPUT DATA:\newline
Discharge Diagnoses: \{discharge\_diagnosis\}\newline
Past Medical History: \{past\_medical\_history\}\newline

OUTPUT FORMAT:\newline
Yes/No\newline

\end{prompt}

\begin{prompt}
\label{prompt:reformat_physical_exam_into_json}
\textbf{Prompt to structure physical examination results as JSON.}\newline
You are a highly skilled and detail-oriented medical expert. Your task is to analyze a given physical exam report written in free text and convert it into a structured JSON format. Each JSON entry should represent a single examination that is typically completed in one step (e.g., height and weight measured together, blood pressure as a single reading). Do not split examinations into smaller components unless they are explicitly presented as separate tests in the input. Only include pre-admission physical examinations that are relevant to the initial diagnosis. Do not include any post-admission or discharge-related physical examination information. Just output the JSON without any additional text or explanation.\newline

INPUT DATA:\newline
Physical Exam: \{physical\_exam\}\newline

OUTPUT FORMAT:
\begin{verbatim}
[
    {
        "exam_name": "Name of the exam",
        "exam_results": "A string containing the results of the exam"
    }
]
\end{verbatim}
\end{prompt}

\begin{prompt}
\label{prompt:generate_differential_diagnosis_data}
\textbf{Prompt to generate differential diagnosis data based on discharge notes.}\newline
As an experienced physician, you will receive a patient Electronic Health Record focused on diagnosis. Your task is to:\newline
- Summarize only the key clinical information available prior to hospital presentation (i.e., the patient's state before arrival at the hospital, including symptoms, history, and relevant background). Do not include any information from the hospital course, ancillary tests, laboratory or imaging results, treatments, procedures, or discharge summaries.\newline
- Present a Key Pertinent Results section, listing essential diagnostic tests and their results using the provided Pertinent results dict.\newline
- Present a Stepwise Diagnostic Reasoning Timeline, outlining the chronological diagnostic process, including the rationale for each investigation, the corresponding results, and how each step informed subsequent decisions.\newline
- State the final diagnosis and briefly explain the supporting evidence.\newline

Ensure all summaries are concise, accurate, and based solely on the information provided. Do not reference images, tables, or other visual data, as these are not accessible.\newline
If the EHR is incomplete or does not meet the criteria for summarization, simply output: "I can't."\newline

Format to follow:\newline

\#\#\# Case Summary\newline
Provide a detailed medical history of the patient prior to hospital arrival, including chief complaint, history of present illness, past medical history, family history, and any other relevant information for initial diagnosis. Do not include any findings, investigations, or events that occurred after hospital arrival. Do not include any ancillary tests or pertinent results here. Do not mention or imply any diagnostic conclusions. Do not include any language that attributes symptoms to a specific diagnosis, even if this is present in the EHR. \newline

- Patient Information:\newline
- Chief Complaint: If none, write "None."\newline
- History of Present Illness: If none, write "None." This needs to be done very carefully and should only include information from before the visit.\newline
- Past Medical History: If none, write "None."\newline
- Personal History: If none, write "None."\newline
- Family History: If none, write "None."\newline
- Allergy History: If none, write "None."\newline
...(other necessary information before hospital arrival)\newline

\#\#\# Key Pertinent Results\newline
Please output the key diagnostic tests and their results in the following json format. Copy all test names and results exactly as they appear in the given Pertinent results dict. Do not change any word:
\begin{verbatim}
{
  "Test Name 1": "Result 1",
  "Test Name 2": "Result 2",
  ...
}
\end{verbatim}

\#\#\# Stepwise Diagnostic Reasoning Timeline\newline
Present a time-ordered, step-by-step diagnostic reasoning process, as follows:\newline
1. Based on the initial patient presentation, provide a preliminary diagnostic impression.\newline
Current diagnosis: [Analyses and preliminary diagnostic impression based on initial presentation]\newline
2. State which diagnostic tests should be ordered next and give the detailed specific reason for each. You cannot select an examination that does not appear in pertinent results. Do not change the name from the pertinent results.\newline
Test to order: [Test Name]\newline
Reason: [Long, comprehensive and detailed specific reason for ordering the test]\newline
3. For each test, copy the result exactly from the Pertinent results value, do not change any word.\newline
Test result: [Test result]\newline
4. After each set of results, update the diagnostic assessment based on information gathered so far, then explain what additional tests are required. Repeat steps 2-4 as needed.\newline
5. Conclude with the final diagnostic decision and the reasoning based on the available data.\newline
Diagnosis: [Final diagnosis]\newline
Reason: [Explanation for the final diagnosis based on all available information]\newline
6. A maximum of 12 turns is allowed.\newline

The following is an example of the format to follow:\newline

Step 1:\newline
Current diagnosis: [Analyses and preliminary diagnostic impression based on initial presentation]\newline

Based on the patient's initial presentation, the following investigation(s) should be performed: [Test Name].\newline
Reason: [Long, comprehensive and detailed reason for ordering the test]\newline
Test result: [Result]\newline

Step 2:\newline
Current diagnosis: [Updated analyses and diagnostic impression based on available information]\newline

Based on the current findings, the following additional investigation(s) are needed: [Test Name].\newline
Reason: [Long, comprehensive and detailed reason for ordering the test]\newline
Test result: [Result]\newline

...\newline

Step n:\newline
Current diagnosis: [Final diagnostic impression]\newline

The available information is sufficient to make a diagnosis.\newline
Diagnosis: [Final diagnosis]\newline
Reason: [Explanation and justification for the final diagnosis based on the findings above]\newline

\#\#\# Final Diagnosis\newline
Integrate the patient's clinical presentation, test results, and differential diagnosis process to summarize the final diagnosis. Briefly explain the basis for the diagnosis and highlight the key factors supporting this conclusion.\newline

\#\#\# Diagnosis results\newline
Just Output the diagnostic result without any other explanation.\newline

The following is the Electronic Health Record (EHR) of the patient:\newline

\{ehr\_text\}\newline

The following is all the test results of the patient according to time and date. Please copy all test names and results exactly as they appear in the following dictionary:\newline

\{events\}
\end{prompt}

\begin{prompt}
\label{prompt:filter_no_data_leakage_differential_diagnosis_data}
\textbf{Prompt to filter high-quality data for differential diagnoses, ensuring no data leakage.}\newline
Please evaluate whether the case summary provided below is qualified diagnostic task data according to the following standards. Please judge each criterion individually.\newline

\#\#\# Evaluation Criteria\newline

1. Information Leakage\newline
Does the "Case Summary" section directly contain the name or diagnostic results of the "Diagnose Results"?\newline

If diagnostic results appear directly (That is, without any reasoning), it is judged as "Unqualified".\newline

2. Chief Complaint Reasonableness\newline
Is the chief complaint in the "Case Summary" the patient's subjective discomfort or disease manifestation?\newline

If the chief complaint is a surgical procedure name or non-subjective discomfort, it is judged as "Unqualified".\newline

Output Format:
\begin{verbatim}
Information Leakage: xxx
Chief Complaint Reasonableness: xxx
\end{verbatim}

The following is specific information:\newline
Case Summary:\newline
\{case\_summary\}\newline

Key Examination Results:\newline
\{pertinent\_results\}\newline

Diagnose Results:\newline
\{diagnosis\_results\}
\end{prompt}

\begin{prompt}
\label{prompt:instruction_for_llm_diagnose}
\textbf{Prompt to instruct LLMs to make dynamic diagnosis}\newline
You are a medical AI assistant. Help the doctor with diagnosis by analyzing patient information, suggesting relevant tests, and providing a final diagnosis when sufficient information is available.\newline

RESPONSE FORMAT:\newline

If more information is needed:
\begin{verbatim}
Current diagnosis: [your diagnosis according to the information provided]
Based on the patient's initial presentation, the following investigation should be
performed: [one additional test]
Reason: [reason for the test]   
\end{verbatim}

If sufficient information exists for diagnosis:
\begin{verbatim}
The available information is sufficient to make a diagnosis. 

Diagnosis: [Diagnosis result]
Reason: [Diagnosis reason]
\end{verbatim}

\end{prompt}

\begin{prompt}
\label{prompt:instruction_to_check_if_exam_in_the_list}
\textbf{Prompt to check if the examination name occurs in the key examination list}\newline
Please determine if the predicted examination name matches any of the valid examination names for this case.\newline

Predicted examination: \{pred\_exam\}\newline
Valid examinations for this case: \{gt\_exam\}\newline

Please respond with "SAME" if the predicted examination matches any of the valid examinations (they refer to the same medical test or examination), or "DIFFERENT" if the predicted examination does not match any of the valid examinations.\newline

Consider examinations as the same if they:\newline
1. Are exactly the same name\newline
2. Are different names for the same medical test/procedure\newline
3. Are abbreviations or full forms of the same examination\newline
4. Are sub-items or components of any valid examination in gt\_exam\newline
5. Are encompassed by any valid examination in gt\_exam\newline

Please output SAME or DIFFERENT directly.\newline
\end{prompt}

\begin{prompt}
\label{prompt:instruction_to_generate_as_simulator}
\textbf{Prompt to instruct LLMs to simulate examination results}\newline
You are an expert medical AI assistant specialized in predicting medical examination results based on patient case summaries and past events. Your task is to analyze the provided patient information and predict the most likely results for a specific medical examination.\newline

Instructions:\newline
1. Carefully analyze the patient case summary, including diagnosis, symptoms, and clinical presentation\newline
2. Consider all past examination results and their implications\newline
3. Based on the medical context, predict realistic and clinically appropriate results for the requested examination\newline
4. Provide only the examination results without additional explanation or reasoning\newline
5. Format your response as concise, medically accurate examination findings\newline
6. If multiple measurements or findings are typical for the exam, include all relevant components\newline
7. Ensure your predictions are consistent with the patient's overall clinical picture\newline

Patient Case Summary:\newline
\{context\}\newline
\{past\_events\_text\}\newline

Current Examination to Predict:\newline
Exam name: \{exam\_name\}\newline

Please predict the most likely results for this examination based on the patient's clinical information and past results. Provide only the examination results.
\end{prompt}

\begin{prompt}
\label{prompt:instruction_to_generate_as_simulator_labevent}
\textbf{Prompt to instruct LLMs to simulate labevent results}\newline
You are an expert medical AI assistant specialized in predicting medical examination results. Your task is to analyze patient information and predict numerical values for specific laboratory tests.\newline

CRITICAL FORMATTING REQUIREMENTS:\newline
- You must provide numerical values for each requested sub-test\newline
- Format each result as: "Sub-test Name: Numeric Value: [number] Units: [unit]"\newline
- Use the exact units specified for each sub-test\newline
- If multiple sub-tests are requested, provide each on a separate line\newline
- Use realistic medical values appropriate for the patient's condition\newline
- Be precise with numbers (use decimals when appropriate)\newline

For the examination ``\{exam\_name\}'', you need to provide values for these specific measurements: \{subevents\_text\}\newline

Example format:\newline
Hemoglobin: Numeric Value: 12.5 Units: g/dL\newline
White Blood Cell Count: Numeric Value: 7200 Units: cells/$\mu$L\newline
Platelet Count: Numeric Value: 250000 Units: cells/$\mu$L
\end{prompt}

\begin{prompt}
\label{prompt:instruction_to_generate_as_simulator_radiology}
\textbf{Prompt to instruct LLMs to simulate radiology results}\newline
You are an expert radiologist AI assistant specialized in generating realistic radiology examination results. Your task is to analyze patient information and generate comprehensive radiology findings.\newline

CRITICAL FORMATTING REQUIREMENTS:\newline
- Generate a detailed and realistic radiology findings section for the specified examination\newline
- Include relevant anatomical findings.\newline
- Use appropriate medical terminology and standard radiological language\newline
- Provide specific details that would be clinically relevant\newline
- Ensure the findings are consistent with the patient's clinical presentation\newline

For the examination ``\{exam\_name\}'', generate a comprehensive radiology report that includes:\newline
1. Detailed findings of anatomical structures.\newline
2. Any abnormalities or normal variations observed.\newline

The report should be professional, detailed, and clinically appropriate.
\end{prompt}

\begin{prompt}
\label{prompt:precision_in_exam_recommendation}
\textbf{Prompt to instruct LLMs to count the number of the predicted examinations appearing in the key examinations list}\newline
Please determine how many of the recommended exams appear in the key exam list. Note that even if the expressions are different, if they refer to the same examination, they should be considered as matches.\newline

Key exam list: \{key\_exam\_names\}\newline
Recommended exam list: \{recommended\_exam\_names\}\newline

Please analyze each item in the recommended exam list and determine if it has a corresponding item in the key exam list (even with different expressions).\newline

Please only output the number of matches as an integer. For example: 3
\end{prompt}

\begin{prompt}
\label{prompt:recall_in_exam_recommendation}
\textbf{Prompt to instruct LLMs to count the number of key examinations appearing the predicted examinations}\newline
Please determine how many of the key exams appear in the recommended exam list. Note that even if the expressions are different, if they refer to the same examination, they should be considered as matches.\newline

Key exam list: \{key\_exam\_names\}\newline
Recommended exam list: \{recommended\_exam\_names\}\newline

Please analyze each item in the key exam list and determine if it has a corresponding item in the recommended exam list (even with different expressions).\newline

Please only output the number of matches as an integer. For example: 2
\end{prompt}

\begin{prompt}
\label{prompt:accuracy_in_diagnose_result}
\textbf{Prompt to instruct LLMs to assess the accuracy of the predicted final diagnosis.}\newline
\# Task Description
You are a professional medical diagnosis evaluation system. Now, you will receive two diagnosis results: one is the diagnosis predicted by the model ([pred\_diag]), and the other is the verified correct diagnosis ([gt\_diag]). Your task is to judge whether the model-predicted diagnosis([pred\_diag]) is correct.\newline

When evaluating, please consider the following factors:\newline
1.The same disease may have multiple aliases, for example, “Heart disease” may also be called “Cardiac disease”.\newline
2.There may be diversity in language expression, for example, “heart attack” and “myocardial infarction” may refer to the same disease.\newline
3.Only judge whether the diagnosis result is correct, information such as the cause of the disease, symptoms, and treatment recommendations are not included in the evaluation scope.\newline
4.If the correct diagnosis[gt\_diag] is included in the predicted diagnosis but some additional complications are mentioned, it is also considered correct\newline

\# Output Requirements\newline
Only output your judgment result on the model-predicted [pred\_diag] as “Correct|Wrong”, do not output any other content.\newline

\# Format to Follow:\newline
[Correct|Wrong]\newline

Below is the diagnosis result predicted by the model and the correct diagnosis:\newline

\{pred\_diag\}\newline

\{gt\_diag\}\newline
\end{prompt}

\begin{prompt}
\label{prompt:check_each_rubric_hit}
\textbf{Prompt to Assess Agent's Adherence to Diagnostic Rubric}\newline
Given a list of messages representing a medical conversation, evaluate whether the agent's diagnostic trajectory, represented by its recommendations throughout the messages, meets the provided criterion.\newline

- Output "Yes" if the agent's recommendations throughout the messages have satisfied the criterion.\newline
- Output "No" if the agent's recommendations throughout the messages have not satisfied the criterion.\newline

Conversation:\newline
\{messages\}

Criterion:\newline
\{criterion\}

Please only respond with "Yes" or "No".
\end{prompt}

\begin{prompt}
\label{prompt:eval_similarity_step_wise}
\textbf{Prompt to evaluate step-wise similarity between generated examination result and ground truth examination result}\newline
You are a medical expert evaluating similarity between AI predictions and ground truth results.\newline

Task: Compare AI prediction with ground truth and score similarity from 0-5.\newline

Scoring (0-5):\newline
- 0: Completely different, opposite findings\newline
- 1: Major differences, different clinical implications  \newline
- 2: Significant differences, different interpretations\newline
- 3: Moderate similarity, similar clinical direction\newline
- 4: High similarity, minor differences\newline
- 5: Excellent similarity, essentially equivalent\newline

Input:\newline
- Exam: \{exam\_name\}\newline
- Ground Truth: \{ground\_truth\}\newline
- AI Prediction: \{prediction\}\newline

Focus: Direct comparison of values/findings and clinical equivalence.\newline

Output Format:
\begin{verbatim}
{
    "score": [0-5],
    "explanation": "Brief explanation of similarity assessment and score reasoning"
}
\end{verbatim}

\end{prompt}

\begin{prompt}
\label{prompt:eval_full_chain_consistency}
\textbf{Prompt to evaluate full-chain consistency}\newline
You are a medical expert.\newline
Task: Determine if the complete AI prediction chain is internally consistent and aligns with the patient case (0=Fail, 1=Pass). You may use the ground truth chain (GT) as a reference. The evaluation criteria should be lenient, allowing for flexibility and minor discrepancies. If the AI prediction chain is similar to the GT chain, it should be considered correct. If the AI prediction chain is not similar to the GT chain but contains no significant factual errors, aligns with the patient case, and maintains clinical coherence, it should still be considered correct. Only assign a fail (0) if there are clear, significant factual errors or contradictions that severely undermine the clinical coherence or logic of the AI prediction chain. Minor differences or deviations are acceptable and should not lead to a fail.\newline

Input:\newline

Case Summary: \{case\_summary\}\newline
AI Prediction Chain: \{predicted\_chain\}\newline
Ground Truth Chain: \{ground\_truth\_chain\}\newline
Evaluation: Check for:\newline

- Internal contradictions between different results in the AI prediction chain\newline
- Alignment with the patient's clinical condition\newline
- Comparison with the ground truth chain (GT) for reference\newline
- Overall clinical coherence and logic\newline

Scoring:\newline

- 0 (Fail): Only assign a fail if there are clear and significant internal contradictions, conflicts with the patient case or GT, or clinically incoherent reasoning with factual errors that make the prediction chain unreliable.\newline
- 1 (Pass): Assign a pass if the AI prediction chain is internally consistent, aligns with the patient case, and forms a coherent clinical picture. This includes cases where the AI prediction chain differs from the GT chain, as long as it contains no significant factual errors and is clinically coherent. Even if there are notable differences from the GT, it should still pass unless there are explicit, critical contradictions or errors.\newline

Output Format:
\begin{verbatim}
{
    "score": [0 or 1],
    "explanation": "Brief analysis of chain coherence, comparison with ground truth, 
    and reasoning for pass/fail decision"
}
\end{verbatim}

\end{prompt}





\begin{prompt}
\label{prompt:generate_patient_profile}
\textbf{Prompt to extract patient profile from raw medical text.}

You are an expert clinical documentation specialist. Your task is to extract a Patient Admission Profile and a Final Diagnosis from the provided [Medical Text], mimicking the strict style and format of a MIMIC-IV Discharge Note.\newline

Instructions:\newline
1. Sufficiency Check: - Read the Case Report. If the information is too sparse (e.g., extremely short abstract, non-clinical text), output only: "Insufficient".\newline
   - If sufficient, proceed to generate the note.\newline

2. Extraction Scope (Strict Truncation):\newline
   - Extract sections in this exact order: Header -> History (HPI/PMH/SH/FH) -> Physical Exam -> Final Diagnosis.\newline
   - CRITICAL: Stop extraction immediately after the "Physical Exam". Do NOT include "Pertinent Results" (Labs/Imaging data tables) or the "Brief Hospital Course" (daily treatment progress).\newline
   - Jump directly to the "Final Diagnosis" section at the end.\newline

3. Format \& Style:\newline
   - Use the exact headers from the Example below.\newline
   - Use `\_\_\_` for any missing privacy info (Name, Unit No, Dates, Attending).
   - Physical Exam: Group findings by system (General, HEENT, CV, Lungs, Abdomen, Ext, Neuro). Distinguish "Admission Exam" if specified.\newline

4. Diagnosis Constraint:\newline
   - Final Diagnosis: Identify the ONE single primary diagnosis responsible for the admission or the main condition confirmed by the end of the case. Do not list multiple secondary diagnoses here unless they are inseparable from the main condition.\newline

\#\#\# Meidcal Text\newline
{medical\_text}\newline

Please output the patient profile directly.\newline
\end{prompt}

\begin{prompt}
\label{prompt:extract_all_exams}
\textbf{Prompt to extract relevant examination names from unstructured clinical text.}

As an experienced physician, you will receive a Transcribed Medical Note. Your task is to:\newline

- Identify and extract the diagnostic examinations mentioned throughout the entire Transcribed Medical Note. You must strictly limit your extraction to physical exams, Laboratory Events, Microbiology Events, or Radiology Events. Ignore procedures (like surgeries), or unrelated tests. No other types of examinations are permitted. Related events can be found in the Related Events Section. \newline
- Extract every occurrence of these events, regardless of whether the result was normal, abnormal, or incidental. The goal is a complete inventory of tests performed.\newline
- A maximum of 30 most related events. Similar events can be combined to avoid duplication.\newline

Output format:\newline
A list of all related examination names.\newline
["Complete Blood Count", "MR BRAIN", ...]\newline

The following is the Transcribed Medical Note of the patient:\newline

\{case\_report\_text\}\newline

Please output the list directly without any other explanation.

\end{prompt}

\begin{prompt}
\label{prompt:consistency_patient_profile_and_diag}
\textbf{Prompt to assess logical consistency between patient profile and diagnostic trajectory.}

You are a medical expert quality controller. \newline
Task: Determine if the "Profile Diagnosis" and the "Diagnostic Info" refer to the same or synonymous medical condition.\newline

Profile Diagnosis:\newline
\{profile\_diagnosis\}\newline

Diagnostic Info / Trajectory:\newline
\{diagnose\_info\_text\}\newline

Question: Is the final diagnosis in the [Diagnostic Info] consistent with the [Profile Diagnosis]? \newline
Focus on clinical equivalence (e.g., "CML" == "Chronic Myelogenous Leukemia").\newline
Answer only "Yes" or "No".\newline
\end{prompt}